\documentclass[10pt,twocolumn,letterpaper,table,hyphens]{article}

\usepackage{booktabs} %
\usepackage{colortbl}
\usepackage{tabularx}
\usepackage{multirow}
\usepackage[table,x11names,dvipsnames,table]{xcolor}
\usepackage{tikz}
\usepackage{pgfplots}
\usetikzlibrary{spy}
\usetikzlibrary{matrix}

\definecolor{rowblue}{RGB}{220,230,240}
\definecolor{myorchid}{RGB}{150,10,30}
\definecolor{myblue}{RGB}{10,30,250}
\definecolor{mygreen}{RGB}{10,190,10}

\newcommand{\supp}[1]{#1}

\newcommand{\camready}[1]{#1}
\newcommand{\camlast}[1]{#1}

\newcommand{\myparagraph}[1]{\vspace{.2cm} \noindent \textbf{#1} \quad}
\newcommand{\etal}{\textit{et al.}}

\usepackage{enumitem}

\usepackage{iccv}
\usepackage{times}
\usepackage{epsfig}
\usepackage{graphicx}
\usepackage{amsmath}
\usepackage{amssymb}
\usepackage{bbm}
\usepackage{pifont}

\usepackage{subfiles}
\usepackage{color}
\usepackage{enumitem}
\usepackage{dsfont}
\usepackage{booktabs}
\usepackage{float}
\usepackage{capt-of,etoolbox}
\usepackage[font={small}]{caption}
\usepackage{subcaption}
\usepackage{multirow}
\usepackage{arydshln}

\usepackage[pagebackref=true,breaklinks=true,letterpaper=true,colorlinks,bookmarks=false,hypertexnames=false]{hyperref}

\iccvfinalcopy %

\def\iccvPaperID{1648} %

\ificcvfinal\fi
\begin{document}

\title{Detecting Photoshopped Faces by Scripting Photoshop}

\author{Sheng-Yu Wang\textsuperscript{1}
\and
Oliver Wang\textsuperscript{2}
\and
Andrew Owens\textsuperscript{1}
\and
Richard Zhang\textsuperscript{2}
\and
Alexei A. Efros\textsuperscript{1}
\\\and 
UC Berkeley\textsuperscript{1}
\and
Adobe Research\textsuperscript{2}
}

\twocolumn[{%
\renewcommand\twocolumn[1][]{#1}%
\maketitle
\newcommand\myspy[4]{\spy [white,
		spy connection path={\draw[thick, white] (tikzspyonnode) -- (tikzspyinnode);},
		every spy on node/.append style={thick},
		every spy in node/.append style={thick}] on ({#1},{#2}) in node [left] at ({#3},{#4})}
\begin{center}
    \centering
    \resizebox{1.\linewidth}{!}{
    \begin{tabular}{*{4}{c@{\hspace{1px}}}}
    \begin{tikzpicture}[spy using outlines={circle, magnification=3, size=1.85cm, connect spies},inner xsep=0cm]	
        \node (O) [] {
        	\includegraphics[width=.24\linewidth]{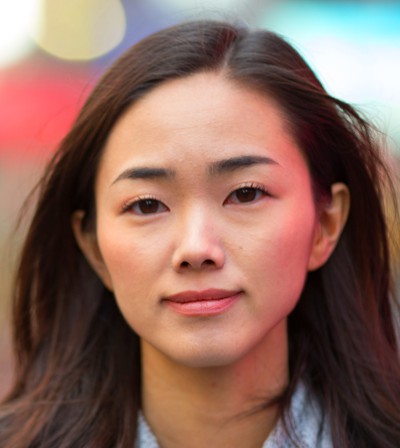}
        };
        \myspy{1.0}{-0.9}{2.0}{-2.4};
        \myspy{-0.33}{-0.82}{-0.1}{-2.1};
    \end{tikzpicture} &
    \begin{tikzpicture}[spy using outlines={circle, magnification=3, size=1.85cm, connect spies},inner xsep=0cm]	
        \node (O) [] {
        	\includegraphics[width=.24\linewidth]{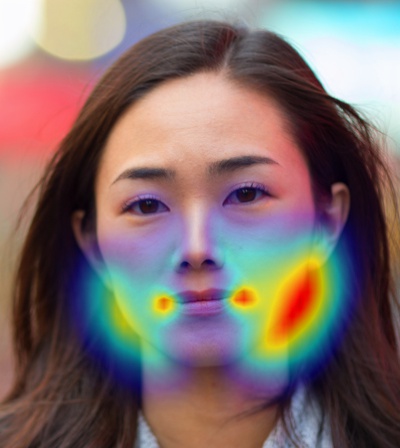}
        };
        \myspy{1.0}{-0.9}{2.0}{-2.4};
        \myspy{-0.33}{-0.82}{-0.1}{-2.1};
    \end{tikzpicture} &
    \begin{tikzpicture}[spy using outlines={circle, magnification=3, size=1.85cm, connect spies},inner xsep=0cm]	
        \node (O) [] {
        	\includegraphics[width=.24\linewidth]{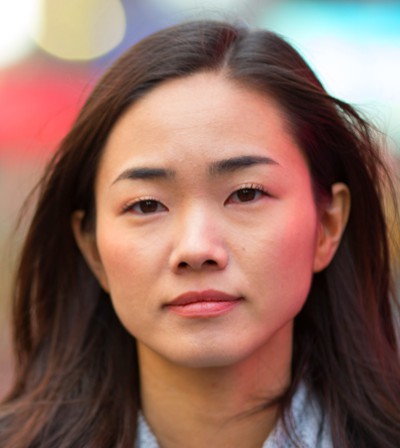}
        };
        \myspy{1.0}{-0.9}{2.0}{-2.4};
        \myspy{-0.33}{-0.82}{-0.1}{-2.1};
    \end{tikzpicture} &
    \begin{tikzpicture}[spy using outlines={circle, magnification=3, size=1.85cm, connect spies},inner xsep=0cm]	
        \node (O) [] {
        	\includegraphics[width=.24\linewidth]{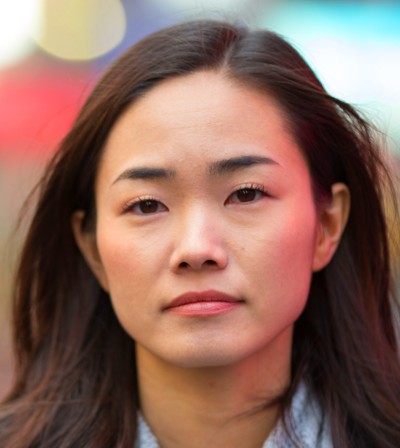}
        };
        \myspy{1.0}{-0.9}{2.0}{-2.4};
        \myspy{-0.33}{-0.82}{-0.1}{-2.1};
    \end{tikzpicture} \\
{\small (a) Manipulated photo} & 
{\small (b) Detected manipulations} & 
{\small (c) Suggested ``undo''} & 
{\small (d) Original photo} \\
    \end{tabular}
    }
    \vspace{-.1in}
    \captionof{figure}{Given an input face (a), our tool can detect that the face has been warped with the Face-Aware Liquify tool from Photoshop, predict where the face has been warped (b), and attempt to ``undo'' the warp (c) and recover the original image (d). \label{fig:teaser}
    }
\end{center}
}]

\maketitle

\begin{abstract}
\vspace{-.1in}

Most malicious photo manipulations are created using standard image editing tools, such as Adobe\textsuperscript{\textregistered} Photoshop\textsuperscript{\textregistered}.  We present a method for detecting one very popular Photoshop manipulation -- image warping applied to human faces -- using a model trained entirely using fake images that were automatically generated by scripting Photoshop itself. We show that our model outperforms humans at the task of recognizing manipulated images, can predict the specific location of edits, and in some cases can be used to ``undo" a manipulation to reconstruct the original, unedited image.  We demonstrate that the system can be successfully applied to real, artist-created image manipulations. 
\vspace{-.1in}

 \end{abstract}

\footnotetext[1]{The code and instructions can be found in our github repository (\href{https://peterwang512.github.io/FALdetector} {https://peterwang512.github.io/FALdetector}).}

\section{Introduction}

In an era when digitally edited visual content is ubiquitous, the public is justifiably eager to know whether the images they see on TV, in glossy magazines, and on the Internet are, in fact, {\em real}.  While the popular press has mostly focused on ``DeepFakes'' and other GAN-based methods that may one day be able to convincingly simulate a real person's appearance, movements, and facial expressions~\cite{wang2018vid2vid,chan2018dance,bansal2018,kim2018deep}, for now, such methods are prone to degeneracies and exhibit visible artifacts~\cite{McDoland18}.
Rather, it is the more subtle image manipulations, performed with classic image
processing techniques, typically in Adobe Photoshop, that have been the largest contributors to the proliferation of manipulated visual content~\cite{farid2016photo}. 
While such editing operations have helped enable creative expression, if done without the viewer's knowledge, they can have serious negative implications, ranging from body image issues set by unrealistic standards, to the consequences of ``fake news'' in politics.

\begin{figure*}[t]
    \centering
    {\def\arraystretch{.25}%
    \begin{tabular}{ c@{\hspace{1px}} c@{\hspace{1px}} c@{\hspace{1px}} c@{\hspace{1px}} c@{\hspace{1px}} c@{\hspace{8px}} c@{\hspace{1px}} c@{\hspace{1px}} c@{\hspace{1px}} c@{\hspace{1px}} c@{\hspace{1px}} c   }
    \includegraphics[width=.08\linewidth]{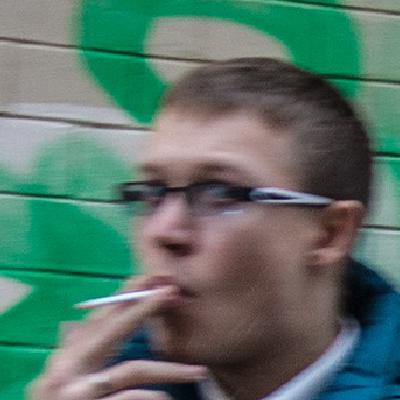} &
    \includegraphics[width=.08\linewidth]{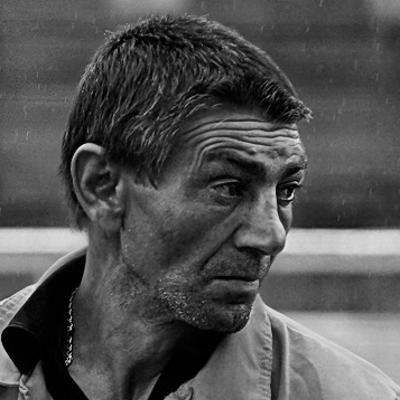} &
    \includegraphics[width=.08\linewidth]{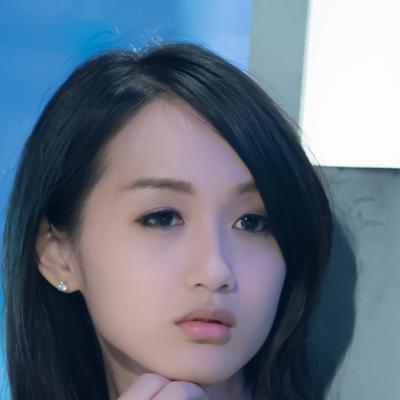} &
    \includegraphics[width=.08\linewidth]{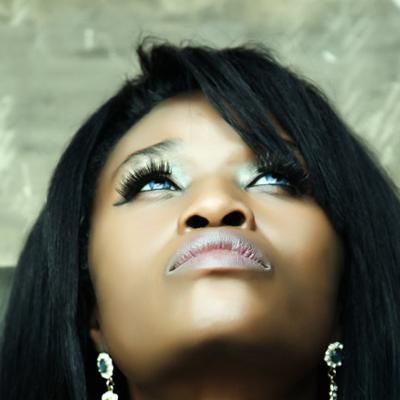} &
    \includegraphics[width=.08\linewidth]{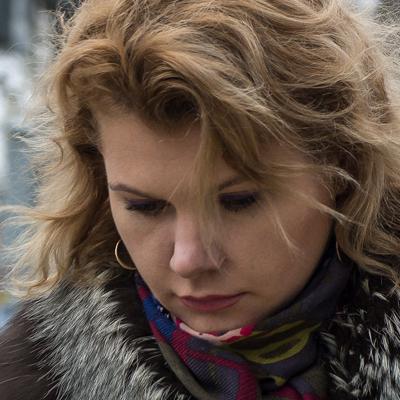} &
   \includegraphics[width=.08\linewidth]{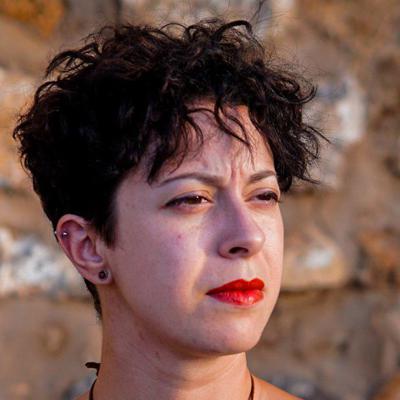} &
    \includegraphics[width=.08\linewidth]{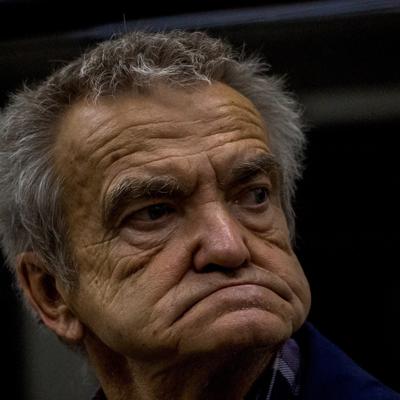} &
    \includegraphics[width=.08\linewidth]{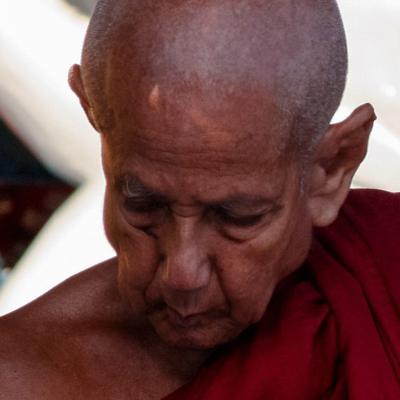} &
    \includegraphics[width=.08\linewidth]{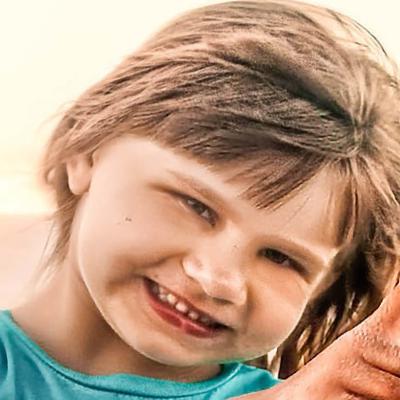} &
    \includegraphics[width=.08\linewidth]{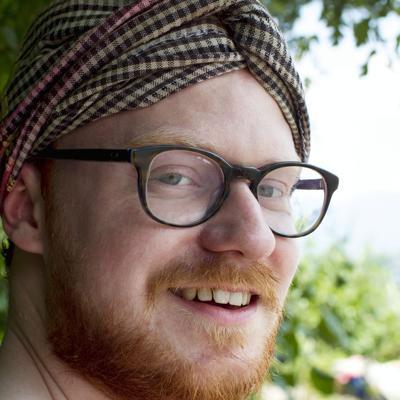} & 
    \includegraphics[width=.08\linewidth]{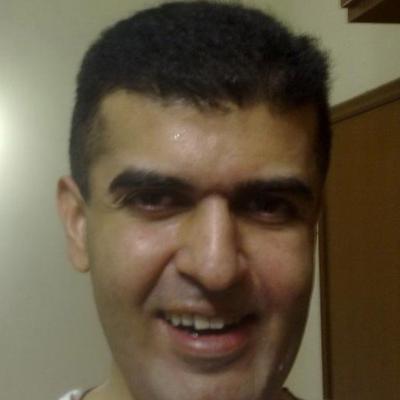} &
    \includegraphics[width=.08\linewidth]{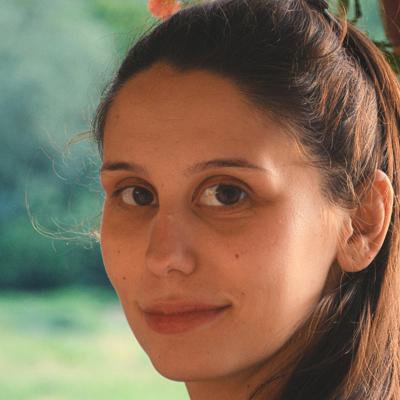} \\
    \includegraphics[width=.08\linewidth]{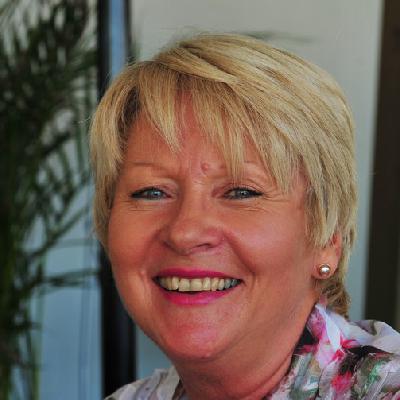} &
    \includegraphics[width=.08\linewidth]{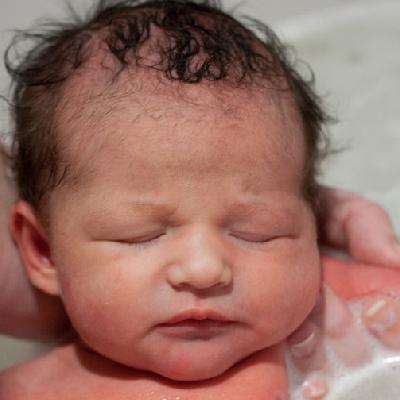} &
    \includegraphics[width=.08\linewidth]{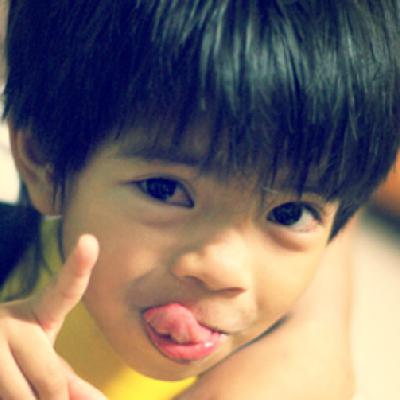} &
    \includegraphics[width=.08\linewidth]{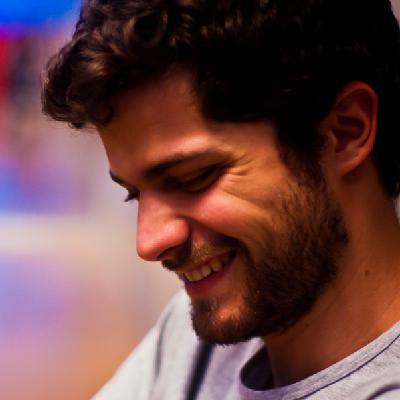} &
    \includegraphics[width=.08\linewidth]{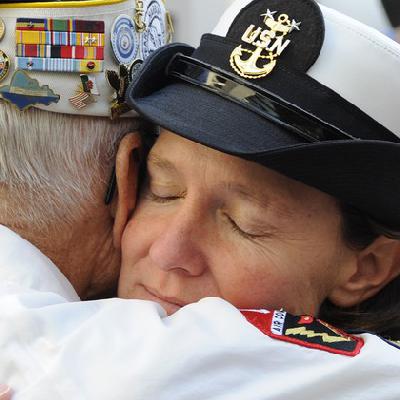} &
    \includegraphics[width=.08\linewidth]{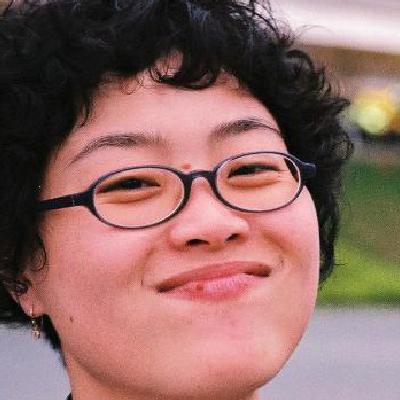} &
    \includegraphics[width=.08\linewidth]{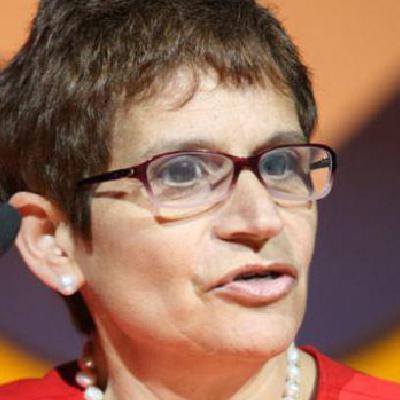} &
    \includegraphics[width=.08\linewidth]{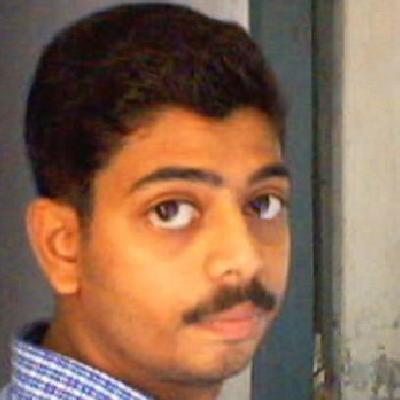} & 
    \includegraphics[width=.08\linewidth]{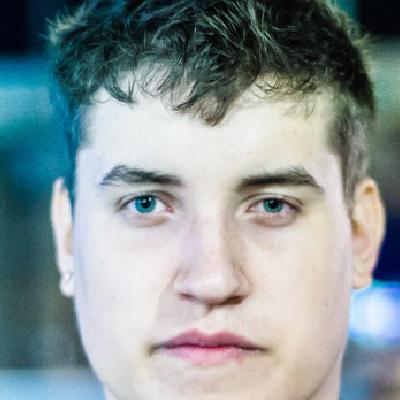} &
    \includegraphics[width=.08\linewidth]{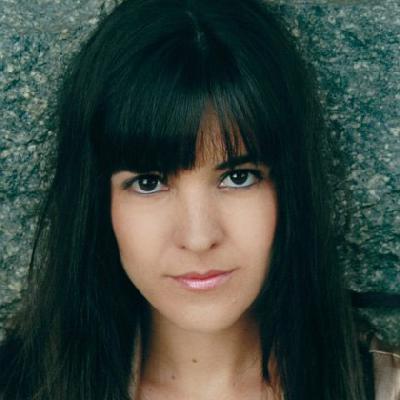} & 
    \includegraphics[width=.08\linewidth]{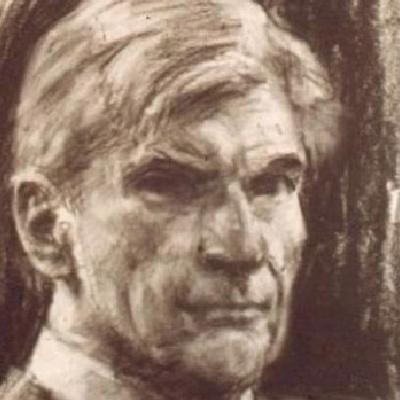} &
    \includegraphics[width=.08\linewidth]{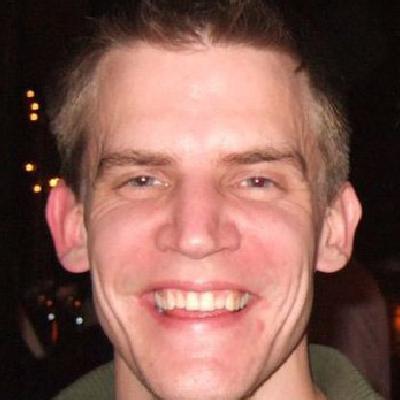} \\
    \addlinespace[3px]
    \multicolumn{6}{c}{\small (a) Real images} &
    \multicolumn{6}{c}{\small (b) Manipulated images} \\
    \end{tabular}}
    \vspace{-.1in}
    \caption{\label{fig:dataset}
    Random samples from our training dataset.   (a) Real images scraped from Flickr portraits (top) and Open Images~\cite{krasin2016openimages} (bottom).  (b) Random warps automatically created with Photoshop's {\em Face-Aware Liquify} tool.  The differences are very subtle.}  
    \vspace{-.15in}
\end{figure*}

In this work, we focus on one specific type of Photoshop manipulation -- image warping applied to faces. 
This is an extremely common task used for ``beautification'' and expression editing.
Face warping is an interesting problem as it is a domain that is surprisingly hard for people to detect, but it is commonly used and has wide reaching implications. 
We show in a user study that humans have only $53.5\%$ accuracy in identifying such edits (chance is $50\%$).
We propose a \emph{lens} through which these subtle edits become visualized, alerting the viewer to the presence of modifications, as shown on Figure~\ref{fig:teaser}. 
Our proposed approach is but one tool in a larger toolbox of techniques that together, could be used to help combat the spread of misinformation, and its effects.

Our approach consists of a CNN carefully trained to detect facial warping modifications in images.
As with any deep learning method, collecting enough supervised training data is always a challenge.  This is especially true for forensics applications, since there are no large-scale datasets of manually created visual fakes.  In this work, we solve this problem by using Photoshop itself to automatically generate realistic-looking fake training data.   
We first collect a large dataset of real face images, scraped from different internet sources (Figure~\ref{fig:dataset}a). 
We then directly script the {\em Face-Aware Liquify} tool in Photoshop, which abstracts facial manipulations into high level semantic operations, such as ``increase nose width'' and ``decrease eye distance''.
By randomly sampling manipulations in this space (Figure~\ref{fig:dataset}b), we are left with a training set consisting of pairs of source images and realistic looking warped modifications. 

We train both global classification and local warping field prediction networks on this dataset. 
In particular, our local prediction method uses a combination of loss functions including flow warping prediction, relative warp preservation, and a pixel-wise reconstruction loss. 
We present a number of applications, including a visualization overlay to draw attention to modified regions, as in Fig.~\ref{fig:teaser}(b), and un-warping the manipulated image to make it more like the original, as in Fig.~\ref{fig:teaser}(c). 
Finally, we evaluate our approach on a number of test cases, including images scraped from various sources, as well as with warping operations performed by other means.

\section{Related work}

Image forensics, or forgery detection, is an increasingly important area of research in computer vision. 
In this section, we focus on works that are either trained from large amounts of data, or directly address the face domain.

\vspace{-2px}
\myparagraph{Face manipulation} 
Researchers have proposed forensics methods to detect a variety of face manipulations.
Zhou \etal ~\cite{zhou2017two} and Roessler \etal~\cite{rossler2018faceforensics,rossler2019faceforensics++} propose neural network models to detect {\em face swapping} and {\em face reenactment} --- manipulations where one face is wholly replaced with another (perhaps taken from the same subject) after splicing, color matching, and blending.
Other work investigates detecting morphed (interpolated) faces~\cite{raghavendra2017transferable} and inconsistencies in lighting from specular highlights on the eye \cite{johnson2007exposing}.
In contrast, we consider {\em facial warps} which undergo subtle geometric deformations, rather than a complete replacement of the face, or the synthesis of new details.

\vspace{-2px}
\myparagraph{Learning photo forensics}
The difficulty in obtaining labeled training data has led researchers to propose a variety of ``self-supervised'' image forensics approaches that are trained on automatically-generated fake images. Chen \etal~\cite{chen2015median} use a convolutional network to detect median filtering.
Zhou \etal \cite{zhou2018learning} propose an object detection model, specifically using steganalysis features to reduce the influence of semantics. The model is pretrained on automatically created synthetic fakes using object segmentations, and subsequently fine-tuned on actual fake images. 
\camready{While we also generate fakes automatically, we use the tools that a typical editor would use, allowing us to detect these manipulations more accurately.}
A complementary approach is exploring unsupervised forensics models that learn only from real images, without explicitly modeling the fake image creation process.
For example, several models have been proposed to detect spliced images by identifying patches which come from different camera models~\cite{bondi2017tampering,mayer2018learned},  by using EXIF metadata~\cite{huh2018fighting}, or by identifying physical inconsistencies~\cite{li2015segmentation}.
These approaches, however, are designed to detect instances of the image splicing problem, while we address a more subtle manipulation --- facial structure warping.

\vspace{-2px}
\myparagraph{Hand-defined manipulation cues} 
Other image forensics work has proposed to detect fake images using hand-defined cues~\cite{farid2016photo}. Early work detected resampling artifacts~\cite{popescu2005exposing,kirchner2008fast} by finding periodic correlations between nearby pixels. There has also been work that detects inconsistent quantization~\cite{agarwal2017photo}, double-JPEG artifacts~\cite{barni2017aligned,amerini2017localization}, and geometric inconsistencies~\cite{o2012exposing}. However, the operations performed by interactive image editing tools are often complex, and can be difficult to model. 
Our approach, by contrast, learns features appropriate for its task from a large dataset of manipulated images.

\section{Datasets}

We obtain a large dataset of real face images from the Open Images dataset~\cite{krasin2016openimages} and Flickr, and create two datasets of fakes: a large, automatically generated set of manipulated images for training a forensics model, and a smaller set of actual manipulations done by an artist for evaluation. Details of the data collection process are provided in Appendix~\ref{sec:apdxdatainfo}. %

\myparagraph{Generating manipulated face images} 
Our goal is to automatically create a dataset of manipulated images that, when leveraged for training, generalizes to artist-created fakes. We script the {\em Face-Aware Liquify (FAL)} tool~\cite{fal2016} in Adobe Photoshop to generate a variety of face manipulations, using built-in support for JavaScript execution. 
We choose Photoshop, since it is one of the most popular image editing tools, and this operation, as it is a very common manipulation in portrait photography.
FAL represents manipulations using 16 parameters, corresponding to higher-level semantics (\eg, adjusting the width of the nose, eye  distance, chin height, etc.). A facial landmark detector registers a mesh to the input image, and the parameters control the mesh's vertex displacements.
As shown in Figure~\ref{fig:teaser}, the tool can be used to make subtle, realistic manipulations, such as making a face more symmetrical. We \camready{randomly} sample the FAL parameter space.
While these parameter choices are unlikely to match the changes an artist would make, we argue, and validate, that randomly sampling the space will cover the space of ``realistic'' operations.
We modify each image from our real face dataset randomly 6 times. 
In all, the data we used for training is 1.295M faces -- 185K unmodified, and 1.1M modified. Additionally, we hold out 5K real faces each from Open Images and Flickr, leaving half of the images unmodified and the rest modified in the same way as the training data. In total, the validation data consists of 2.5K images in each categories -- \{Open Images, Flickr\} $\times$ \{unmanipulated, manipulated\}. Table~\ref{tab:datasettable} summarizes that data and  Figure~\ref{fig:dataset} shows random samples.

\begin{table}[t]
\scalebox{.9} {
 \begin{tabular}{l c c c}
 \toprule
 & {\bf Train} & {\bf Val} & {\bf Test} \\ \hline
 {\bf Source} & \multicolumn{2}{c}{OpenImage \& Flickr} & Flickr \\
 {\bf Total Images} & 1.1M & 10k & 100 \\
 {\bf Unmanipulated images} & 157k & 5k & 50 \\
 {\bf Manipulated images} & 942k & 5k & 50 \\
 {\bf Manipulations} & \multicolumn{2}{c}{Random FAL} & Pro Artist \\

\bottomrule
\end{tabular}
}
\vspace{-.1in}
\caption{{\bf Dataset statistics.} This includes our own automatically created data as well as a smaller test set of manipulations created by a professional artist.
\label{tab:datasettable}
\vspace{-.2in}
}
\end{table}

\myparagraph{Test Set: Artist-created face manipulations} We test the generalization ability to ``real" manipulations by contracting a professional artist to manipulate 50 real photographs. Half are manipulated with the intent of ``beautifying'', or increasing attractiveness, and the other half to change facial expression, positively or negatively. This covers two important use cases.
The artist created 50 images with the FAL tool, and 50 images with the more general Liquify tool -- a free-form brush used to warp images. On average, it took 7.8 minutes of editing time per image.
\section{Methods}

Our goal is to train a system to detect facial manipulations. 
We present two models: a {\em global classification} model, tasked with predicting whether a face has been warped, and a {\em local warp predictor}, which can be used to identify \emph{where} manipulations occur, and reverse them.%

\subsection{Real-or-fake classification} We first address the question ``has this image been manipulated?''
\camready{We train a binary classifier using a Dilated Residual Network variant (DRN-C-26)~\cite{Yu2017}. Details of the training procedure are provided in Appendix~\ref{sec:apdxtraininfo}.}

We investigate the effect of resolution by training low and high-resolution models. High-resolution models enable preservation of low-level details, potentially useful for identifying fakes, such as resampling artifacts. On the other hand, a lower-resolution model potentially contains sufficient details to identify fakes and can be trained more efficiently. We try low and high-resolution models, where the shorter side of the image is resized to 400 and 700 pixels, respectively. During training, the images are randomly \camready{left-right} flipped and cropped to 384 and 640 pixels, respectively.

While we control the post-processing pipeline in our test setup, real-world use cases may contain unexpected postprocessing. Forensics algorithms are often sensitive to such operations~\cite{popescu2005exposing}. To increase robustness, we consider more aggressive data augmentation, including resizing methods (bicubic and bilinear), JPEG compression, brightness, contrast, and saturation. We experimentally find that this increases robustness to perturbations at testing, even if they are not in the augmentation set.

\begin{table*}[t]
\begin{center}
\resizebox{1.\linewidth}{!}{
\setlength{\tabcolsep}{0.5em} %
\begin{tabular}{l c c c c c c c c c c c c}
\toprule
\multicolumn{3}{c}{\bf Algorithm} & \multicolumn{5}{c}{\bf Validation (Random FAL)} & \multicolumn{5}{c}{\bf Test (Professional Artist)} \\ \cmidrule(lr){1-3} \cmidrule(lr){4-8} \cmidrule(lr){9-13}
\multirow{2}{*}{\bf Method} & {\bf Resol-} & {\bf with} & \multicolumn{3}{c}{\bf Accuracy} & \multirow{2}{*}{\bf AP} & \multirow{2}{*}{\bf 2AFC} & \multicolumn{3}{c}{\bf Accuracy} & \multirow{2}{*}{\bf AP} & \multirow{2}{*}{\bf 2AFC} \\ \cmidrule(lr){4-6} \cmidrule(lr){9-11}
& {\bf ution} & {\bf Aug?} & {\bf \camlast{Total}} & {\bf Orig} & {\bf Mod} & & & {\bf \camlast{Total}} & {\bf Orig} & {\bf Mod} & & \\  
\midrule
{\bf Chance} & -- & -- & 50.0 & 50.0 & 50.0 & 50.0 & 50.0 & 50.0 & 50.0 & 50.0 & 50.0 & 50.0 \\ \midrule
{\bf Human} & -- & -- & -- & -- & -- & -- & 53.5 & -- & -- & -- & -- & 71.1 \\ \midrule
{\bf FaceForensics++~\cite{rossler2019faceforensics++}} & -- & -- & 51.3 & 86.3 & 16.2 & 52.7 & -- & 50.0 & 85.7 & 14.3 & 55.3 & 61.9 \\
{\bf Self-consistency\camlast{\textsuperscript{*}}~\cite{huh2018fighting}} & -- & -- & -- & -- & -- & 53.7 & -- & -- & -- & -- & 56.4 & 72.0 \\ \midrule

{\bf \camlast{Low-res no aug.}} & 400 &  & 97.0 & 97.2 & 96.9 & 99.7 & 99.5 & 89.0 & 86.0 & 92.0 & 96.8 & {\bf 98.0} \\
{\bf \camlast{Low-res with aug.}} & 400 & \checkmark & 93.7 & 91.6 & 95.7 & 98.9 & 98.9 & 83.0 & 74.0 & 92.0 & 94.4 & 96.0 \\
{\bf \camlast{High-res with aug.}} & 700 & \checkmark & {\bf 97.1} & 99.8 & 94.5 & {\bf 99.8} & {\bf 100.0} & {\bf 90.0} & 96.0 & 84.0 & {\bf 97.4} & {\bf 98.0} \\
\bottomrule
\end{tabular}
}
\vspace{-4mm}
\end{center}
\caption{{\bf \camready{Real-or-fake} classifier performance.} \camlast{We tested models with FAL warping applied both by automated scripting and a professional artist. 
We observe that training with high-resolution inputs performs the best among the three. 
In addition, training without augmentation performs better in this domain, but adding augmentation makes the model more robust to corruptions, both within and outside of the augmentation set (see Appendix~\ref{sec:apdxrobustness}). *Self-consistency was tested on a 2k random subset of the validation set due to running time constraints.}} 
\label{tab:global}
\vspace{-4mm}
\end{table*}

\subsection{Predicting what moved where}

Upon detecting whether a face has been modified, a natural question for a viewer is \textit{how} the image was edited: which parts of an image were warped, and what did the image look like prior to manipulation? To do this, we predict an optical flow field $\hat{U}\in \mathbb{R}^{H\times W\times 2}$ from the original image $X_{orig} \in \mathbb{R}^{H\times W\times 3}$ to the warped image $X$, which we then use to try to ``reverse" the manipulation and recover the original image.

We train a flow prediction model $\mathcal{F}$ to predict the per-pixel warping field, measuring its distance to an approximate ``ground-truth" flow field $U$ for each training example (computed estimating optical flow between the original and modified images). Fig.~\ref{fig:quallocal_artist} shows examples of these flow fields. To remove erroneous flow values, we discard pixels that fail a forward-backward consistency test, resulting in binary mask $M\in \mathbb{R}^{H\times W\times 1}$. %
\begin{equation}
    \mathcal{L}_{epe}(\mathcal{F}) = ||M \odot \big(\mathcal{F}(X) - U \big)||_2,
    \label{eqn:local}
\end{equation}
where $X$ is a manipulated image, $U$ is its ``ground-truth" flow, $\odot$ is the Hadamard product, and $\mathcal{L}_{epe}$ is a measure of flow error (also known as {\em endpoint error}). We compute this loss for each pixel of the image, and compute the mean. Following~\cite{ummenhofer2017demon}, we encourage the flow to be smooth by minimizing a multiscale loss on the flow gradients:
\begin{equation}
    \mathcal{L}_{ms}(\mathcal{F}) = \sum_{s \in S} \sum_{t \in \{x,y\} } || M \odot \big( \nabla^s_t (\mathcal{F}(X)) - \nabla^s_t (U) \big) ||_2,
    \label{eqn:multiscale}
\end{equation}
where $\nabla^s_x, \nabla^s_y$ are horizontal and vertical gradients of a flow field, decimated by stride $s \in \{2, 8, 32, 64\}$.

\myparagraph{Undoing a warp}
With the correct flow field predicted from the original image to the modified image, one can retrieve the original image by inverse warping.
This leads to a natural reconstruction loss, 
\begin{equation}
    \mathcal{L}_{rec}(\mathcal{F}) = || \mathcal{T} \big( X; \mathcal{F}(X) \big) - X_{orig} ||_1,
    \label{eqn:recons}
\end{equation}
where $\mathcal{T}(X; U)$ warps $X$ by resampling with flow $U$.
In this case, the loss is applied to the \emph{unwarped image} directly, after warping with a differentiable bilinear interpolation layer. 
We note that this approach is similar to the main loss used in flow-based image synthesis models \cite{zhou2016view,xue2017video}.

Applying only the reconstruction loss leads to ambiguities in low-texture regions, which often results in undesirable artifacts. 
Instead, we jointly train with all three losses: $\mathcal{L}_{total} = \lambda_{e}\mathcal{L}_{epe} + \lambda_{m}\mathcal{L}_{ms} + \lambda_{r}\mathcal{L}_{rec}$. We find $\lambda_{e}=1.5$, $\lambda_{m}=15$, and $\lambda_{r}=1$ work well and perform ablations in Section \ref{sec:exp-local}.

\myparagraph{Architecture} We use a Dilated Residual Network variant (DRN-C-26)~\cite{Yu2017}, pretrained on the ImageNet~\cite{russakovsky2015imagenet} dataset, as our base network for local prediction. 
The DRN architecture was designed originally for semantic segmentation, and we found it to work well for the warp prediction task.

We found that directly training the flow regression network performed poorly. We first recast the problem into multinomial classification, commonly used in regression problems (e.g., colorization~\cite{larsson2016learning,zhang2016colorful}, surface normal prediction~\cite{wang2015designing}, and generative modeling~\cite{oord2016pixel}), and then fine-tune with a regression loss. We computed ground truth flow fields using PWC-Net~\cite{sun2018pwc}. Details of the training procedure are provided in Appendix~\ref{sec:apdxtraininfo}.  %

\section{Experiments}
\label{sec:experiments}
We evaluate our ability to detect and undo image manipulations, using both automatic and artist-created images.

\begin{table*}[t]
\centering
\setlength{\tabcolsep}{0.5em} %
\resizebox{1.\linewidth}{!}{
\begin{tabular}{l c@{\hspace{3px}} c@{\hspace{3px}} c@{\hspace{12px}}
c@{\hspace{5px}} c@{\hspace{5px}} c@{\hspace{12px}}
c@{\hspace{5px}} c@{\hspace{5px}} c@{\hspace{12px}}
c@{\hspace{5px}} c@{\hspace{5px}} c@{\hspace{12px}}
c@{\hspace{5px}} c@{\hspace{5px}} c@{\hspace{0px}}}
\toprule
& & & & \multicolumn{6}{c}{\bf Face-Aware Liquify (FAL)} & \multicolumn{6}{c}{\bf Other Manipulations} \\ \cmidrule(lr){5-10} \cmidrule(lr){11-16}
& \multicolumn{3}{c}{\bf Losses} & \multicolumn{3}{c}{\bf Val (Rand-FAL)} & \multicolumn{3}{c}{\bf Artist-FAL} & \multicolumn{3}{c}{\bf Artist-Liquify} & \multicolumn{3}{c}{\bf Portrait-to-Life~\cite{averbuch2017bringing}} \\ \cmidrule(lr){2-4} \cmidrule(lr){5-7} \cmidrule(lr){8-10} \cmidrule(lr){11-13} \cmidrule(lr){14-16}
& \multirow{2}{*}{\bf EPE} & {\bf Multi-} & {\bf Pix} & {\bf EPE} & {\bf IOU-3} & {\bf $\Delta$PSNR} & {\bf EPE} & {\bf IOU-3} & {\bf $\Delta$PSNR} & {\bf EPE} & {\bf IOU-3} & {\bf $\Delta$PSNR} & {\bf EPE} & {\bf IOU-3} & {\bf $\Delta$PSNR} \\
& & {\bf scale} & {\bf $\ell_1$} & $\downarrow$ & $\uparrow$ & $\uparrow$ & $\downarrow$ & $\uparrow$ & $\uparrow$ & $\downarrow$ & $\uparrow$ & $\uparrow$ & $\downarrow$ & $\uparrow$ & $\uparrow$ \\ \midrule
{\bf EPE-only} & $\checkmark$ & & & {\bf 0.51} & {\bf 0.45} & +2.67 & 0.74 & {\bf 0.33} & +2.09 & 0.63 & {\bf 0.12} & -1.21 & {\bf 1.74} & {\bf 0.42} & -- \\ %
{\bf MultiG} & $\checkmark$ & $\checkmark$ & & 0.53 & 0.42 & +2.38 & 0.75 & 0.30 & +2.07 & 0.59 & 0.11 & -0.84 & 1.75 & 0.41 & -- \\ %
\cdashline{1-16}
{\bf Full} & $\checkmark$ & $\checkmark$ & $\checkmark$ & 0.52 & 0.43 & {\bf +2.69} & {\bf 0.73} & 0.28 & {\bf +2.21} & {\bf 0.56} & {\bf 0.12} & {\bf -0.72} & {\bf 1.74} & 0.40 & -- \\ %
\bottomrule
\end{tabular}
\vspace{-2mm}
}
\caption{ {\bf \camready{Warping localization and undoing} performance.} We show performance of our local prediction models across several evaluations: (1) EPE, which measures average flow accuracy, (2) IOU-$3$, which measures flow magnitude prediction accuracy and (3) $\Delta$PSNR, which measures how closely the predicted unwarping recovers the original image from the manipulated; $\uparrow, \downarrow$ indicate if higher or lower is better. Our full method with all losses (flow prediction, multiscale flow gradient, and pixel-wise reconstruction) performs more strongly than ablations, both across datasets which use Face-Aware Liquify and other manipulations. }
\label{tab:local_accuracy}
\vspace{-4mm}
\end{table*}

\subsection{Real-or-fake classification}
We first investigate whether manipulated images can be detected by our global classifier on our validation set. We test the robustness of the classifier by perturbing the images, and measure its generalization ability to manipulations by a professional artist (Table~\ref{tab:global}).

We evaluate several variants: \textbf{(1)} \textbf{\camlast{Low-res with aug.}}: a lower-resolution model (400 pixels on the smaller side), with data augmentation (compression, resizing methods, and photometric changes) and the whole training set (including low-resolution images). \textbf{(2)} \textbf{\camlast{Low-res no aug.}}: We test the augmentation methods above by omitting them. Note that all models still include random flipping and cropping. \textbf{(3)} \textbf{\camlast{High-res with aug.}}: We test if \camlast{training on} higher resolution (700 pixels on shorter side) may allow the network to pick up on more details. We keep the lower resolution images by upsampling them.

\myparagraph{Baselines} We compare our approach to several recent methods, which were trained for other, related forensics tasks. \textbf{(1)} \textbf{FaceForensics++}~\cite{rossler2019faceforensics++}: A network trained on face swapping and reenactment data; we use the Xception~\cite{chollet2017xception} model trained on raw video frames. \textbf{(2)} \textbf{Self-consistency}~\cite{huh2018fighting}: A network trained to spot low-level inconsistencies within an image.

\myparagraph{Evaluations}  First, we evaluate our model's raw accuracy on the binary prediction task. Second, we use ranking-based scores that are not sensitive to the ``base rate" of the fraction of fake images (which may be difficult to know in practice). For this, we use Average Precision (AP), as well as a Two Alternative Force Choice (2AFC) score that is directly comparable to human studies, where we provide our model with two images, one real and one manipulated, and measure the fraction of the time it assigns a higher manipulation probability to the fake.

\myparagraph{Evaluation on auto-generated fakes} We first explore performance on our validation set, shown in Table~\ref{tab:global} (left), containing automatically-generated manipulated images. 

We began by running a human studies test on Amazon Mechanical Turk (AMT). We showed real and manipulated images, side-by-side for 6 seconds, and ask participants to identify the one that was modified. We gave 15 example pairs to ``train'' each person and then collected 35 test examples (for 40 participants total). 
Since the manipulations we trained with are subtle, this was a challenging task; participants were able to identify the manipulated image $53.5\%$ of the time (chance = $50\%$). 
This indicates that it is difficult to use high-level semantics alone for this task. 

\camlast{The low-res model trained with augmentation} performs at $93.7\%$ accuracy and $98.9\%$ average precision.
\camready{Without augmenting for different resampling techniques, our network performance increases to $97.0\%$ accuracy and $99.7\%$ AP, but leaves the network less robust to different image creation and editing pipelines.}
\camlast{Processing at a higher resolution, 700 pixels, the performance also increases to $97.1\%$ accuracy and $99.8\%$ AP.
}
Details of robustness experiments of our models are presented in Appendix~\ref{sec:apdxrobustness}, along with an analysis of the Class Activation Maps of the global classifier in Appendix~\ref{sec:apdxquals}.

\begin{figure*}[t]
    \centering
    \small
    \def\valqualwidth{0.12\linewidth}
    \def\qualheighta{2.25cm}
    \def\qualheightb{1.95cm}
    \def\qualheightc{2.0cm}
    \def\qualheightd{2.15cm}
    \def\qualheighte{2.15cm}

    \resizebox{1\linewidth}{!}{
    \begin{tabular}{*{8}{c@{\hspace{5px}}}}
    \includegraphics[height=\qualheighta]{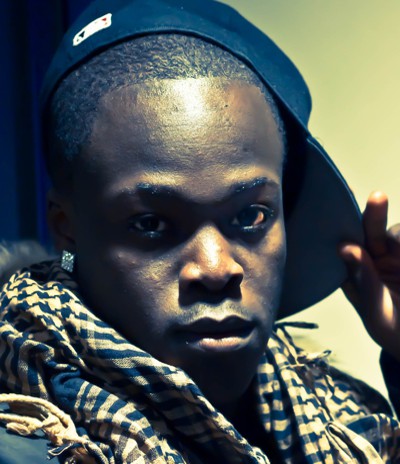} &
    \includegraphics[height=\qualheighta]{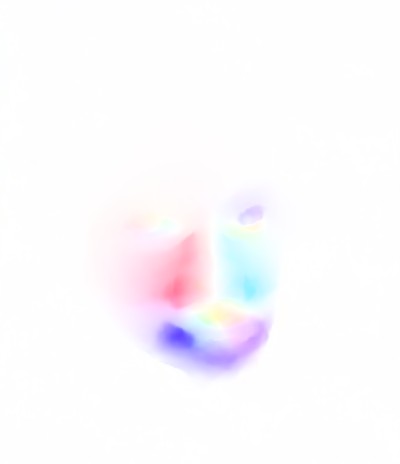} & 
    \includegraphics[height=\qualheighta]{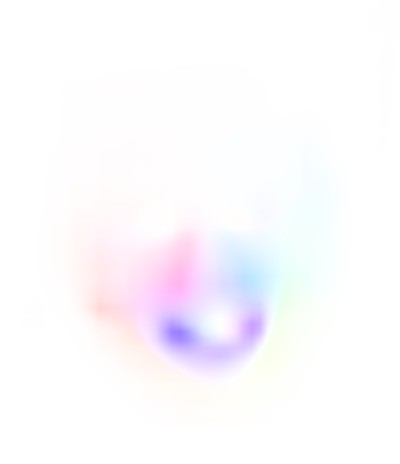} & 
    \includegraphics[height=\qualheighta]{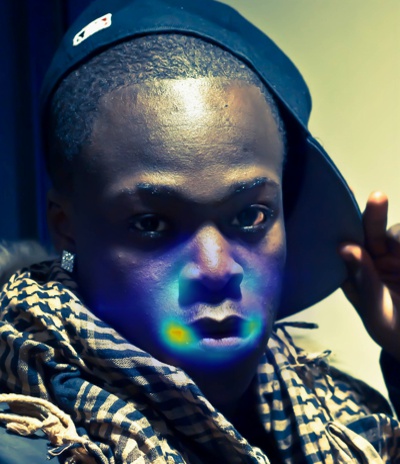} &

    \includegraphics[height=\qualheighta]{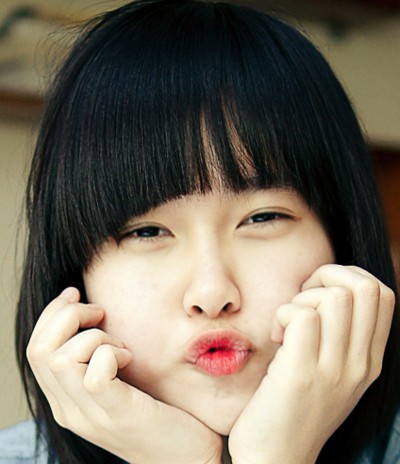} &
    \includegraphics[height=\qualheighta]{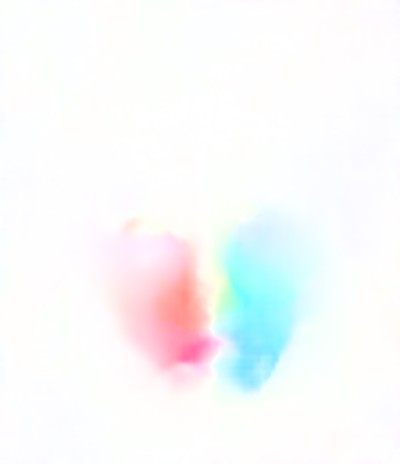} & 
    \includegraphics[height=\qualheighta]{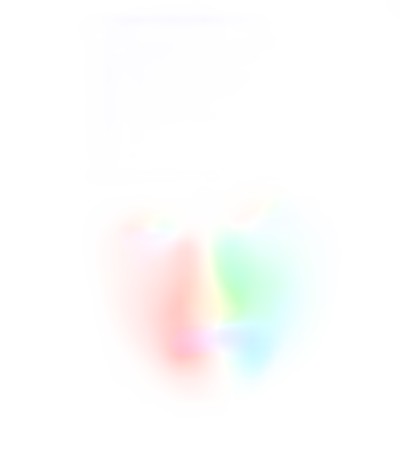} & 
    \includegraphics[height=\qualheighta]{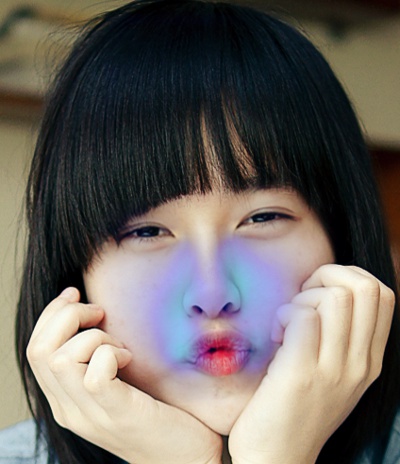} \\

    \includegraphics[height=\qualheightb]{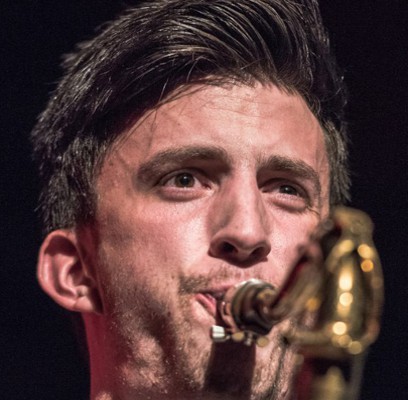} &
    \includegraphics[height=\qualheightb]{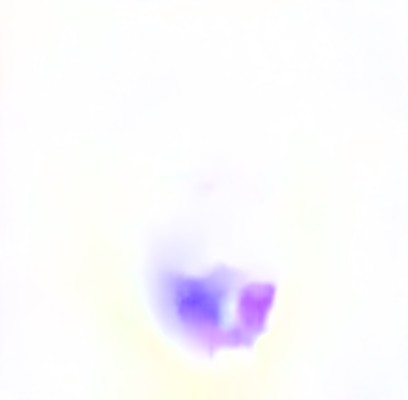} & 
    \includegraphics[height=\qualheightb]{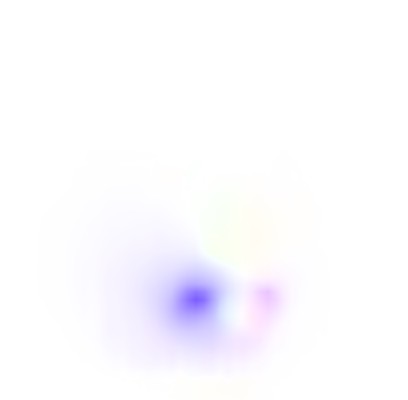} &
    \includegraphics[height=\qualheightb]{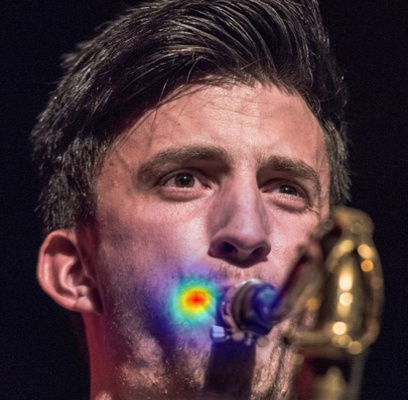} &

    \includegraphics[height=\qualheightb]{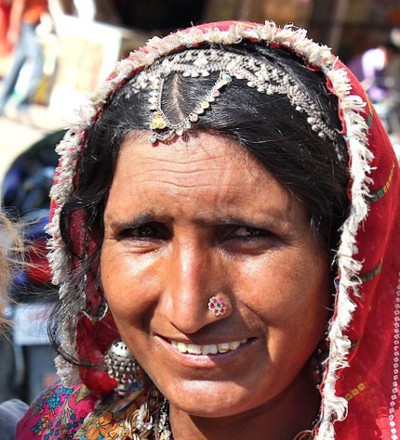} &
    \includegraphics[height=\qualheightb]{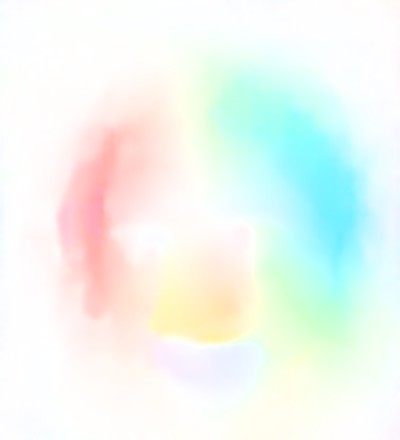} & 
    \includegraphics[height=\qualheightb]{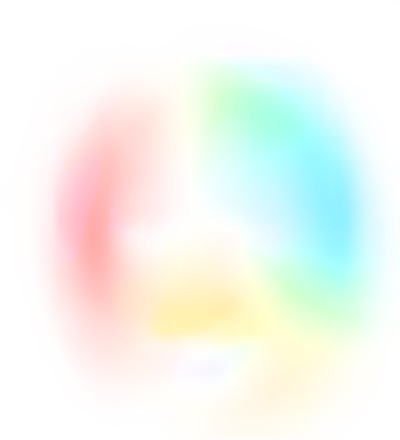} &
    \includegraphics[height=\qualheightb]{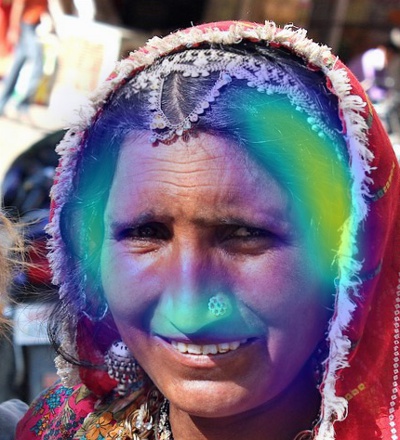} \\

    \includegraphics[height=\qualheightc]{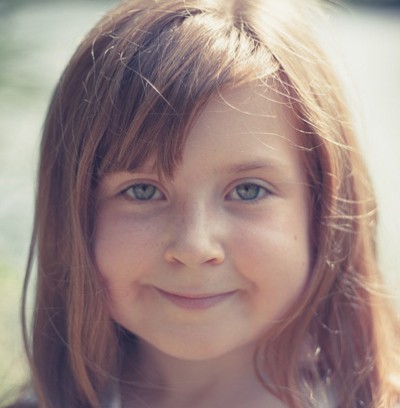} &
    \includegraphics[height=\qualheightc]{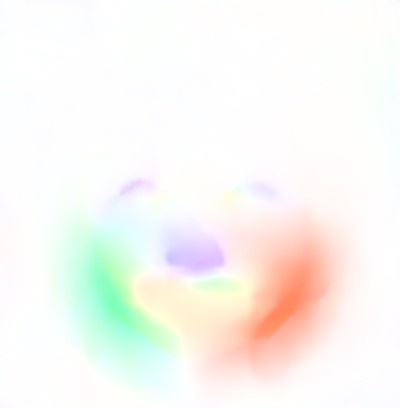} &
    \includegraphics[height=\qualheightc]{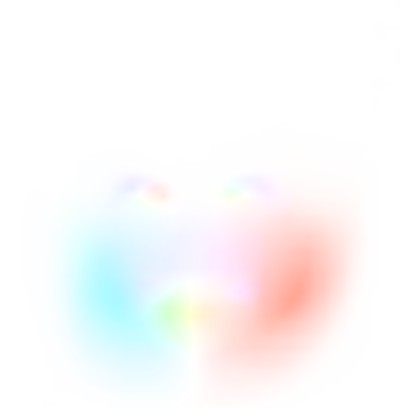} &
    \includegraphics[height=\qualheightc]{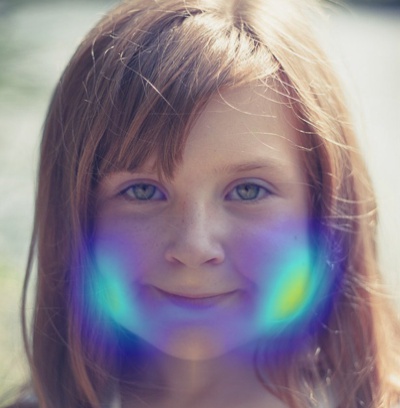} &

    \includegraphics[height=\qualheightc]{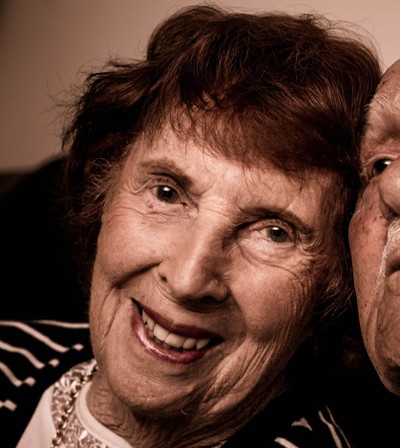} &
    \includegraphics[height=\qualheightc]{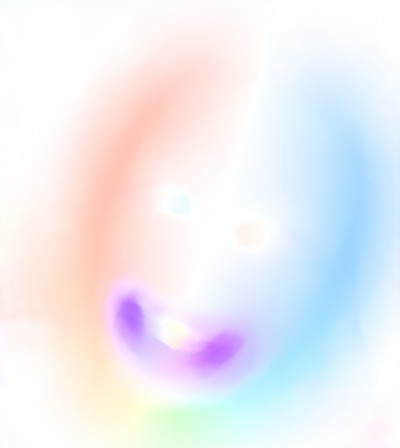} &
    \includegraphics[height=\qualheightc]{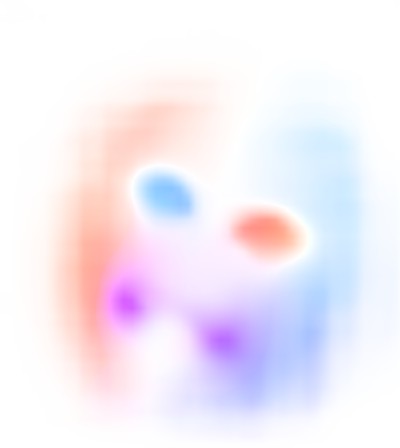} &
    \includegraphics[height=\qualheightc]{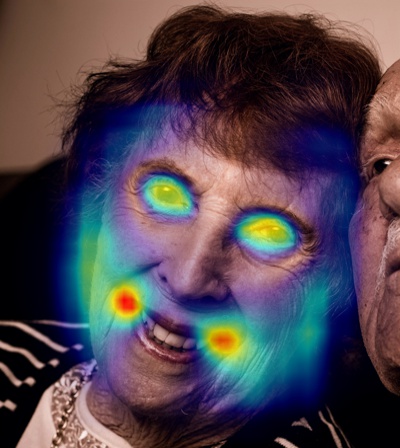} \\
    
    \includegraphics[height=\qualheightd]{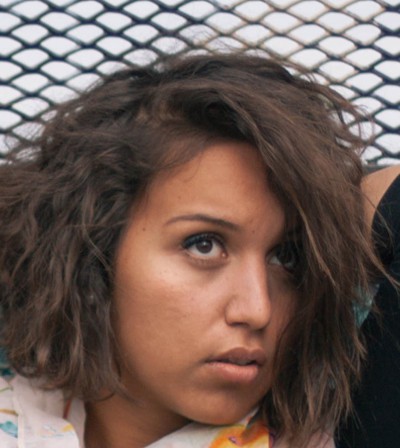} &
    \includegraphics[height=\qualheightd]{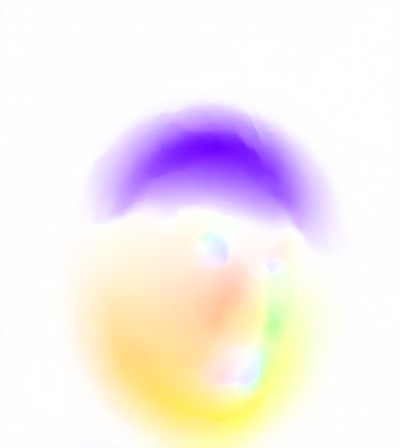} &
    \includegraphics[height=\qualheightd]{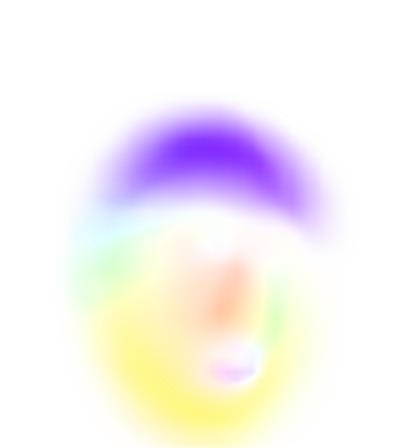} &
    \includegraphics[height=\qualheightd]{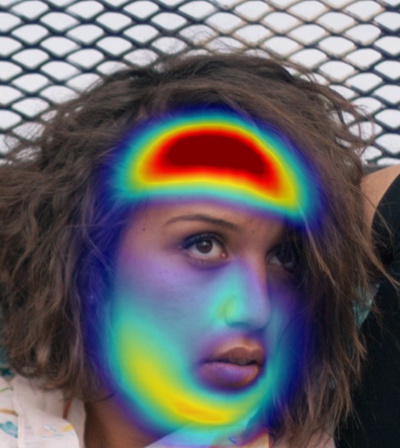} &

    \includegraphics[height=\qualheightd]{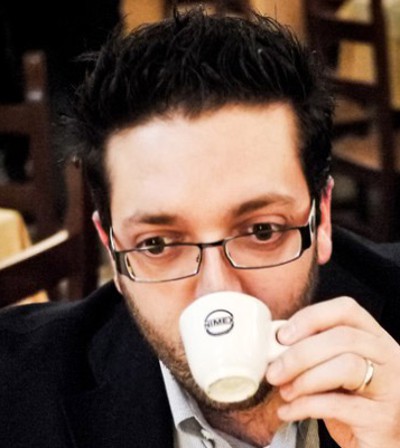} &
    \includegraphics[height=\qualheightd]{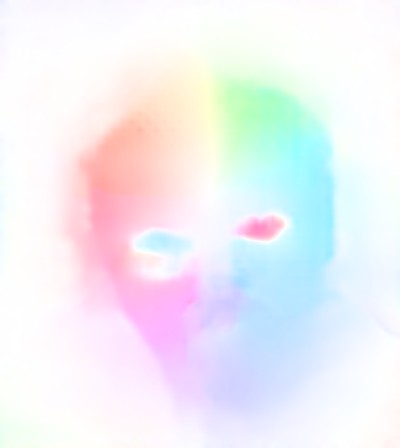} &
    \includegraphics[height=\qualheightd]{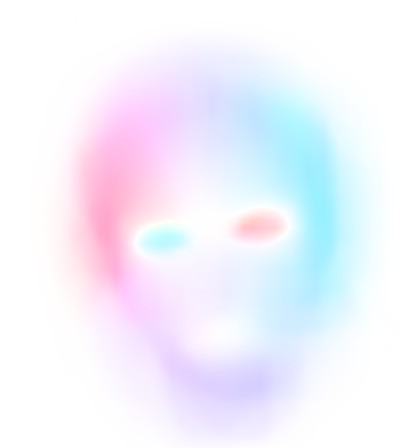} &
    \includegraphics[height=\qualheightd]{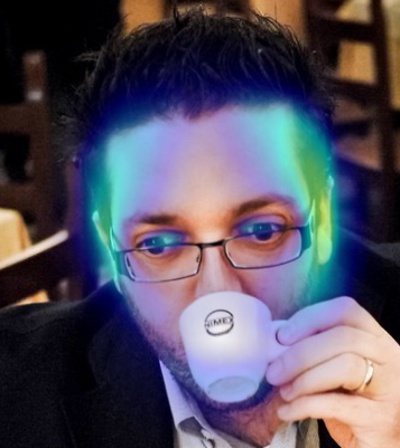} \\
    
    \includegraphics[height=\qualheighte]{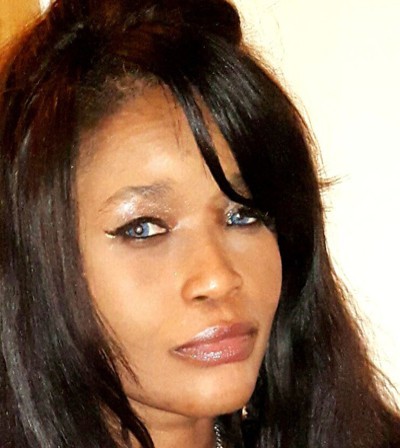} &
    \includegraphics[height=\qualheighte]{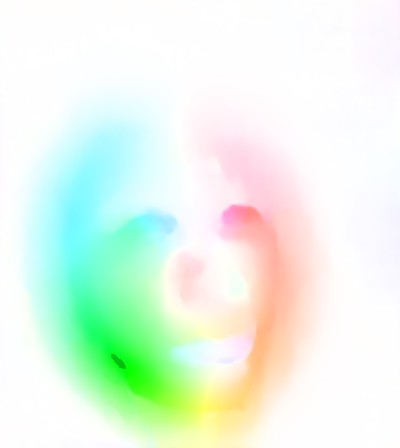} &
    \includegraphics[height=\qualheighte]{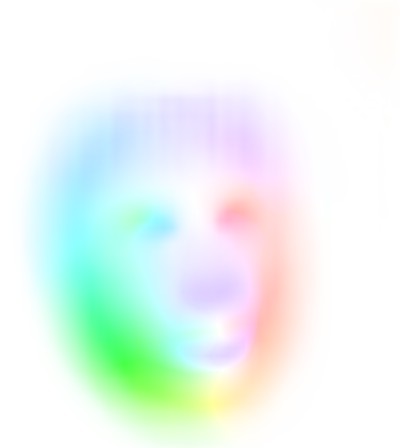} &
    \includegraphics[height=\qualheighte]{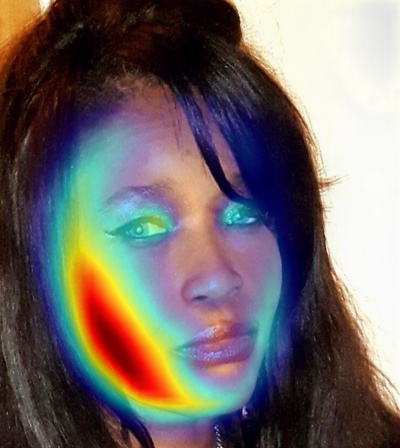} &

    \includegraphics[height=\qualheighte]{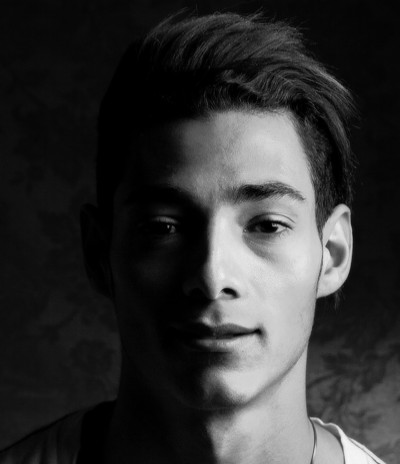} &
    \includegraphics[height=\qualheighte]{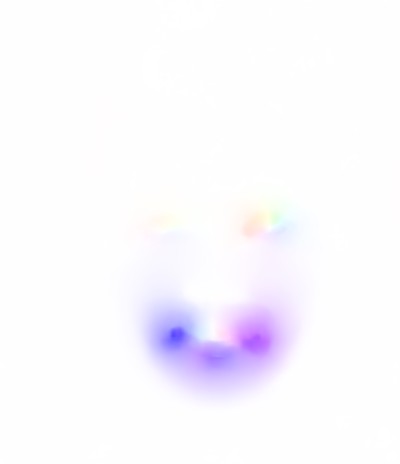} &
    \includegraphics[height=\qualheighte]{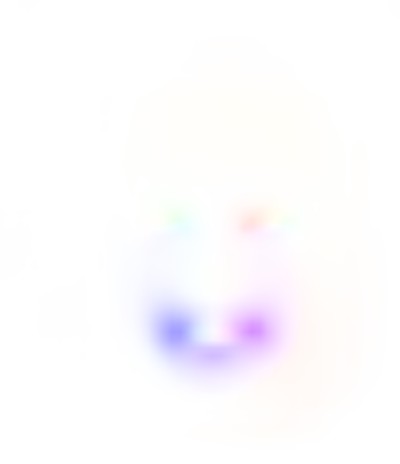} &
    \includegraphics[height=\qualheighte]{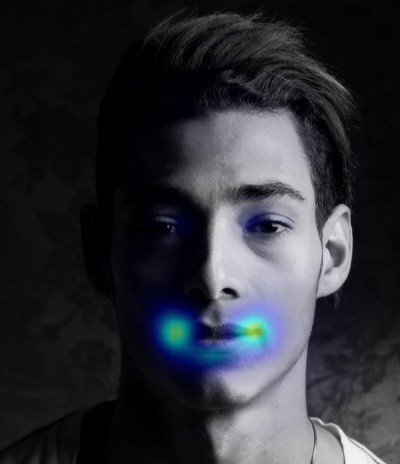} \\
    
    Input & GT flow & Our prediction & Flow overlay & Input & GT flow & Our prediction & Flow overlay \\
    \end{tabular}}
    \vspace{-.1in}
    \caption{\textbf{\camready{Qualitative results on artist-created and auto-generated data.}} We show examples of our flow prediction on images manipulated from an external artist and from our \camready{auto-generated} validation set. {\bf (Input)} Input manipulated image. {\bf (GT flow)} The "ground truth" optical flow from original to manipulated image. {\bf (Our prediction)} Predicted flow from our network. {\bf (Flow overlay)} Magnitude of predicted flow overlaid.
    \supp{See Appendix~\ref{sec:apdxquals} for additional examples.}}
    \label{fig:quallocal_artist}
\end{figure*}

\begin{figure*}[t]
    \centering
    \small
    \newcommand\myspy[4]{\spy [white,
		spy connection path={\draw[thick, white] (tikzspyonnode) -- (tikzspyinnode);}
		every spy on node/.append style={thick},
		every spy in node/.append style={thick}] on ({#1},{#2}) in node [left] at ({#3},{#4})}
    \def\valqualwidth{2.5cm}
    \resizebox{1.0\linewidth}{!}{
    \begin{tabular}{*{6}{c@{\hspace{2px}}}}
    \begin{tikzpicture}[spy using outlines={circle, magnification=2.5, size=1.5cm, connect spies},inner xsep=0cm]	
        \node (O) [] {
        	\includegraphics[height=\valqualwidth]{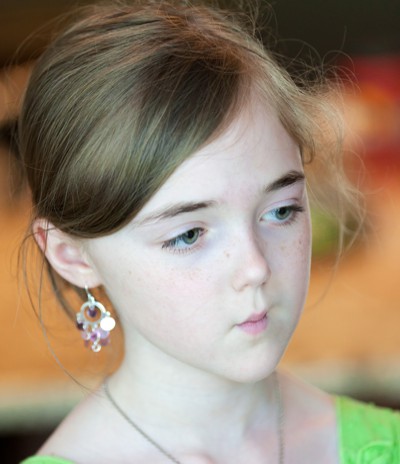}
        };
        \myspy{0.3}{-0.5}{0}{0.7};
    \end{tikzpicture} &
        \begin{tikzpicture}[spy using outlines={circle, magnification=2.5, size=1.5cm, connect spies},inner xsep=0cm]	
        \node (O) [] {
        	\includegraphics[height=\valqualwidth]{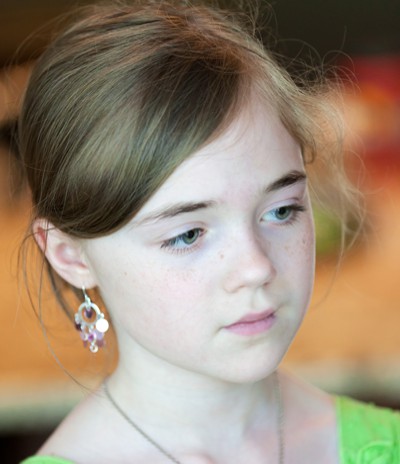}
        };
        \myspy{0.3}{-0.5}{0}{0.7};
    \end{tikzpicture} &
    \begin{tikzpicture}[spy using outlines={circle, magnification=2.5, size=1.5cm, connect spies},inner xsep=0cm]	
        \node (O) [] {
        	\includegraphics[height=\valqualwidth]{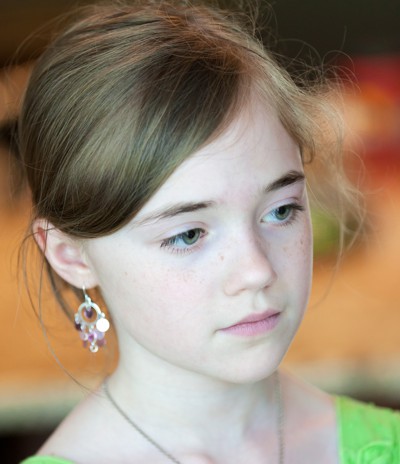}
        };
        \myspy{0.3}{-0.5}{0}{0.7};
    \end{tikzpicture} &
    \begin{tikzpicture}[spy using outlines={circle, magnification=2.5, size=1.5cm, connect spies},inner xsep=0cm]	
        \node (O) [] {
        	\includegraphics[height=\valqualwidth]{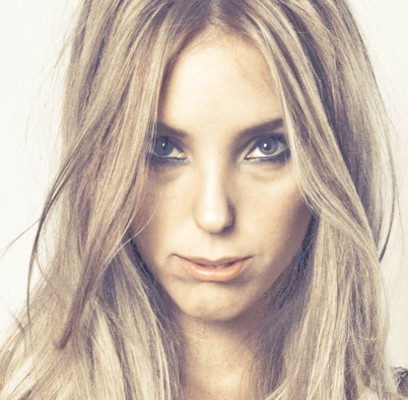}
        };
        \myspy{0.1}{-0.1}{0}{0.7};
    \end{tikzpicture} &
    \begin{tikzpicture}[spy using outlines={circle, magnification=2.5, size=1.5cm, connect spies},inner xsep=0cm]	
        \node (O) [] {
        	\includegraphics[height=\valqualwidth]{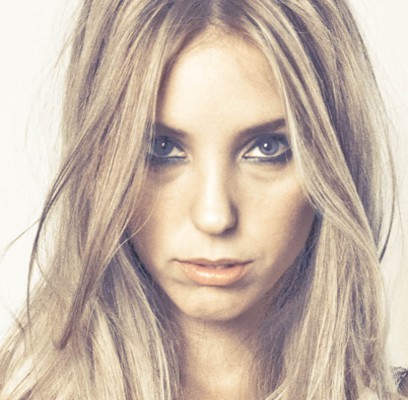}
        };
        \myspy{0.1}{-0.1}{0}{0.7};
    \end{tikzpicture} &
    \begin{tikzpicture}[spy using outlines={circle, magnification=2.5, size=1.5cm, connect spies},inner xsep=0cm]	
        \node (O) [] {
        	\includegraphics[height=\valqualwidth]{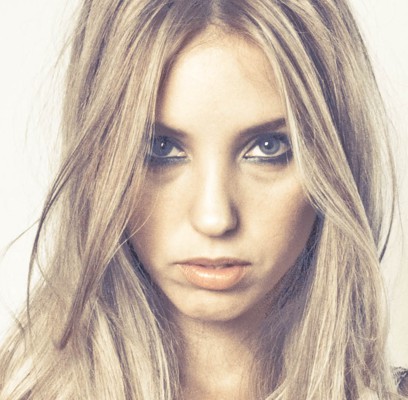}
        };
        \myspy{0.1}{-0.1}{0}{0.7};
    \end{tikzpicture} \\
        \begin{tikzpicture}[spy using outlines={circle, magnification=2.5, size=1.5cm, connect spies},inner xsep=0cm]	
        \node (O) [] {
        	\includegraphics[height=\valqualwidth]{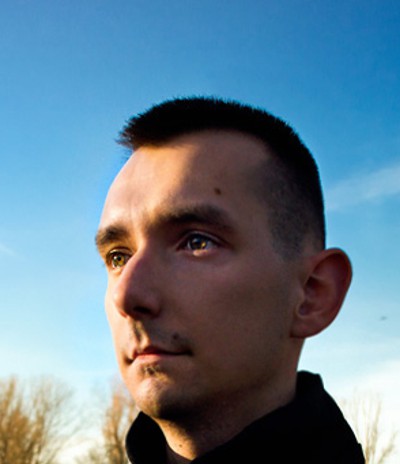}
        };
        \myspy{-0.28}{-0.3}{0}{0.7};
    \end{tikzpicture} &
        \begin{tikzpicture}[spy using outlines={circle, magnification=2.5, size=1.5cm, connect spies},inner xsep=0cm]	
        \node (O) [] {
        	\includegraphics[height=\valqualwidth]{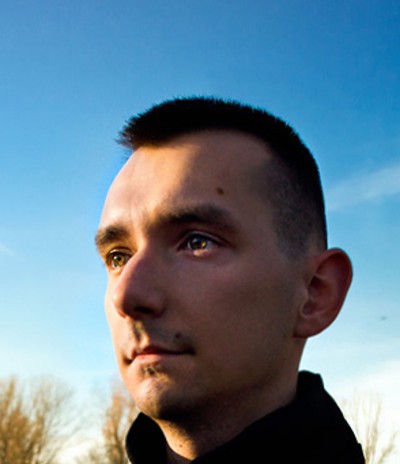}
        };
        \myspy{-0.28}{-0.3}{0}{0.7};
    \end{tikzpicture} &
    \begin{tikzpicture}[spy using outlines={circle, magnification=2.5, size=1.5cm, connect spies},inner xsep=0cm]	
        \node (O) [] {
        	\includegraphics[height=\valqualwidth]{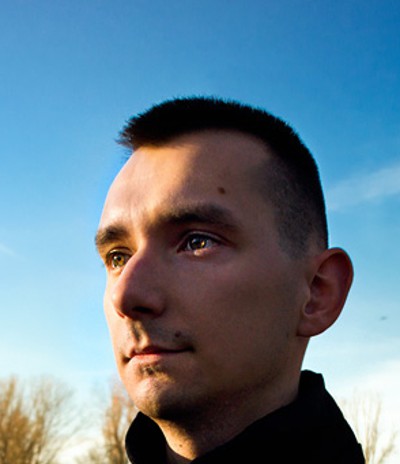}
        };
        \myspy{-0.28}{-0.3}{0}{0.7};
    \end{tikzpicture} &
    \begin{tikzpicture}[spy using outlines={circle, magnification=4.5, size=1.5cm, connect spies},inner xsep=0cm]	
        \node (O) [] {
        	\includegraphics[height=\valqualwidth]{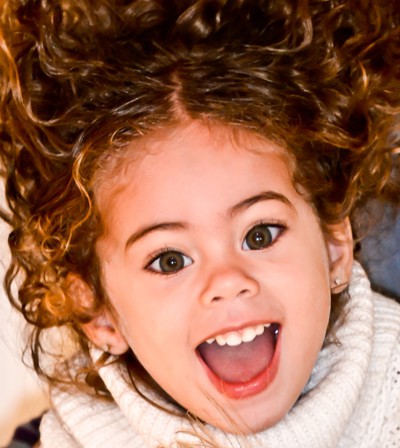}
        };
        \myspy{0.31}{-0.05}{0}{0.7};
    \end{tikzpicture} &
    \begin{tikzpicture}[spy using outlines={circle, magnification=4.5, size=1.5cm, connect spies},inner xsep=0cm]	
        \node (O) [] {
        	\includegraphics[height=\valqualwidth]{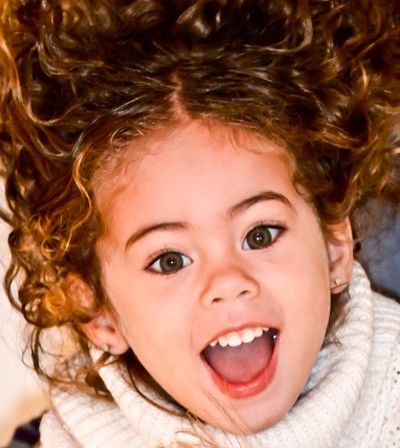}
        };
        \myspy{0.31}{-0.05}{0}{0.7};
    \end{tikzpicture} &
    \begin{tikzpicture}[spy using outlines={circle, magnification=4.5, size=1.5cm, connect spies},inner xsep=0cm]	
        \node (O) [] {
        	\includegraphics[height=\valqualwidth]{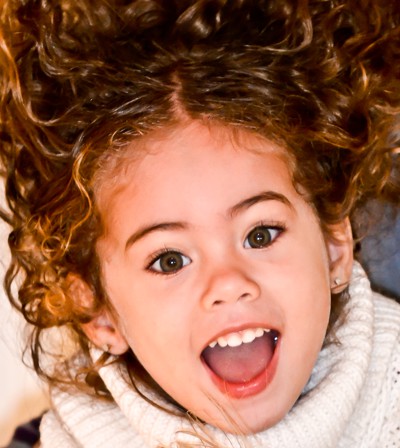}
        };
        \myspy{0.31}{-0.05}{0}{0.7};
    \end{tikzpicture} \\

    {\small Manipulated} & 
    {\small Unwarped} & 
    {\small Original} & 
    {\small Manipulated} & 
    {\small Unwarped} & 
    {\small Original}\\
    \end{tabular}
    }
    \vspace{-.1in}
    \caption{\textbf{Unwarping results.} These images show results from the artist edited test dataset, where the manipulations are reversed by our model. Among other edits, the mouth and nose in the top row were expanded. In the bottom row, the nose shape was made less round and the eye was shrunk.}
    \label{fig:unwarping}
\end{figure*}

\myparagraph{Artist test set} Critically, we investigate if training on our random perturbations generalizes to a more real-world setting. We collect data from a professional artist, tasked with the goal of making a subject more attractive, or changing the subject's expression. 
Since the edits here are made to be more noticeable, and study participants were able to identify the modified image with $71.1\%$ accuracy. Our \camlast{high-res} classifier achieves $98.0\%$ in the 2AFC setting. Our accuracy drops from $97.1$ in the validation setting to $90.0$. However, the AP drops much less, from $99.8$ to $97.4$. This indicates that there is some domain gap between our random perturbations and an artist, which can be reduced by a certain extent by ``recalibrating'' the classifier's detection threshold. 

\myparagraph{Baselines} We compare to two recent baselines for image forensics, FaceForensics++~\cite{rossler2019faceforensics++} and Self-consistency~\cite{huh2018fighting}.
Neither of these methods are designed for our application: FaceForensics++ is split into three manipulation types: face swapping,  ``deepfakes'' face replacement, and face2face reenactment~\cite{rossler2019faceforensics++}.
Self-consistency, on the other hand, is designed to detect low-level differences in image characteristics. Both methods perform around chance on our dataset, indicating that generalizing to facial warping manipulations is challenging.

However, our method is able to generalize to some of the FaceForensics++ datasets. \camlast{The low-res model with augmentation performs significantly better than chance ($50.0\%$ acc; $50.0\%$ AP) on FaceSwap ($65.4\%$ acc; $71.8\%$ AP), Face2Face ($69.9\%$ acc; $77.4\%$ AP) and DeepFake ($77.2\%$ acc; $87.1\%$ AP) tasks. On the other hand, the high-res model doesn't generalize as well to the task: FaceSwap ($59.4\%$ acc; $64.7\%$ AP), Face2Face ($55.7\%$ acc; $55.9\%$ AP) and DeepFake ($65.0\%$ acc; $71.3\%$ AP).} %
This indicates that \camlast{training with lower resolution images might allow the model to learn more high-level features (e.g., geometric inconsistencies), where the features can then be used to detect other face manipulations, while training with high-resolution images allows the model to leverage low-level image features that allow it to perform better within the narrower domain. Moreover,} training on synthetically generated subtle facial warping data could be an interesting technique to generalize to other, more complex, editing tasks. 

\subsection{Localizing and undoing manipulations}
\label{sec:exp-local}

\noindent Next, we evaluate manipulation localization and reversal.

\myparagraph{Model variations}  To help understand what parts of our model contributed to its performance, we ablate the loss functions for our local prediction model. Since previous methods have not considered the problem of predicting or reversing warps, we consider variations of our own model. {\bf (1) Our full method}: trained with endpoint error (EPE) (Eqn.~\ref{eqn:local}), multiscale gradient (Eqn.~\ref{eqn:multiscale}), and reconstruction (Eqn.~\ref{eqn:recons}) losses. 
{\bf (2) EPE}: an ablation only trained with endpoint loss.
{\bf (3) MultiG}: trained with endpoint and multiscale, but without reconstruction loss.

\myparagraph{Evaluations} 
We evaluate our model in several ways, capturing both localization ability and warp reversal ability. \textbf{(1) End Point Error (EPE)} similarity between predicted and ground-truth flow between the original and manipulated images (Eqn~\ref{eqn:local}).
\textbf{(2) Intersection Over Union (IOU-$\tau$)} We apply threshold $\tau$ to predicted and ground truth flows magnitudes and compute IOU. \textbf{(3) Delta Peak Signal-to-Noise Ratio ($\Delta$PSNR)} effectiveness of our predicted unwarping, PSNR (original, unwarped manipulated) minus PSNR (original, manipulated)

\begin{figure}
    \centering
    \def\portraitwidth{0.24\linewidth}
    \begin{tabular}{*{4}{c@{\hspace{1px}}}}
    \includegraphics[width=\portraitwidth]{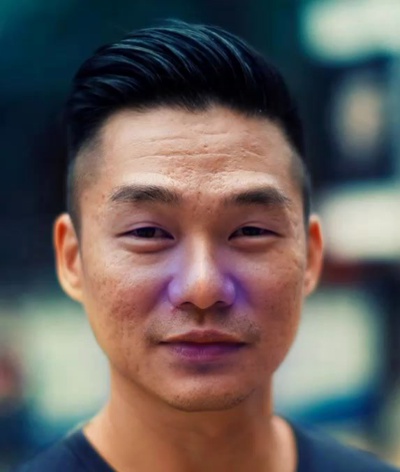} &
    \includegraphics[width=\portraitwidth]{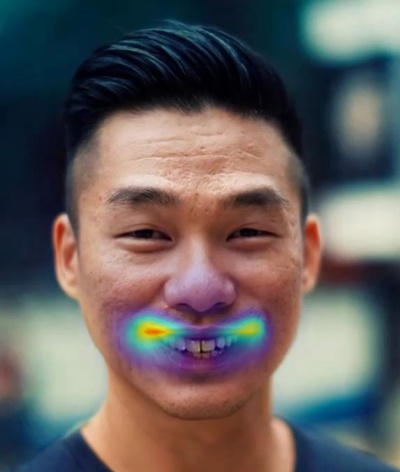} &
    \includegraphics[width=\portraitwidth]{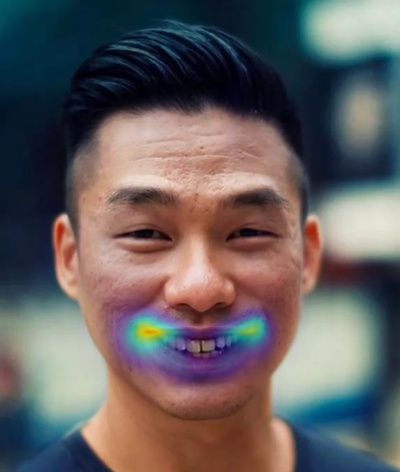} &
    \includegraphics[width=\portraitwidth]{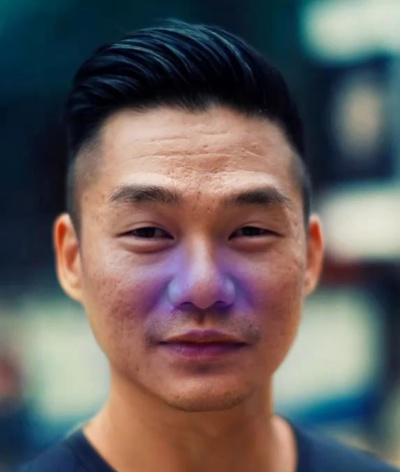} \\
    \multicolumn{4}{c}{
    \includegraphics[width=.90\linewidth,height=.34\linewidth]{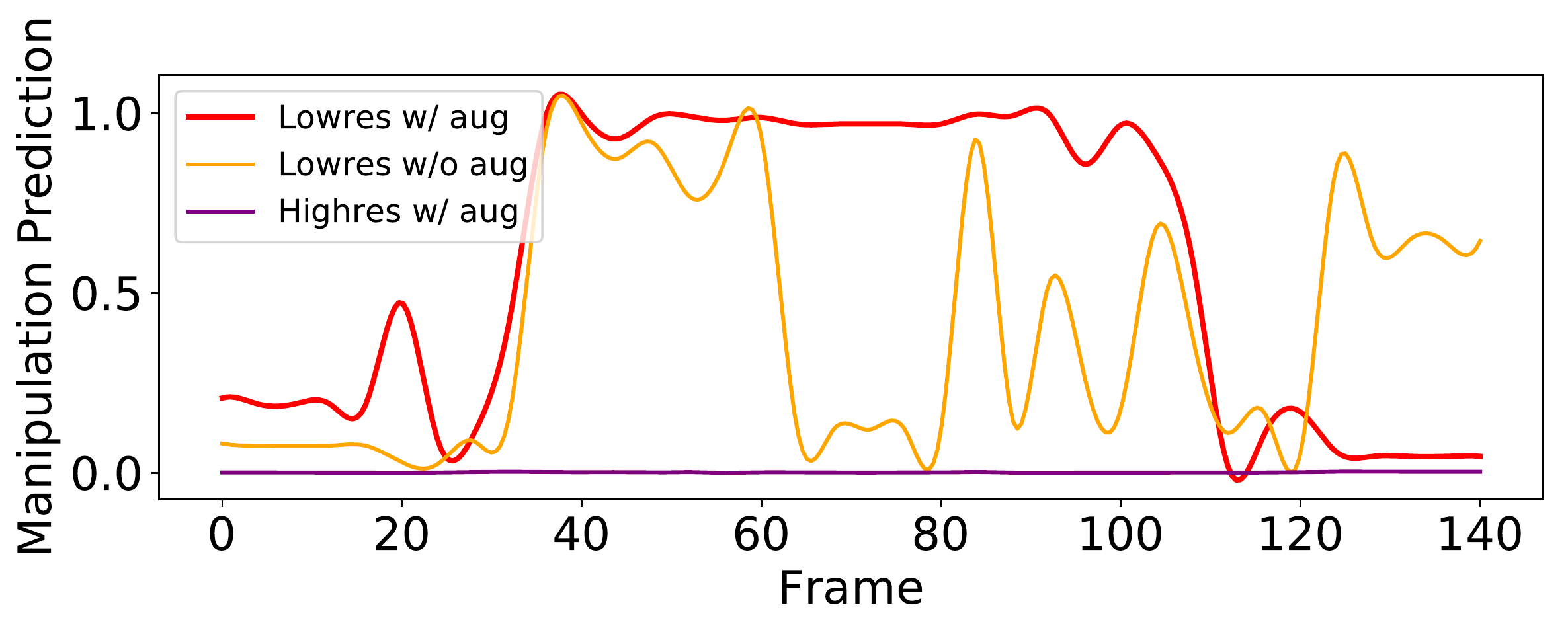}
    }\\
    \end{tabular}
    \vspace{-4mm}
    \caption{
    {\bf Analysis of a Puppeteered  video~\cite{averbuch2017bringing}}. A single input image (first and last frame) is driven by a smiling animation.  {\bf (top)} Our local analysis correctly identifies manipulated regions (corners of the mouth). {\bf (bottom)} The global prediction over time shows how the animation moves from input to smile and back.
    }
    \vspace{-4mm}
    \label{fig:puppeteering}
\end{figure}

\myparagraph{Analysis}
As shown in Table~\ref{tab:local_accuracy}, we found that removing a loss reduced performance.  In particular, we found that directly optimizing the reconstruction loss led to better image reconstructions.
In Figures~\ref{fig:quallocal_artist} and~\ref{fig:unwarping}, we show several qualitative results on the automatically-generated and artist-created data.
\supp{We include more qualitative results, randomly sampled from the validation set, in Appendix~\ref{sec:apdxquals}.}

\subsection{Out-of-distribution manipulations}
\camlast{While our model is trained to detect face warping manipulations made by Photoshop, we also evaluate its ability to detect other kinds of image editing, and discuss its limitations.}

\myparagraph{Puppeteering} We conduct an experiment to see whether our method can be used to detect the results of recent image-puppeteering work~\cite{averbuch2017bringing}.
In this work, a video (from a different subject) is used to animate an input image via image warping and the additional of extra details, such as skin wrinkles and texture in the eyes and mouth. 
We apply our manipulation detection model to this data, and show that despite not being trained on this data, we are still able to make reasonable predictions.
Fig.~\ref{fig:puppeteering} shows a qualitative result of running both local and global predictors on this data, where it correctly identifies a puppeted smile animation that starts and returns to a (real) rest pose.
\camlast{We observe that our low-res model with augmentation produces more stable predictions over time than the one trained without augmentation. Moreover, the high-res model doesn't generalize to detecting such manipulations.}
We note that PSNR comparisons on this data are not possible, due to the addition of non-warping image details. 

\begin{table}[h]
\centering
\resizebox{0.92\linewidth}{!}{
    \begin{tabular}{@{}lccccc@{}}
    
    \toprule & \multicolumn{2}{c}{\bf Global} && \multicolumn{2}{c}{\bf Local}\\
    
    \cmidrule{2-3} \cmidrule{5-6} & {\bf Accuracy} & {\bf AP} && {\bf $\Delta$PSNR} & {\bf EPE}\\ \midrule

    {\bf \camlast{High-res with aug.}} & 55.0 & 64.0 && -- & --\\
    {\bf Low-res no aug.} & 57.0 & 67.7 && +0.15 & 0.99\\
    {\bf \camlast{Low-res with aug.}} & {\bf 67.0} & {\bf 79.6} && {\bf +0.61} & {\bf 0.91} \\
    
    \bottomrule
    \end{tabular}
}
\caption{{\bf Results on Facebook \camready{post-processing}}. We tested our global and local models with the artist test set and compare the performance of \camlast{our different models}.}
\label{tab:social}
\vspace{-2mm}
\end{table}

\myparagraph{Social media \camready{post-processing} pipeline}
\label{sec:fb-compression}
\camready{
We also evaluated our model's robustness to post-processing operations performed by Facebook (e.g., extra JPEG compression). We uploaded our artist-created fakes to Facebook, and then evaluated our method with the post-processed images. Table~\ref{tab:social} shows results of our \camlast{low-res} models trained with and without augmentation, \camlast{along with the high-res global classifier}. We note that \camlast{the high-res model doesn't generalize to such scenario, and} both global and local models trained with augmentation perform better in this scenario. %
}

\begin{figure}[t]
    \centering
    \small
    \newcommand\myspy[4]{\spy [white,
		spy connection path={\draw[thick, white] (tikzspyonnode) -- (tikzspyinnode);}
		every spy on node/.append style={thick},
		every spy in node/.append style={thick}] on ({#1},{#2}) in node [left] at ({#3},{#4})}
    \def\valqualheighte{2.7cm}
    \resizebox{1.0\linewidth}{!}{
    \begin{tabular}{*{3}{c@{\hspace{1.5px}}}}
    
    \begin{tikzpicture}[spy using outlines={circle, magnification=3.5, size=1.5cm, connect spies},inner xsep=0cm]	
        \node (O) [] {
        	\includegraphics[height=\valqualheighte,clip,trim=0 0 0 120]{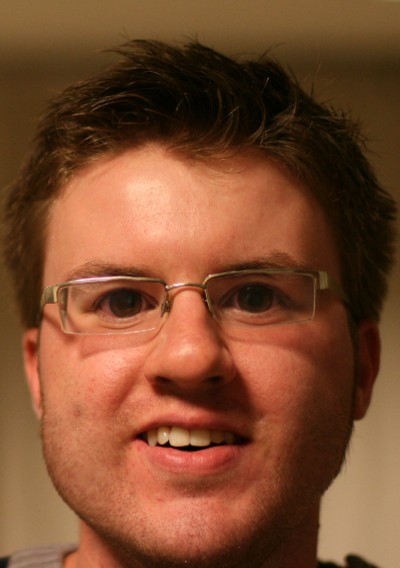}
        };
        \myspy{-0.5}{0.25}{1.6}{0.7};
    \end{tikzpicture} &
    
     \begin{tikzpicture}[spy using outlines={circle, magnification=3.5, size=1.5cm, connect spies},inner xsep=0cm]	
        \node (O) [] {
        	\includegraphics[height=\valqualheighte]{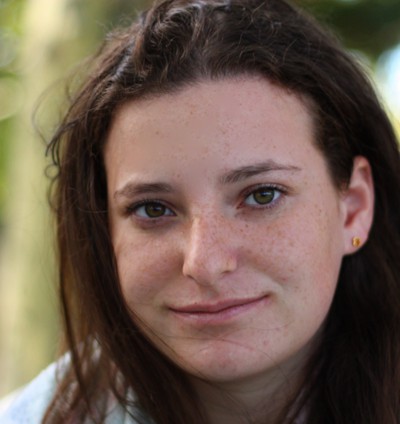}
        };
        \myspy{0.3}{-0.6}{1.6}{0.7};
    \end{tikzpicture} &
    
    \begin{tikzpicture}[spy using outlines={circle, magnification=4.0, size=1.5cm, connect spies},inner xsep=0cm]	
        \node (O) [] {
        	\includegraphics[height=\valqualheighte]{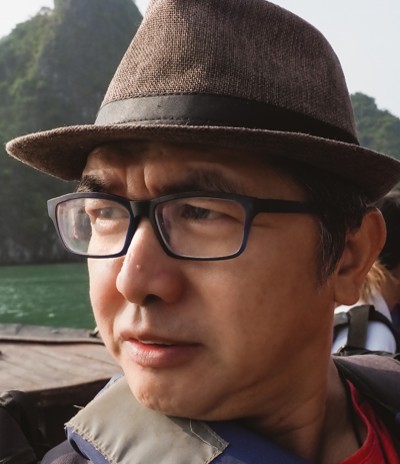}
        };
        \myspy{-0.25}{-0.35}{1.6}{0.7};
    \end{tikzpicture} \\

    \begin{tikzpicture}[spy using outlines={circle, magnification=3.5, size=1.5cm, connect spies},inner xsep=0cm]	
        \node (O) [] {
        	\includegraphics[height=\valqualheighte,clip,trim=0 0 0 120]{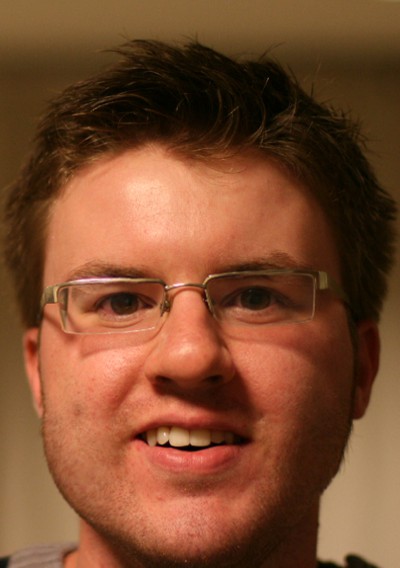}
        };
        \myspy{-0.5}{0.25}{1.6}{0.7};
    \end{tikzpicture} &

    \begin{tikzpicture}[spy using outlines={circle, magnification=3.5, size=1.5cm, connect spies},inner xsep=0cm]	
        \node (O) [] {
        	\includegraphics[height=\valqualheighte]{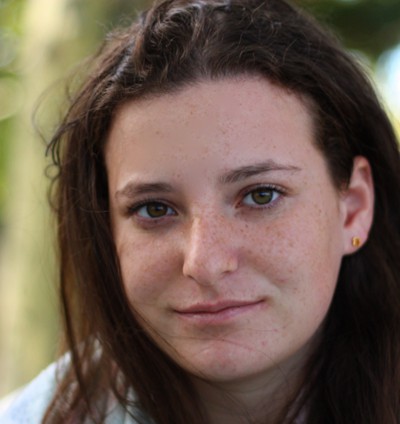}
        };
        \myspy{0.3}{-0.6}{1.6}{0.7};
    \end{tikzpicture} &
    
    \begin{tikzpicture}[spy using outlines={circle, magnification=4.0, size=1.5cm, connect spies},inner xsep=0cm]	
        \node (O) [] {
        	\includegraphics[height=\valqualheighte]{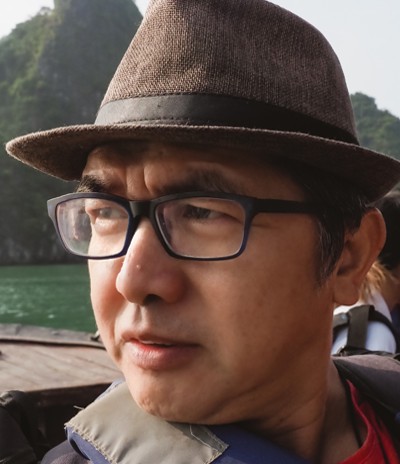}
        };
        \myspy{-0.25}{-0.35}{1.6}{0.7};
    \end{tikzpicture} \\

    \begin{tikzpicture}[spy using outlines={circle, magnification=3.5, size=1.5cm, connect spies},inner xsep=0cm]	
        \node (O) [] {
        	\includegraphics[height=\valqualheighte,clip,trim=0 0 0 120]{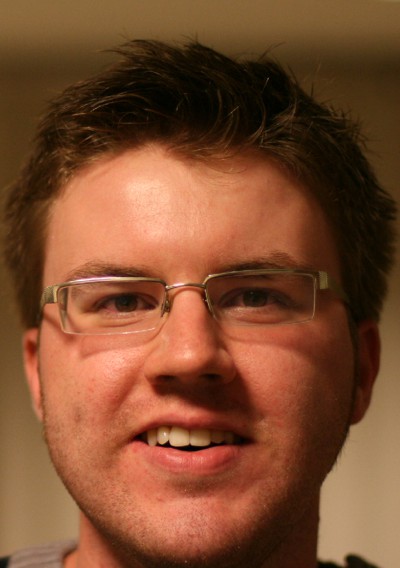}
        };
        \myspy{-0.5}{0.25}{1.6}{0.7};
    \end{tikzpicture} &

    \begin{tikzpicture}[spy using outlines={circle, magnification=3.5, size=1.5cm, connect spies},inner xsep=0cm]	
        \node (O) [] {
        	\includegraphics[height=\valqualheighte]{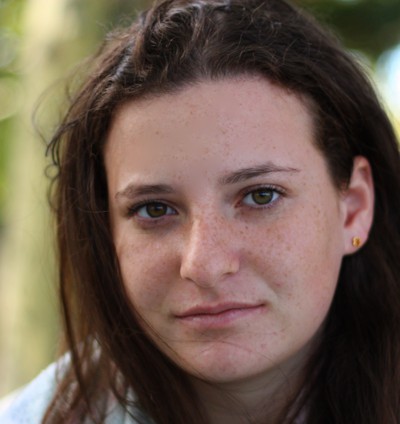}
        };
        \myspy{0.3}{-0.6}{1.6}{0.7};
    \end{tikzpicture} &

    \begin{tikzpicture}[spy using outlines={circle, magnification=4.0, size=1.5cm, connect spies},inner xsep=0cm]	
        \node (O) [] {
        	\includegraphics[height=\valqualheighte]{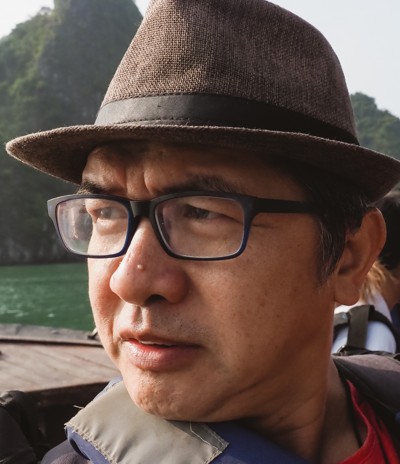}
        };
        \myspy{-0.25}{-0.35}{1.6}{0.7};
    \end{tikzpicture} \\

    \hspace*{-0.5cm} 
    \includegraphics[height=\valqualheighte,clip,trim=0 0 0 120]{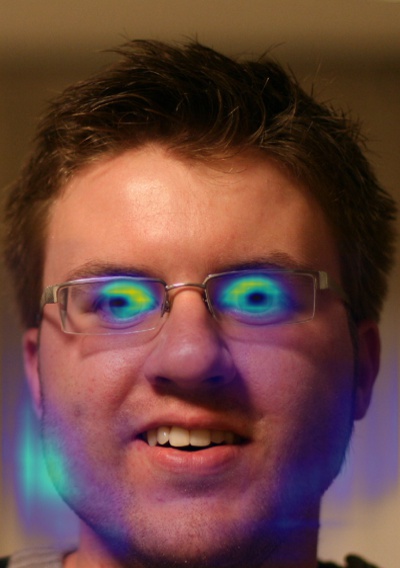} &
    
    \hspace*{-0.5cm} 
    \includegraphics[height=\valqualheighte]{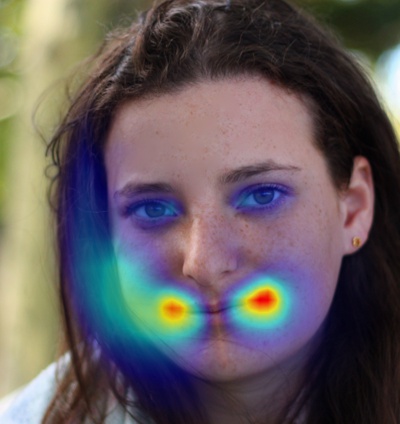} &
    
    \hspace*{-0.5cm} 
    \includegraphics[height=\valqualheighte]{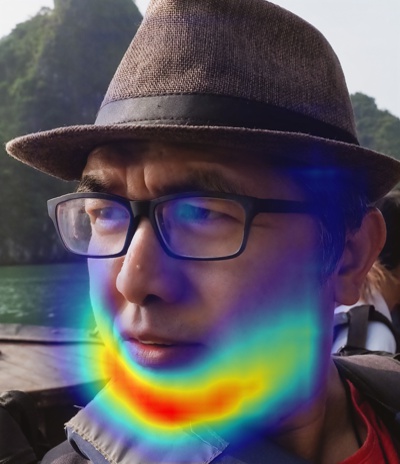} \\

    \hspace*{-0.5cm} 
    {\small Lens Studio} & 
    \hspace*{-0.5cm} 
    {\small Facetune} &
    \hspace*{-0.5cm} 
    {\small Facetune (airbrushed)} \\

    \end{tabular}
    }
    \vspace{-.1in}
    \caption{\textbf{Unwarping with other image editing tools.} We show results of unwarping predictions from Snapchat Lens Studio and Facetune edits. From top to bottom is: \textbf{(1)} Manipulated input. \textbf{(2)} Suggested "undo". \textbf{(3)} Original image. \textbf{(4)} Heatmap overlay. }
    \label{fig:othertool}
    \vspace{-4mm}
\end{figure}

\myparagraph{Other image editing tools}
\camready{
We also tested our local detection model on facial warping by Facetune~\cite{facetune} and Snapchat Lens Studio~\cite{snapchat}. Facetune provides similar warping operations to change a person's expression along with an airbrushing functionality, and Snapchat Lens Studio warps the face by magnifying certain parts of a face. Fig.~\ref{fig:othertool} shows a qualitative result of suggested undo predictions. Notice that our model is able to perform reasonable recovery of the edits even if the model is not trained on these tools.
}

\begin{figure}[t]
    \centering
    \begin{tabular}{*{3}{c@{\hspace{1px}}}}
    \includegraphics[width=.32\linewidth,clip,trim=0 20 0 50]{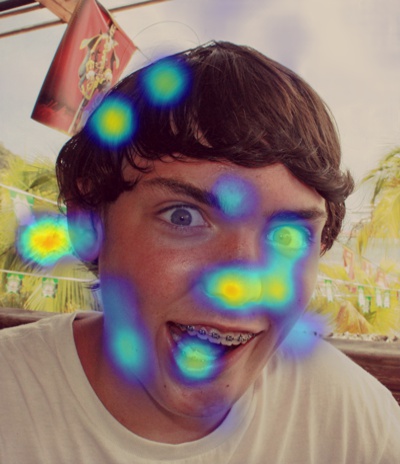} & 
    \includegraphics[width=.32\linewidth,clip,trim=0 20 0 50]{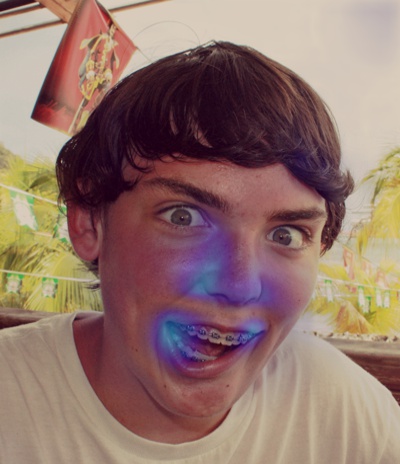} & 
    \includegraphics[width=.32\linewidth,clip,trim=0 20 0 50]{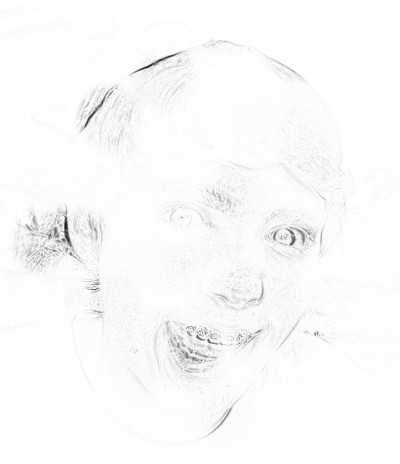} \\
    GT edits & Predicted edits & Unwarped diff \\
    \end{tabular}
    \vspace{-2mm}
    \caption{\textbf{Limitations}. When manipulations are too far outside the training distribution, as with the general Liquify tool experiment. Our local prediction model fails to correctly identify warped regions. This is visible in the overlay as well as in the unwarped image (difference to ground truth after unwarping is shown on the right, darker is worse). }
    \label{fig:limitations}
\end{figure}

\myparagraph{Generic Liquify filter}
Like any data-driven method, we are limited by our training distribution. 
Warping edits that exist outside of this, such as warping applied to hair or body, cannot be detected by our method. This can be seen in our artist experiment with the generic (non-face) Liquify filter, where images are sometimes outside the distribution (Figure~\ref{fig:limitations}). 
Despite this, our method can still predict with success well above chance (64.0 accuracy, 85.6 AP), indicating some generalization. However, the global classifier performs well below the FAL operation (90.0 accuracy, 97.4 AP), and the local prediction accuracy is not enough to improve the PSNR when unwarping (-0.72 $\Delta$PSNR). 
Increasing the range of scripted warping operations is likely to improve this.
In general, reversing the warps is a challenging problem, as there are many configurations of plausible faces. This can be seen in that the PSNR improvement we get on the artist test set is limited to +2.21 db on average. While this manipulation is reduced, the problem of perfectly restoring the original image remains an open challenge.

\section{Conclusion}

We have presented the first method designed to detect facial warping manipulations, and did so using by training a forensics model entirely with images automatically generated from an image editing tool. We showed that our model can outperform human judgments in determining whether images are manipulated, and in many cases is able to predict the local deformation field used to generate the warped images. We see facial warp detection as an important step toward making forensics methods for analyzing images of a human body, and extending these approaches to body manipulations and photometric edits such as skin smoothing are interesting avenues for future work. Moreover, we also see our work as being a step toward toward making forensics tools that learn without labeled data, and which incorporate interactive editing tools into the training process.  

\section*{Acknowledgements}

We thank Daichi Ito and Adam Pintek for contributing to our artist test set, along with Hany Farid, Matthias Kirchner, and Minyoung Huh for the helpful discussions. This work was supported, in part, by DARPA MediFor and UC Berkeley Center for Long-Term Cybersecurity. The views, opinions and/or findings expressed are those of the authors and should not be interpreted as representing the official views or policies of the Department of Defense or the U.S. Government. 

{\small
\bibliographystyle{ieee_fullname}
\bibliography{fal}
}

\renewcommand{\thesection}{A\arabic{section}}
\renewcommand{\thefigure}{A\arabic{figure}}
\setcounter{section}{0}
\setcounter{figure}{0}

\clearpage
\noindent{\Large\bf Appendix}

\section{Supplemental Video}
\camready{
We have included a supplementary video in the following link: \href{https://youtu.be/TUootD36Xm0}{https://youtu.be/TUootD36Xm0}. We invite readers to view this video for better visualizations of our qualitative results.
}

\section{Qualitative results}
\label{sec:apdxquals}

\begin{figure}
    \centering
    \begin{tabular}{{c@{\hspace{2px}}}}
    \includegraphics[width=\linewidth]{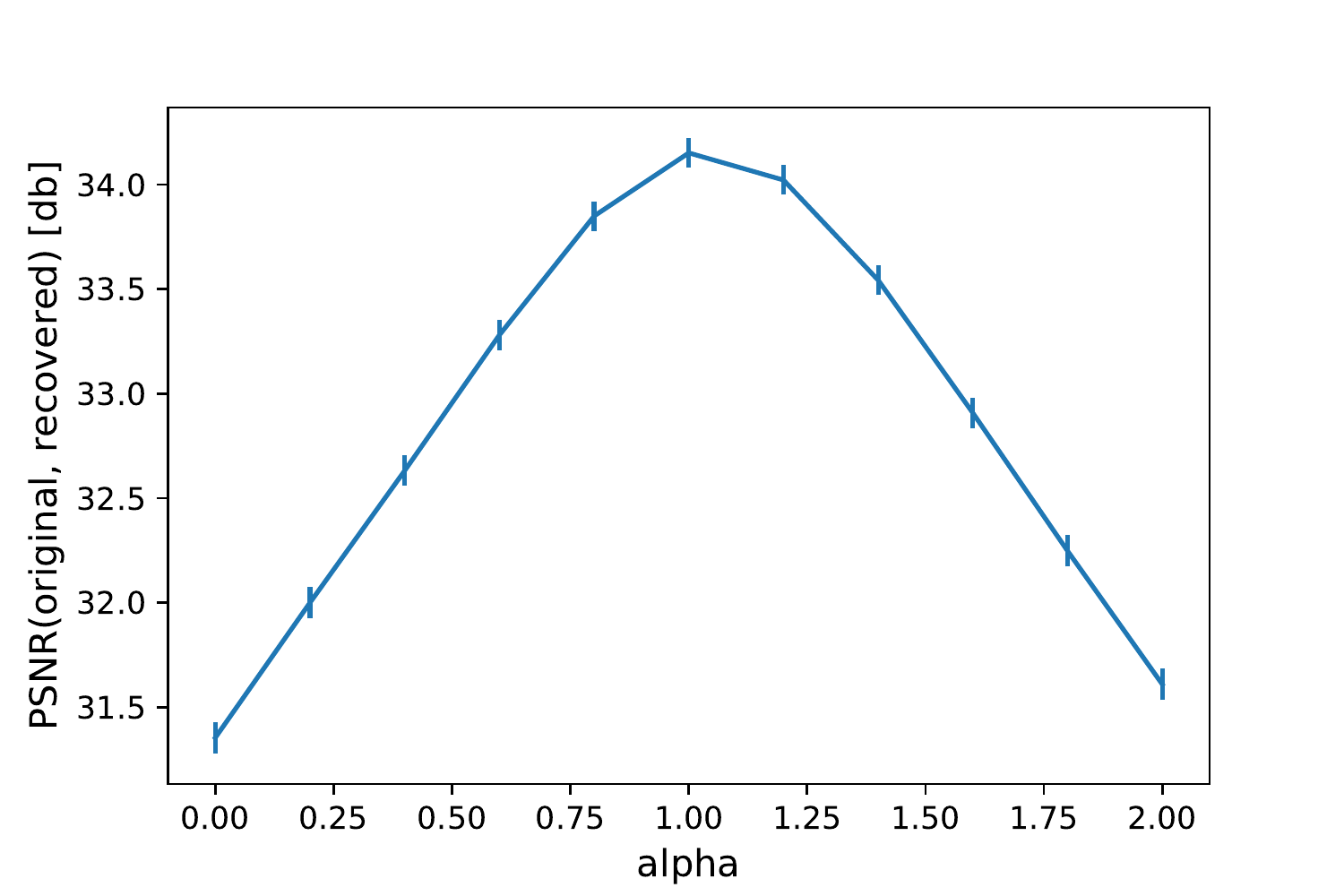}
    \end{tabular}
    \vspace{-7px}
    \caption{PSNR plots from our held-out validation subset. We plot the average PSNRs of the unwarped image to the original (y-axis), with respect to a multiplicative factor on the predicted flow field. The error bars are the standard errors. In the ideal case, this PSNR should peak at 1.0, the predicted flow.}
    \vspace{-12px}
    \label{fig:psnr}
\end{figure}

\begin{figure*}[t]
    \centering
    \footnotesize
    \def\vallinewidth{.13\linewidth}
    \begin{tabular}{*{7}{c@{\hspace{2px}}}}
    \includegraphics[width=\vallinewidth]{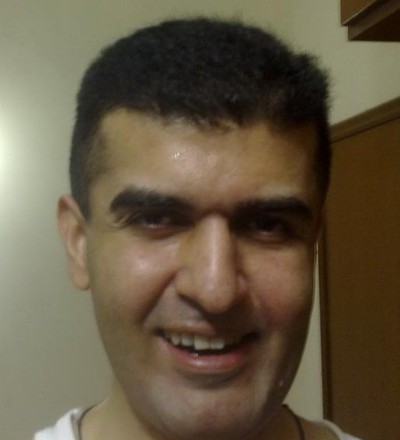} &
    \includegraphics[width=\vallinewidth]{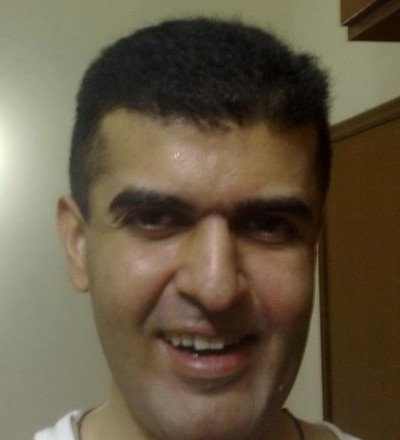} &
    \includegraphics[width=\vallinewidth]{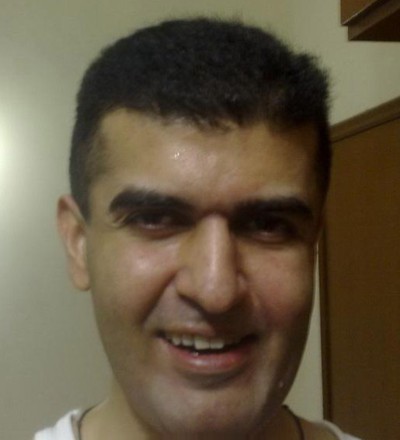} &
    \includegraphics[width=\vallinewidth]{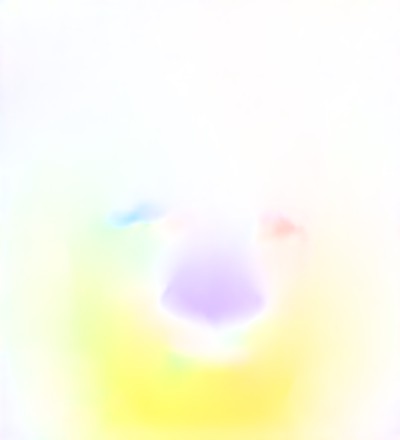} &
    \includegraphics[width=\vallinewidth]{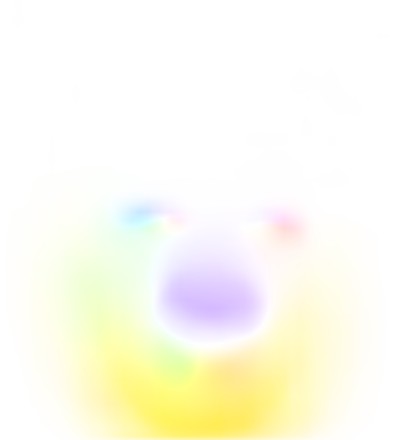} &
    \includegraphics[width=\vallinewidth]{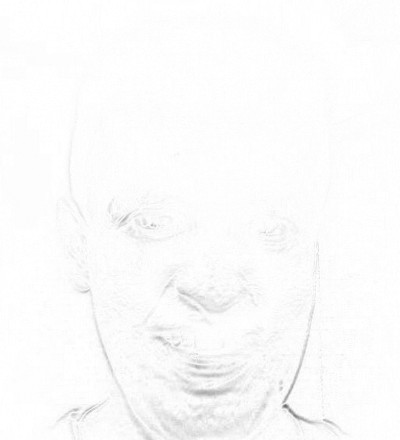} &
    \includegraphics[width=\vallinewidth]{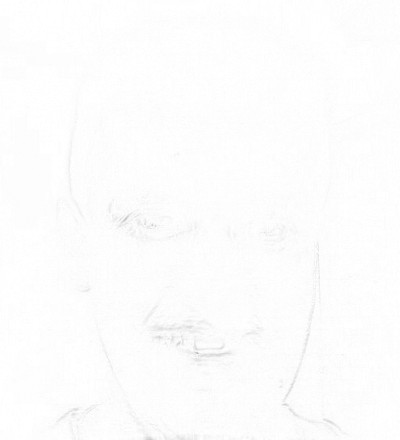} \\
    
    \includegraphics[width=\vallinewidth]{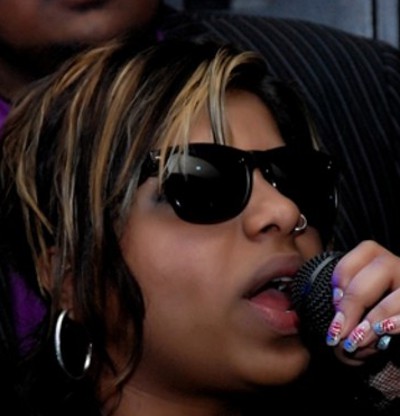} &
    \includegraphics[width=\vallinewidth]{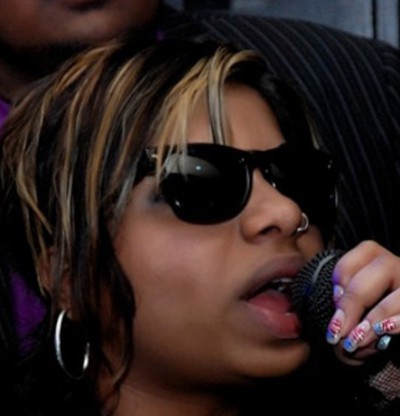} &
    \includegraphics[width=\vallinewidth]{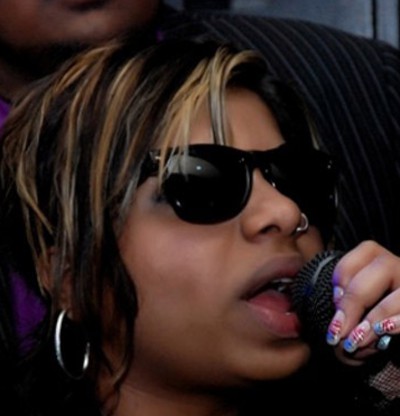} &
    \includegraphics[width=\vallinewidth]{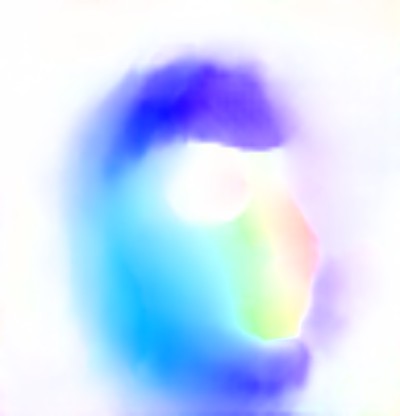} &
    \includegraphics[width=\vallinewidth]{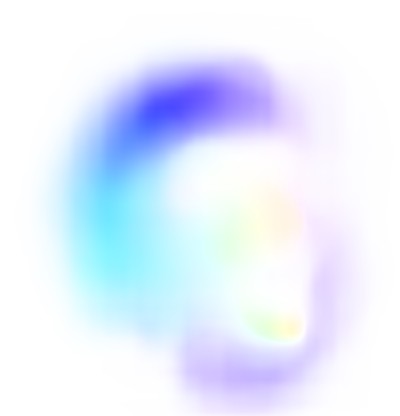} &
    \includegraphics[width=\vallinewidth]{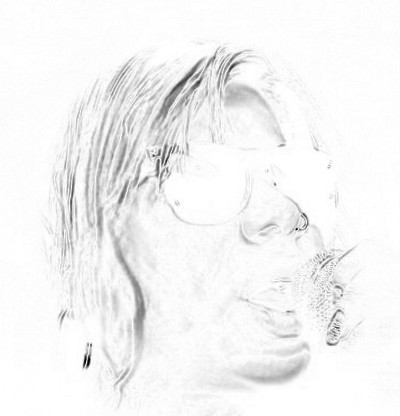} &
    \includegraphics[width=\vallinewidth]{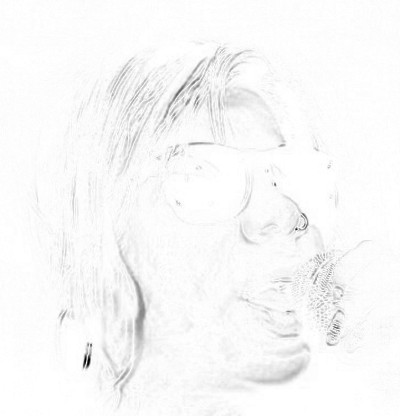} \\
    
    \includegraphics[width=\vallinewidth]{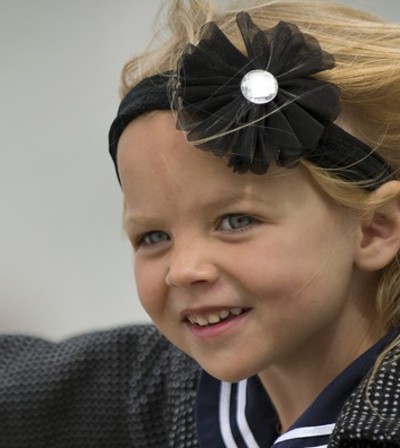} &
    \includegraphics[width=\vallinewidth]{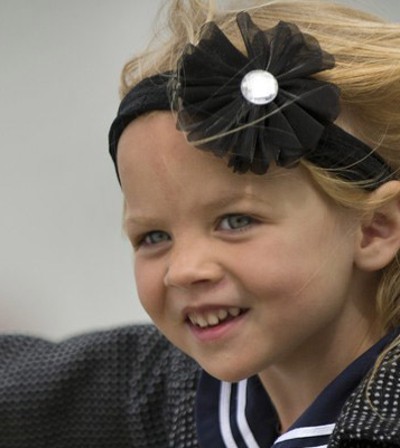} &
    \includegraphics[width=\vallinewidth]{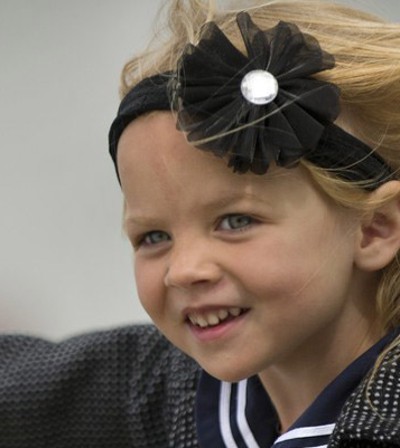} &
    \includegraphics[width=\vallinewidth]{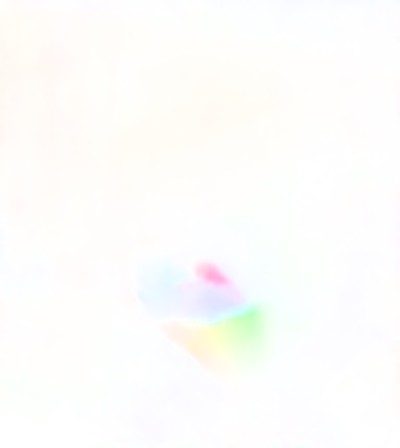} &
    \includegraphics[width=\vallinewidth]{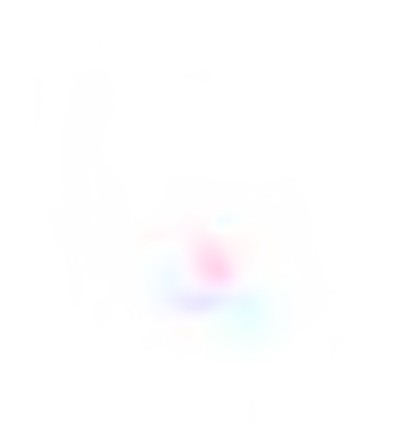} &
    \includegraphics[width=\vallinewidth]{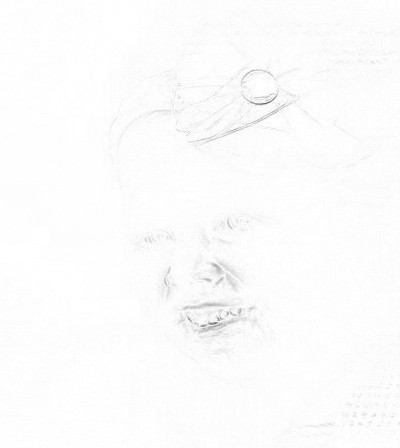} &
    \includegraphics[width=\vallinewidth]{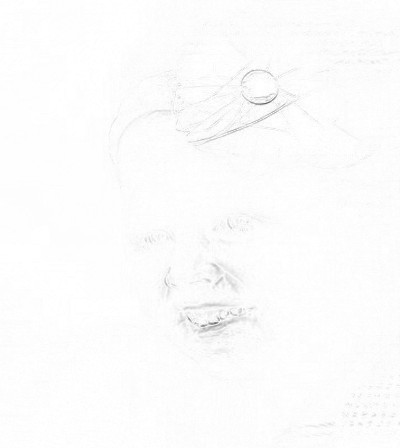} \\
    
    \includegraphics[width=\vallinewidth]{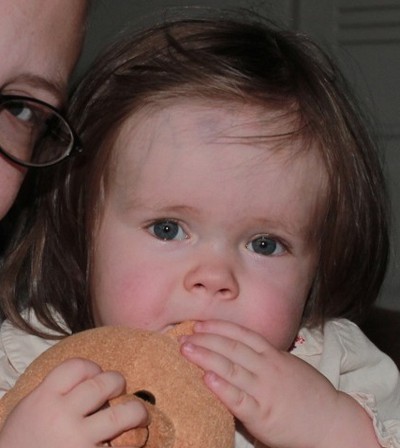} &
    \includegraphics[width=\vallinewidth]{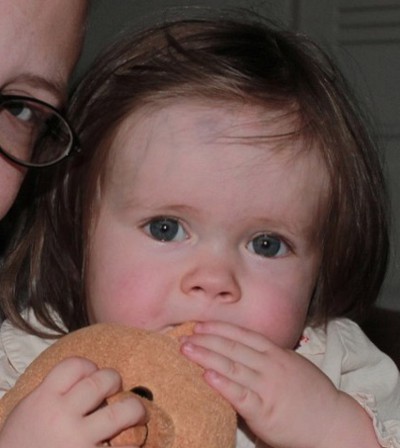} &
    \includegraphics[width=\vallinewidth]{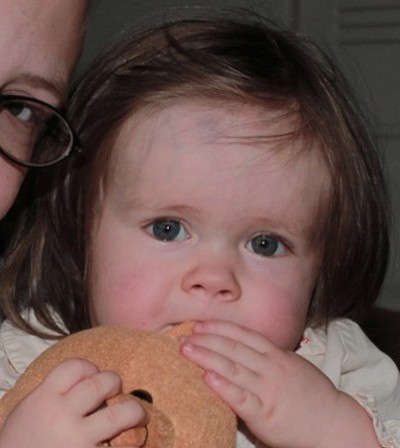} &
    \includegraphics[width=\vallinewidth]{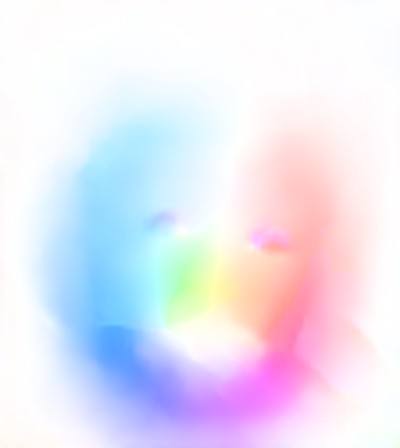} &
    \includegraphics[width=\vallinewidth]{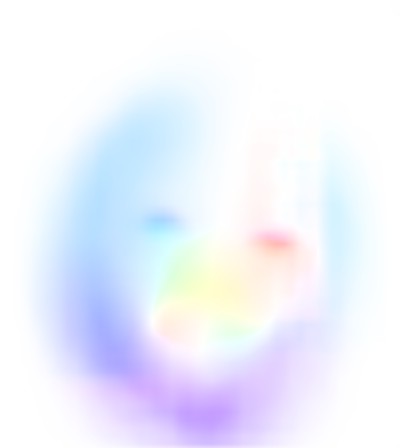} &
    \includegraphics[width=\vallinewidth]{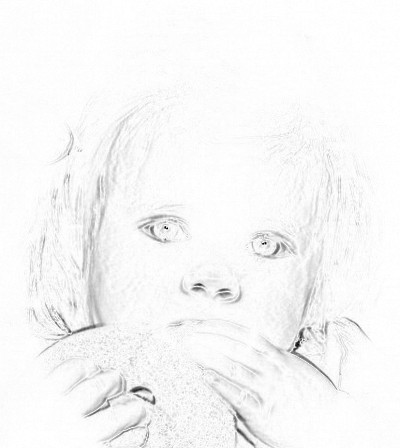} &
    \includegraphics[width=\vallinewidth]{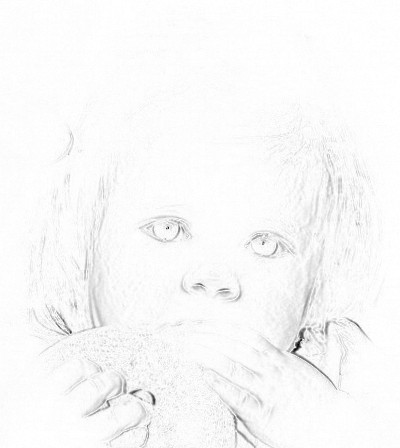} \\
    
    \includegraphics[width=\vallinewidth]{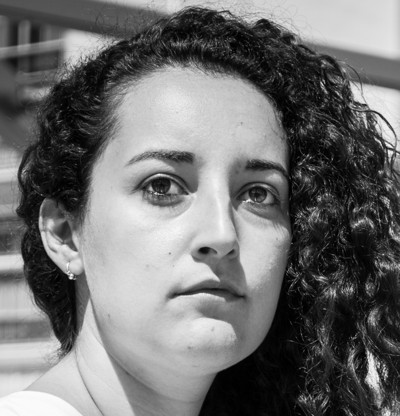} &
    \includegraphics[width=\vallinewidth]{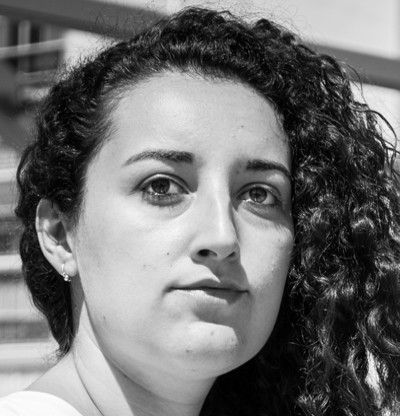} &
    \includegraphics[width=\vallinewidth]{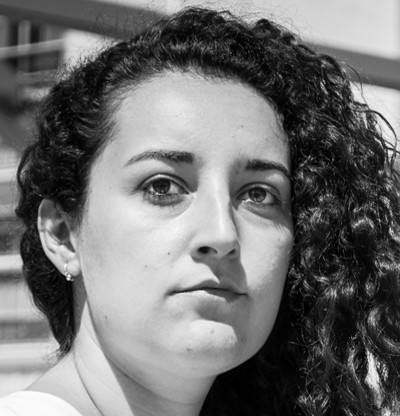} &
    \includegraphics[width=\vallinewidth]{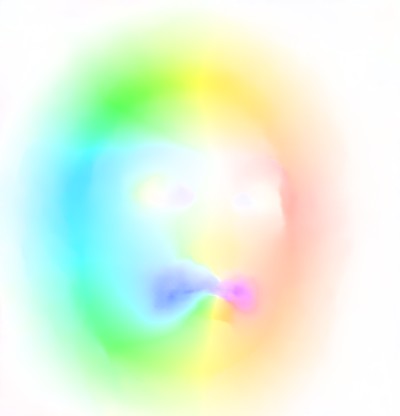} &
    \includegraphics[width=\vallinewidth]{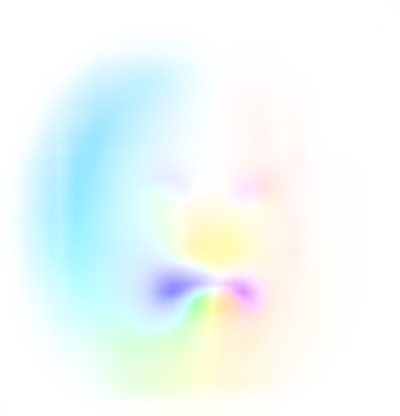} &
    \includegraphics[width=\vallinewidth]{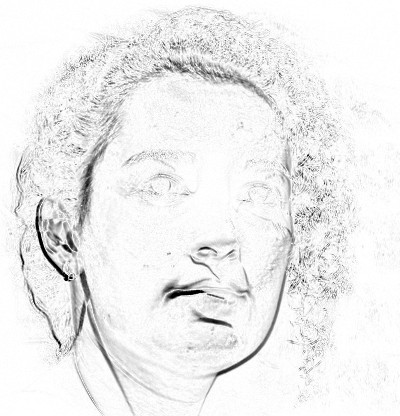} &
    \includegraphics[width=\vallinewidth]{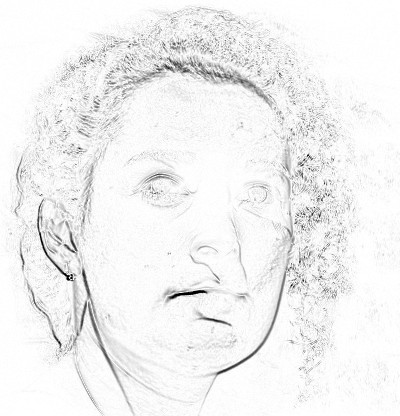} \\
    
    \includegraphics[width=\vallinewidth]{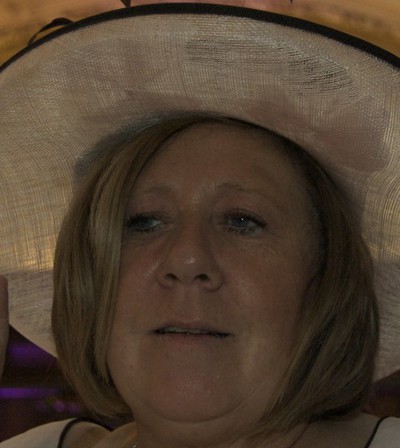} &
    \includegraphics[width=\vallinewidth]{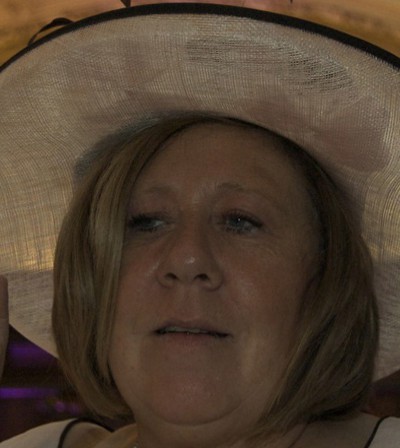} &
    \includegraphics[width=\vallinewidth]{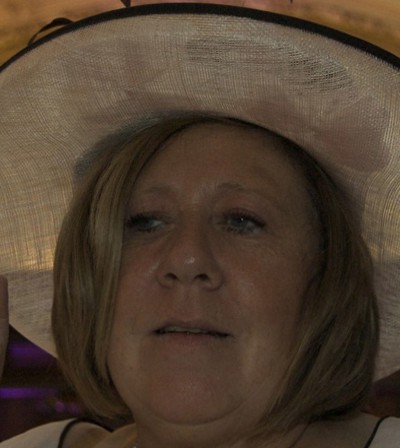} &
    \includegraphics[width=\vallinewidth]{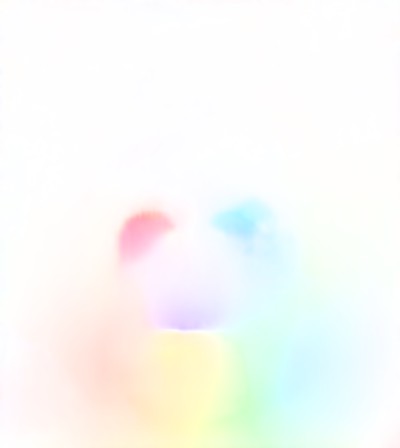} &
    \includegraphics[width=\vallinewidth]{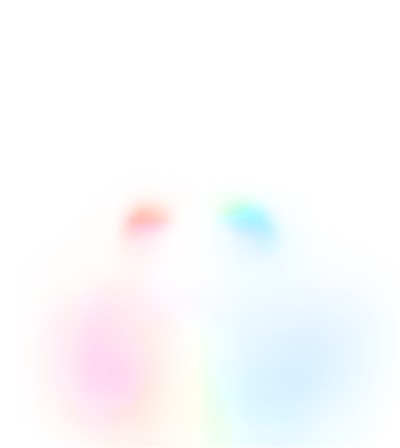} &
    \includegraphics[width=\vallinewidth]{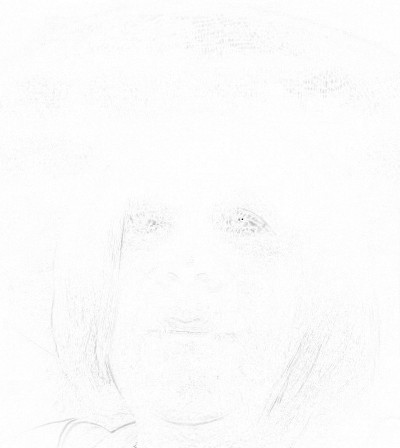} &
    \includegraphics[width=\vallinewidth]{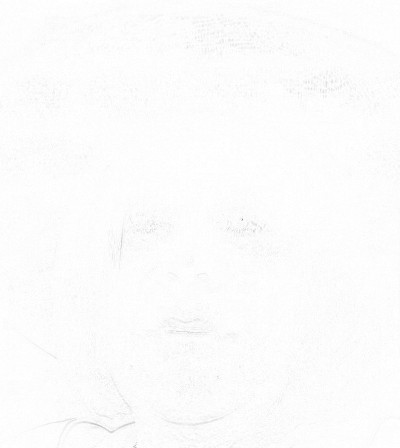} \\
    
    \includegraphics[width=\vallinewidth]{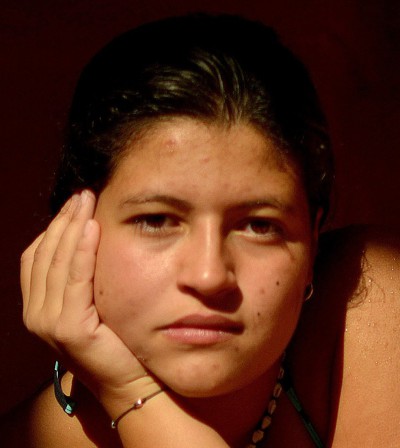} &
    \includegraphics[width=\vallinewidth]{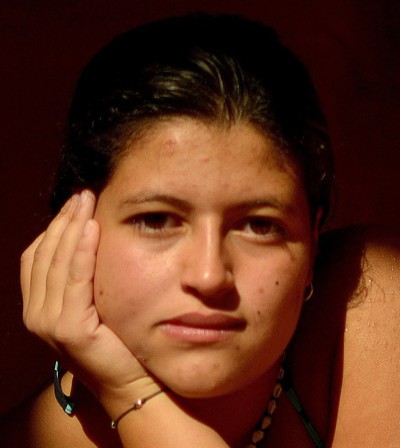} &
    \includegraphics[width=\vallinewidth]{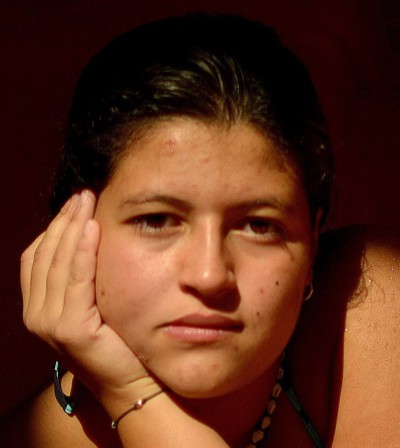} &
    \includegraphics[width=\vallinewidth]{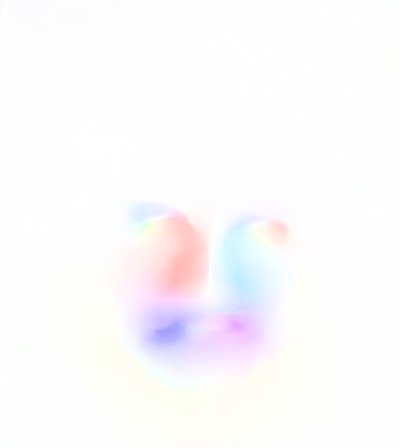} &
    \includegraphics[width=\vallinewidth]{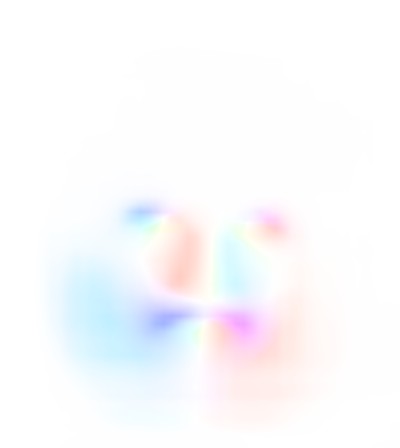} &
    \includegraphics[width=\vallinewidth]{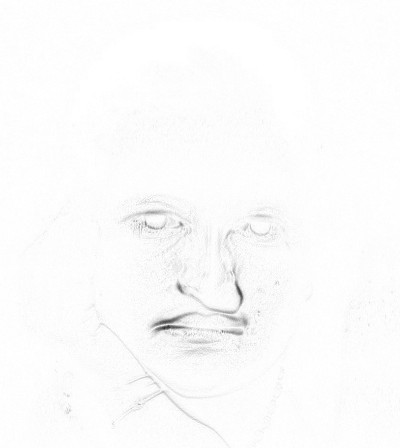} &
    \includegraphics[width=\vallinewidth]{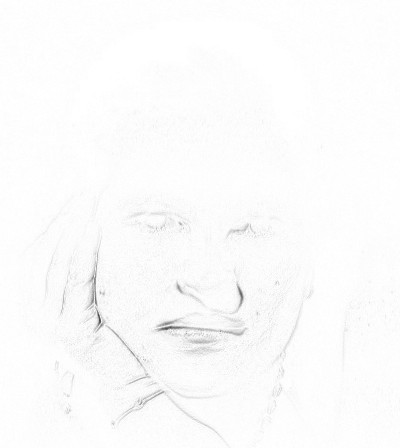} \\

    \includegraphics[width=\vallinewidth]{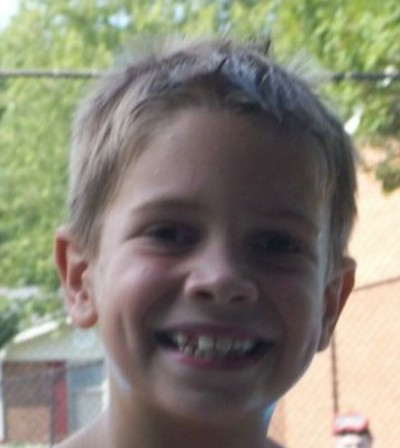} &
    \includegraphics[width=\vallinewidth]{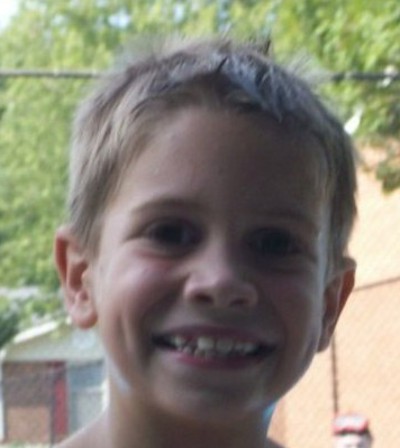} &
    \includegraphics[width=\vallinewidth]{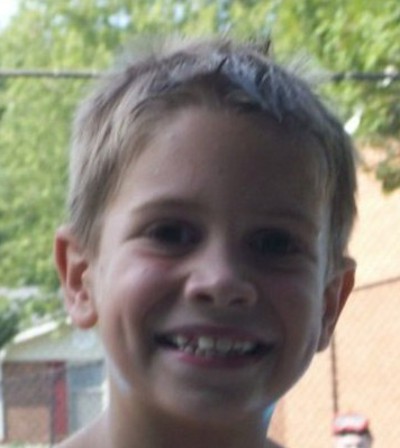} &
    \includegraphics[width=\vallinewidth]{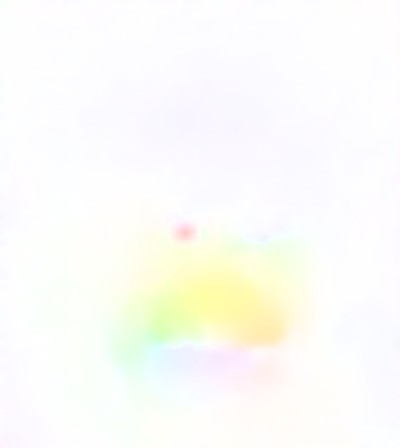} &
    \includegraphics[width=\vallinewidth]{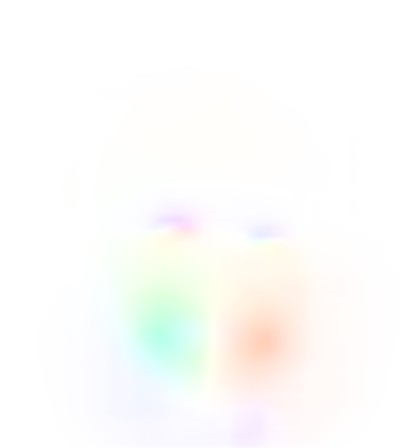} &
    \includegraphics[width=\vallinewidth]{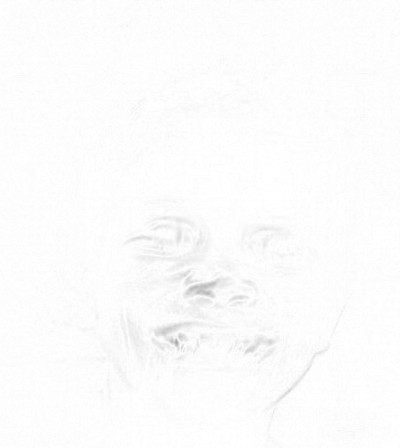} &
    \includegraphics[width=\vallinewidth]{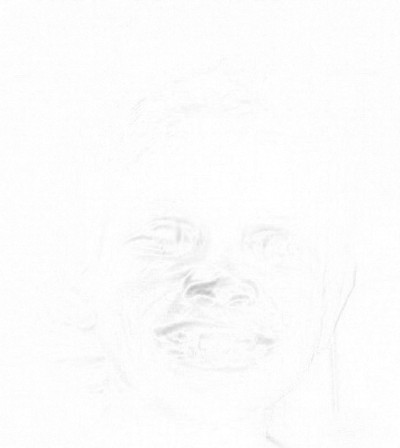} \\
    Original & Manipulated & Unwarped & Ground truth flow & Our flow & Manipulated diff. & Unwarped diff. \\
    \end{tabular}
    \caption{Randomly selected results from our held-out validation dataset, showing the original, warped, and unwarped images. The ground-truth and predicted flow fields, and the difference images between the manipulated and original image, and the unwarped and original images (enhanced for visibility).}
    \label{fig:randomval}
\end{figure*}

\myparagraph{Local predictions}
Figure~\ref{fig:randomval} shows a random selection of results from our validation dataset of automatically-generated manipulations.
We conducted an experiment where the PSNR change with respect to scaled versions of the predicted flow field are shown over the validation set (Figure~\ref{fig:psnr}).
We can see that the highest PSNR gain is where the scale factor is 1.0, which implies that our predicted flow fields do not contain a multiplicative bias, that might result from the regression loss.

\begin{figure*}[t]
{\small
    \def\camheight{3.0cm}
    \def\camheightb{2.9cm}
    \def\camwidth{2.8cm}
    \centering
    \begin{tabular}{ c@{\hspace{2px}} c@{\hspace{2px}} c@{\hspace{5px}} c@{\hspace{2px}} c@{\hspace{2px}} c@{\hspace{2px}} }
    \includegraphics[height=\camheight]{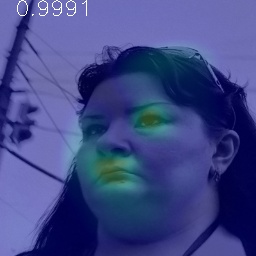} &
    \includegraphics[height=\camheight,width=\camwidth]{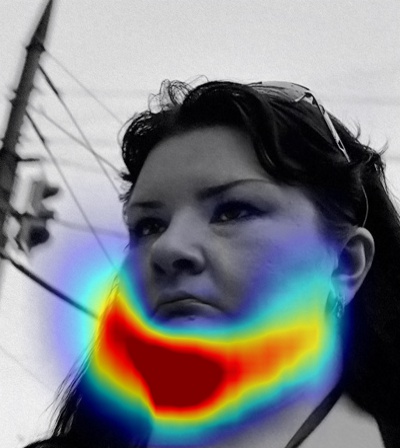} &
    \includegraphics[height=\camheight,width=\camwidth]{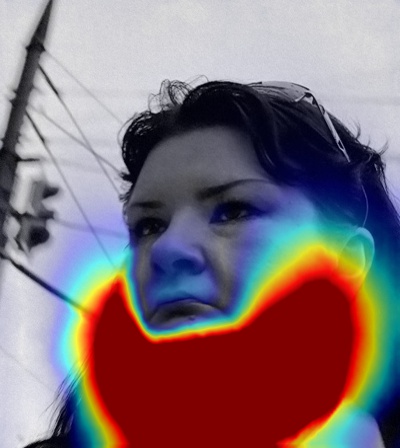} &
    \includegraphics[height=\camheight]{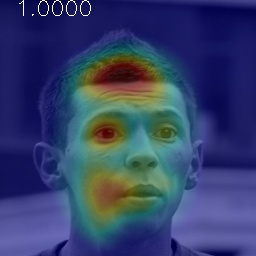} &
    \includegraphics[height=\camheight,width=\camwidth]{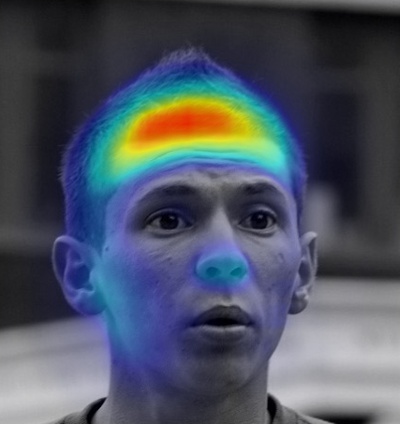} &
    \includegraphics[height=\camheight,width=\camwidth]{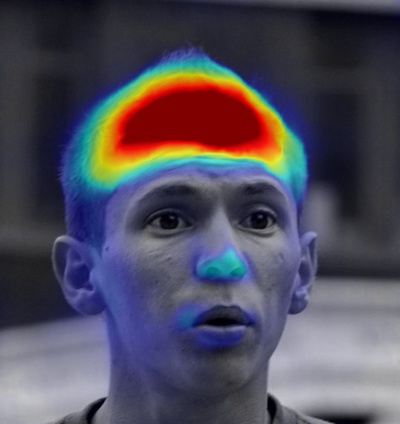} \\
    \includegraphics[height=\camheight]{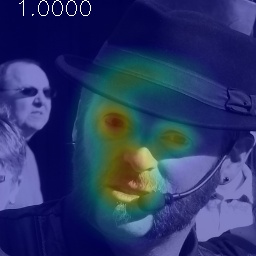} &
    \includegraphics[height=\camheight,width=\camwidth]{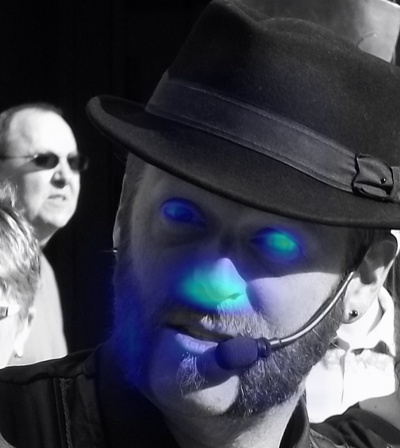} &
    \includegraphics[height=\camheight,width=\camwidth]{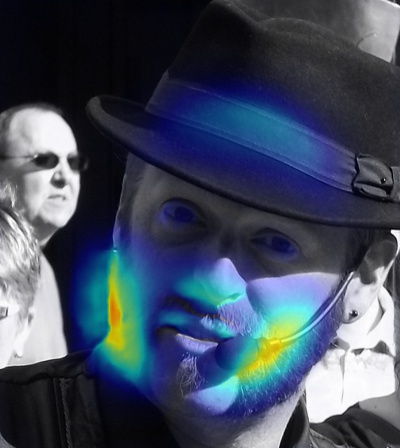} &
    \includegraphics[height=\camheight]{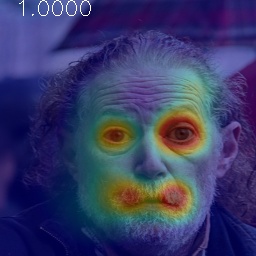} &
    \includegraphics[height=\camheight,width=\camwidth]{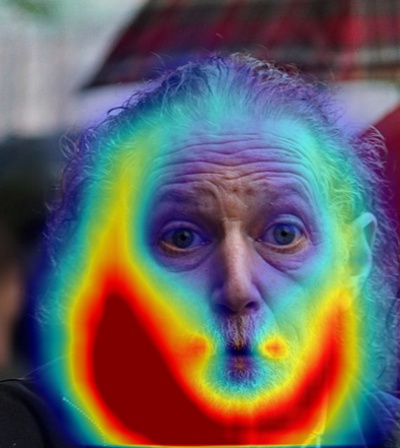} &
    \includegraphics[height=\camheight,width=\camwidth]{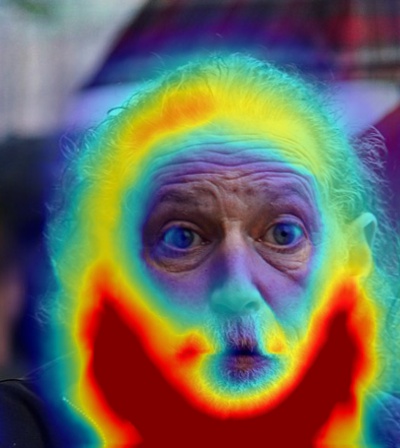} \\    \includegraphics[height=\camheight]{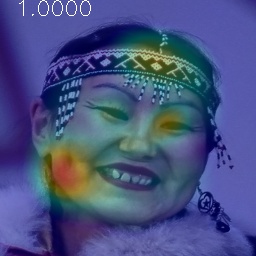} &
    \includegraphics[height=\camheight,width=\camwidth]{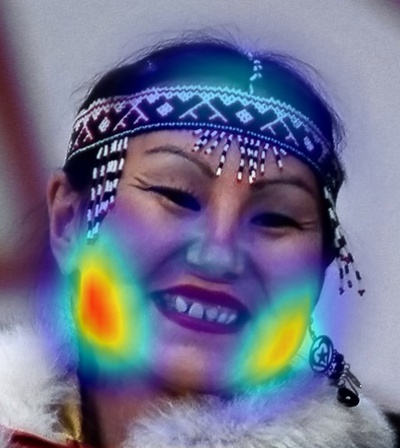} &
    \includegraphics[height=\camheight,width=\camwidth]{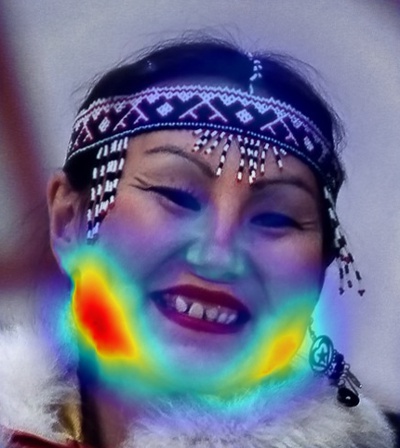} &
    \includegraphics[height=\camheight]{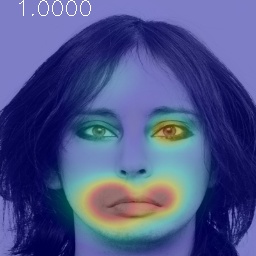} &
    \includegraphics[height=\camheight,width=\camwidth]{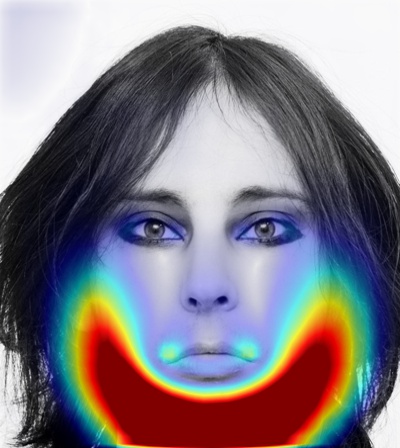} &
    \includegraphics[height=\camheight,width=\camwidth]{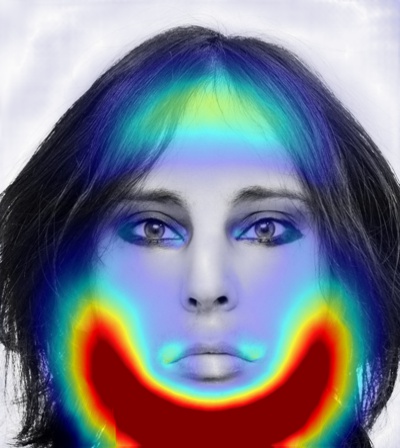} \\    \includegraphics[height=\camheight]{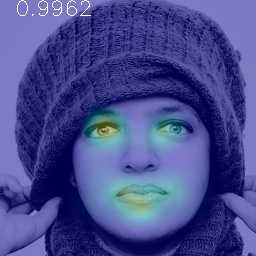} &
    \includegraphics[height=\camheight,width=\camwidth]{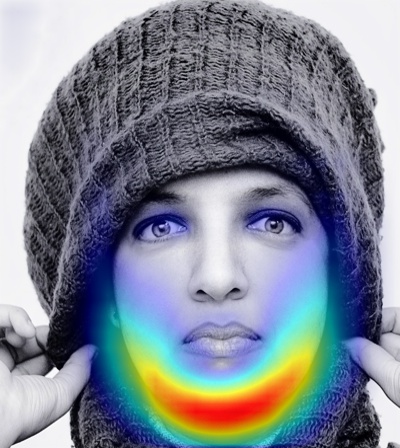} &
    \includegraphics[height=\camheight,width=\camwidth]{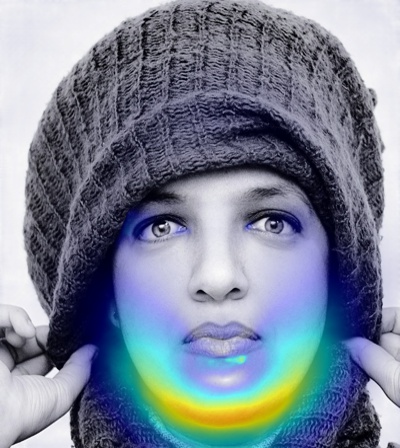} &
    \includegraphics[height=\camheight]{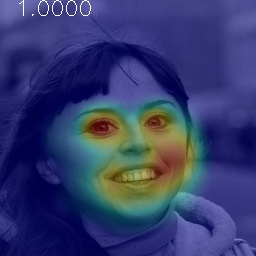} &
    \includegraphics[height=\camheight,width=\camwidth]{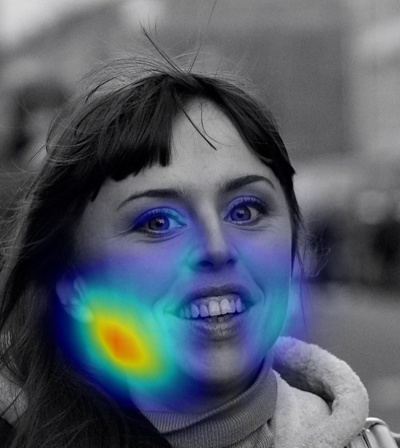} &
    \includegraphics[height=\camheight,width=\camwidth]{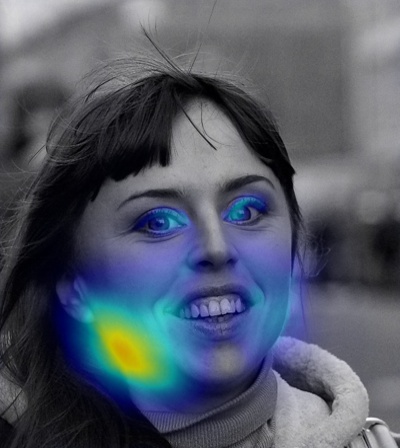} \\
    Class activation map & Predicted flow & Ground truth flow & Class activation map & Predicted flow & Ground truth flow \\
    \end{tabular}}
    \caption{Class activation maps on modified images. Numbers on the upper-left corners of the class activation maps are the modification probability assigned by our model. For reference, we also include the ground truth flow, and our prediction of it.} %
    \label{fig:classactivation}
\end{figure*}

\begin{figure*}[t]
    \def\camheight{2.8cm}
    \centering
    \begin{tabular}{ c@{\hspace{2px}} c@{\hspace{2px}} c@{\hspace{2px}} c@{\hspace{2px}} c@{\hspace{5px}} c@{\hspace{2px}} }
    \includegraphics[height=\camheight]{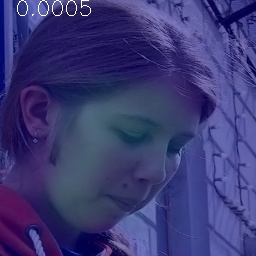} &
    \includegraphics[height=\camheight]{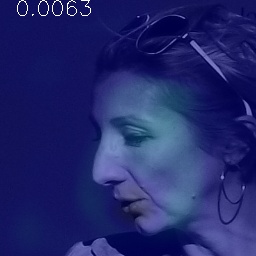} &
    \includegraphics[height=\camheight]{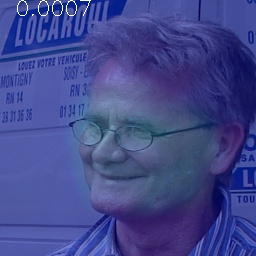} &
    \includegraphics[height=\camheight]{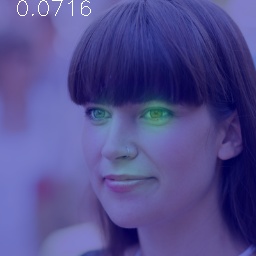} &
    \includegraphics[height=\camheight]{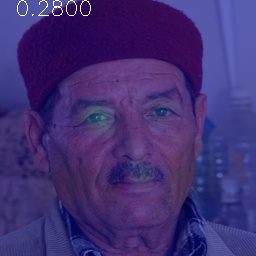} &
    \includegraphics[height=\camheight]{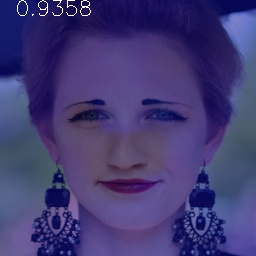} \\
    \includegraphics[height=\camheight]{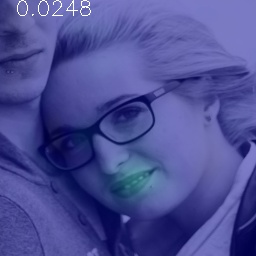} &
    \includegraphics[height=\camheight]{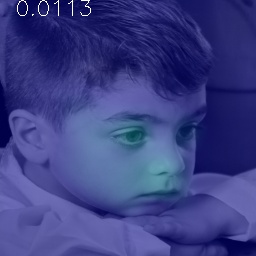} &
    \includegraphics[height=\camheight]{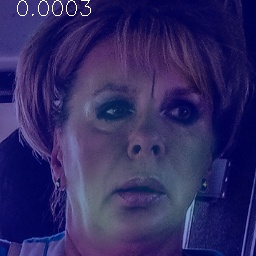} &
    \includegraphics[height=\camheight]{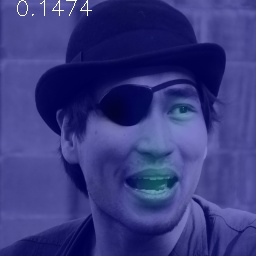} &
    \includegraphics[height=\camheight]{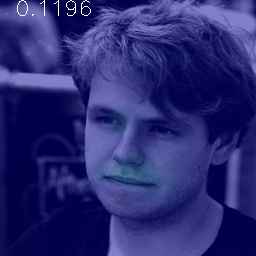} &
    \includegraphics[height=\camheight]{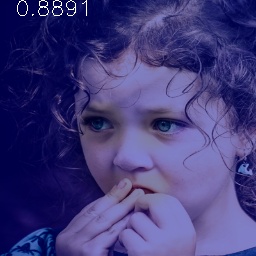} \\
    True negative & True negative & True negative & True negative & True negative & False positive \\
    \end{tabular}
    \caption{Class activation maps on randomly sampled original (unmodified) images. Numbers on the upper-left corners of the class activation maps are the modification probability assigned by our model.} %
    \label{fig:classactivationorg}
\end{figure*}

\myparagraph{Network visualization}
We visualize our global classifier using the class activation map method of Zhou \etal~\cite{zhou2016learning}.
Figures~\ref{fig:classactivation}, \ref{fig:classactivationorg} show a random selection of class activation maps of our global classifier. Note that our global classifier model is able to achieve high accuracy (93.7\%) despite the mismatch between class activation maps and ground truth flow. This suggests that the model may be able to pick up other cues to differentiate between original and manipulated images.

\section{Robustness to corruptions}
\label{sec:apdxrobustness}
\camready{
We tested the robustness of our model by perturbing the low-level statistics of our validation set through common corruptions such as lossy JPEG compression, blurring, and printing and scanning physical prints.} This offers three interesting test cases, as we \textit{did} train on JPEG compressed images, did \textit{not} train on blurring, and \textit{cannot} train on rescanned images due to the cost of dataset acquisition.

\begin{figure*}[t]
    \centering
    \begin{tabular}{*{3}{c@{\hspace{2px}}}}
    \includegraphics[height=3.7cm]{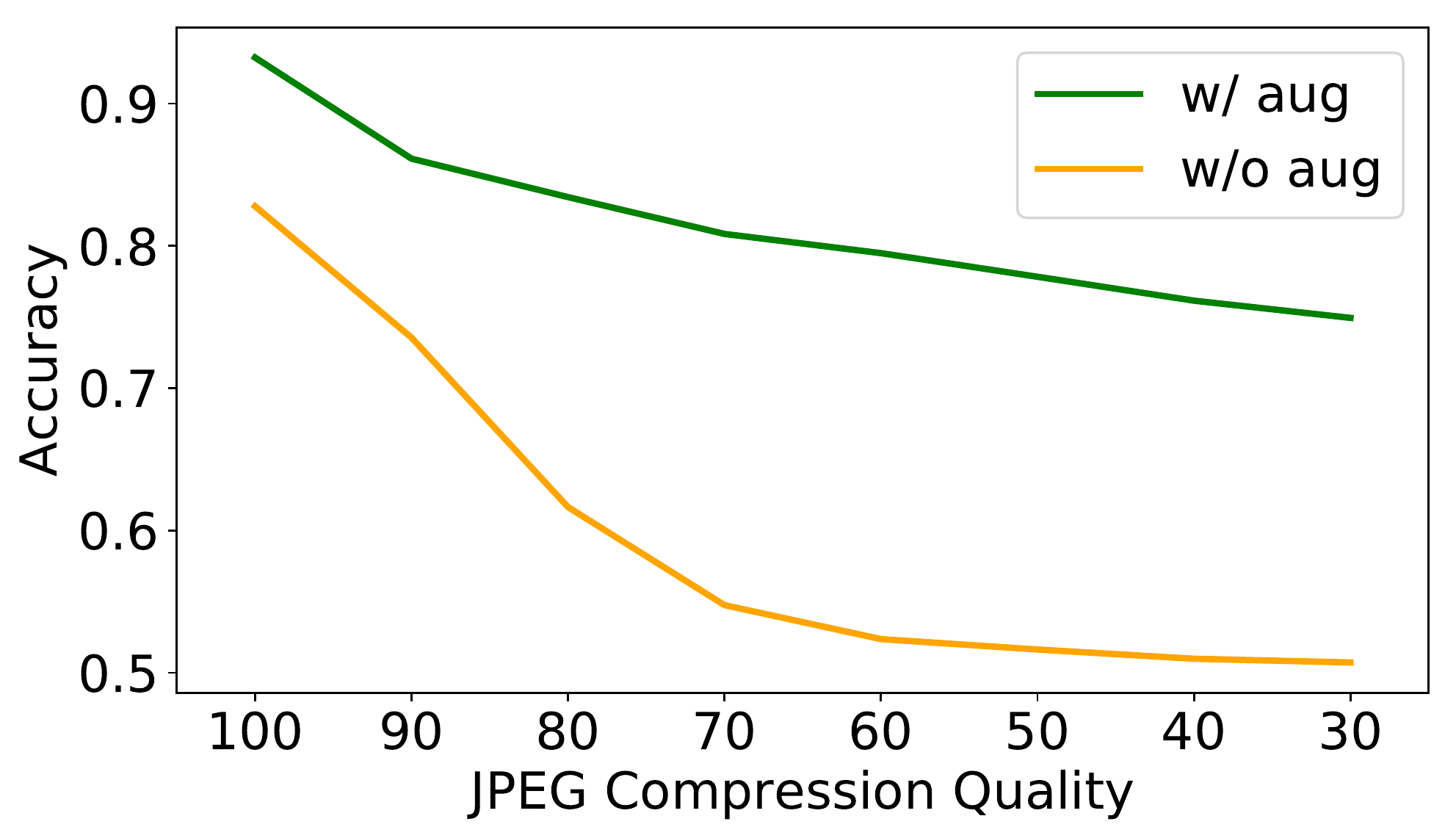} &
    \includegraphics[height=3.7cm]{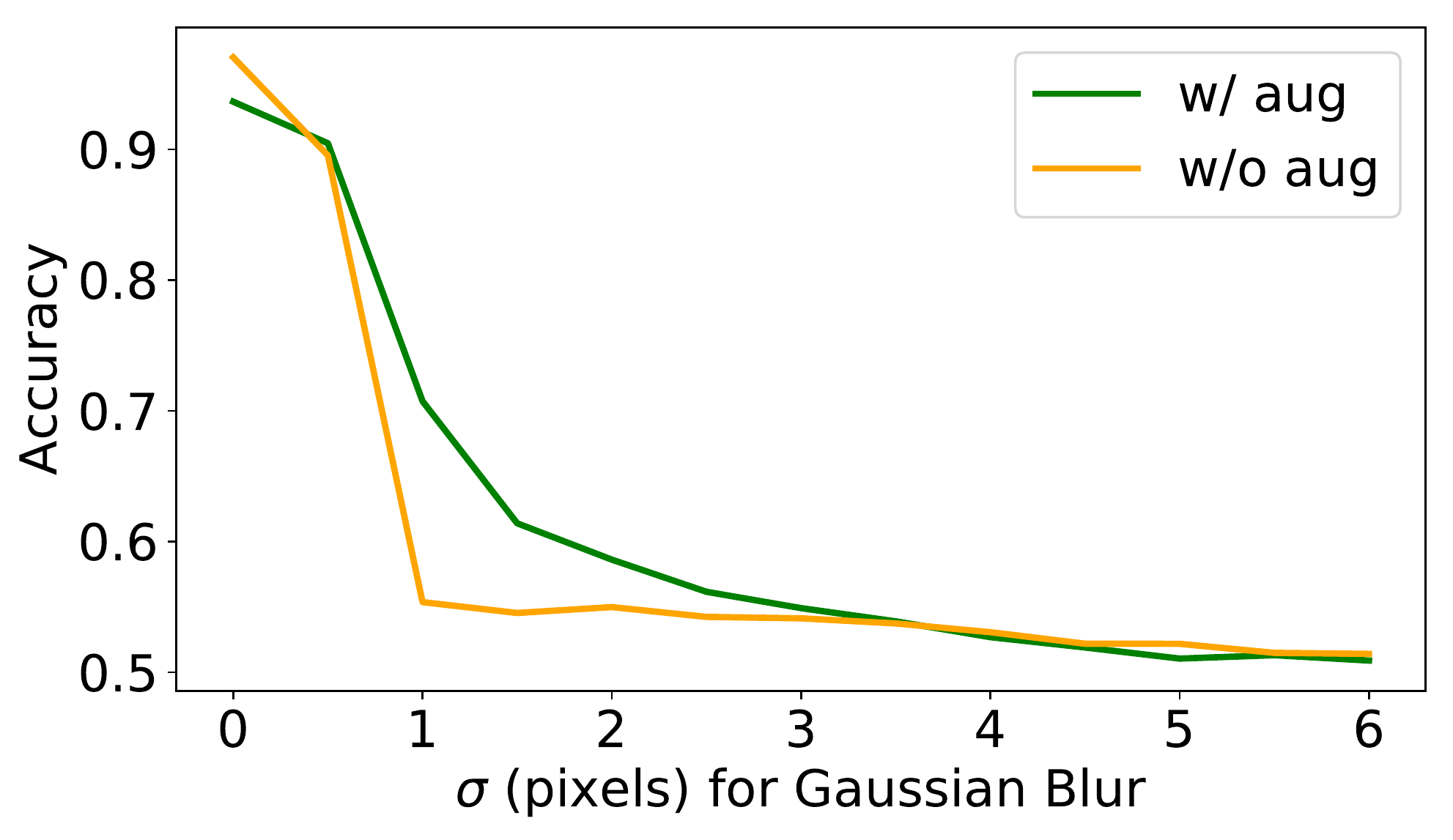} &
    \raisebox{3px}{\includegraphics[height=3.9cm]{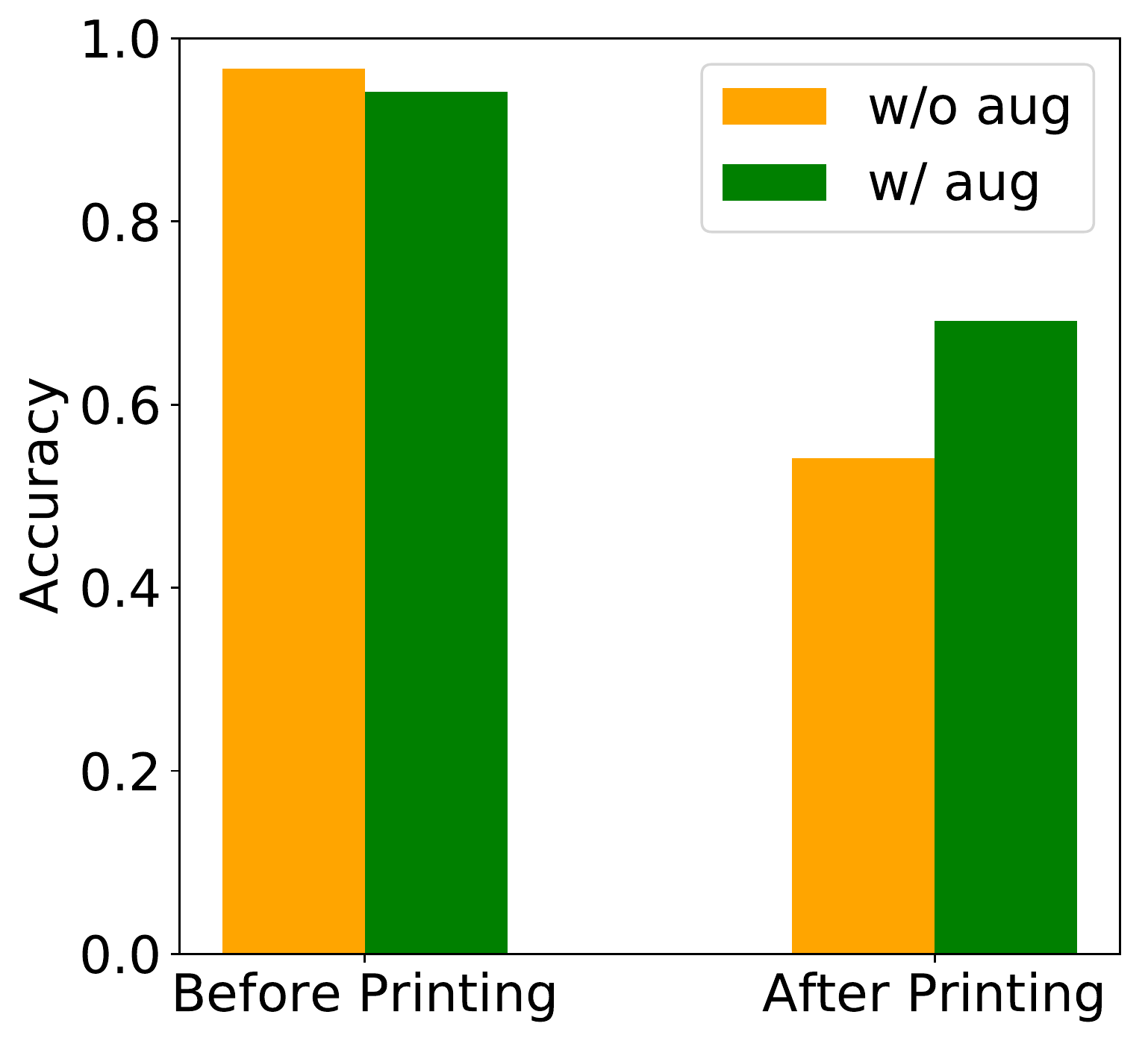}} \
    \end{tabular}
    \vspace{-.1in}
    \caption{{\bf Robustness to corruptions.} Accuracy of global classification with JPEG, Gaussian blur, and after rescanning, with and without data augmentation. (left) JPEG compression: a significant increase in robustness. Though unsurprising, as it is in our augmentation set, it is important, as compression and recompression is commonly applied to images. (middle) Blur: \camlast{although this is not in our augmentation set, we observe a small increase in robustness.} (right) Rescanning: a small increase in robustness. We corrupt by printing and rescanning a subset of photos, a perturbation that cannot be reproduced during training}
    \label{fig:perturbations}
    \vspace{-.1in}
\end{figure*}

As shown in Fig.~\ref{fig:perturbations}, the method with augmentation is fairly robust to JPEG compression. 
\camlast{Though we did not train with blurring augmentations (as images are unlikely to be intentionally blurred), training with other augmentations helps increase resilience. However, with significant blur ($\sigma > 4$), performance degrades to chance levels. This indicates that the classifier is relying on some high frequency information, which is the main component attenuated by the Gaussian filter.}

Lastly, we also test the robustness of our classifier to print rebroadcasting~\cite{fan2018rebroadcast}, testing on images that are printed, and then re-digitized by a scanner (\eg, simulating the task of identifying manipulations in magazine covers). We used a Canon imageRunner Advance C3530i Multifunctional copier and standard 8.5$\times$11 inch paper. 
We randomly selected 30 images each from the Flickr and OpenImages sets.
Classification performance drops from $94.2\%$ to $69.2\%$ (standard error of $6.0\%)$. While rebroadcasting hurts performance, our model still detects manipulated images significantly more accurately than chance.

\begin{table}[h]
\begin{center}
    \begin{tabular}{@{}lccccc@{}}
    
    \toprule & \multicolumn{2}{c}{\bf Global} && \multicolumn{2}{c}{\bf Local}\\
    
    \cmidrule{2-3} \cmidrule{5-6} & {\bf Acc.} & {\bf AP} && {\bf $\Delta$PSNR} & {\bf IOU-3}\\ \midrule
    
    {\bf Face / FAL} & 93.7 & 98.9 && +2.69 & 0.43\\
    {\bf Face / X2face} & 64.7 & 74.0 && +0.13 & 0.05\\
    {\bf Noise / FAL} & 44.5 & 92.9 && -- & 0.43\\
    {\bf Noise / X2face} & 36.5 & 82.0 && -- & 0.03\\
    {\bf Natural / FAL} & 67.7 & 77.3 && +0.12 & 0.05\\
    \bottomrule
    \end{tabular}
\end{center}
\caption{{\bf Generalization results}. We tested the generalization of our global and local models on four out-of-distribution dataset. The top row (Face/FAL) contains the results of our original validation set for comparison.}
\vspace{-4mm}
\label{tab:generalization}
\end{table}

\section{Generalization}
\label{sec:apdxgeneralization}
\camready{
We are interested in what cues in the images the model learns to focus on, in order to detect warping. 
For example, is the model looking at low-level image statistics ({\em e.g.} resampling artifacts) or high-level cues ({\em e.g.} facial geometric inconsistencies)? 
This has larger implications for example in whether the model can detect warps only realizable by FAL, or can it detect more general warping scenarios? 
To investigate, we evaluate our global and local models in four different scenarios: (1) images composed of noise, warped with FAL warps, (2) images composed of noise warped with out-of-domain warps, (3) out-of-domain natural images warped with FAL warps, and (4) portrait images warped with out-of-domain warps.}

\camready{
To generate out-of-domain warps, we randomly sampled the latent space of the optical flow generator in the X2face model~\cite{wiles2018x2face} to generate warps. We note that although the X2face model is trained to generate face-specific warps, the warping field will not necessarily align with the portrait; moreover, since a VAE loss is not included during X2face training, sampling the bottleneck does not guarantee to have realistic warping fields. However, empirically we observed our sampling method generates smooth warping fields that modifies the face in a ``stochastic" fashion. That is, the X2face warping field will not specifically change a face in a meaningful way such as making someone's smile bigger or face smaller.
On the other hand, for out-of-domain images we collected natural images from random samples in Open Images~\cite{krasin2016openimages}, which are not portrait images. Table~\ref{tab:generalization} shows the results. }

\camready{
Note that when there is a domain shift in warping field (face/X2face) or image space (natural/FAL), the performances of both models drop significantly although still perform above chance ($50\%$ Accuracy and $0$ $\Delta$PSNR). \camlast{More interestingly, note that our global model is able to generalize to warped noise with FAL and X2face flows at a certain degree if well-calibrated (92.9, 82.0 AP), and our local model generalizes specifically to FAL-warped noise. This indicates they have learned low-level warping cues, while the local model is more specific to FAL warping field statistics.} However, we trained global and local models solely on noise warped with FAL flows and tested on our validation set, and the models are only able to achieve $49.6 \%$ accuracy and $28.28$ EPE respectively. This suggests that our model {\em has} learned low-level cues, but that low-level cues are not  sufficient: {\em the face warping problem is much more difficult}.}

\begin{figure}
    \centering
    \def\portraitwidth{0.34\linewidth}
    \begin{tabular}{*{3}{c@{\hspace{1px}}}}
    \includegraphics[height=\portraitwidth]{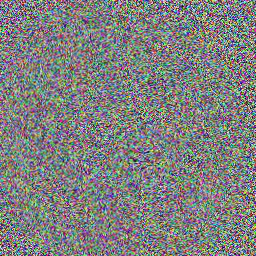} &
    \includegraphics[height=\portraitwidth]{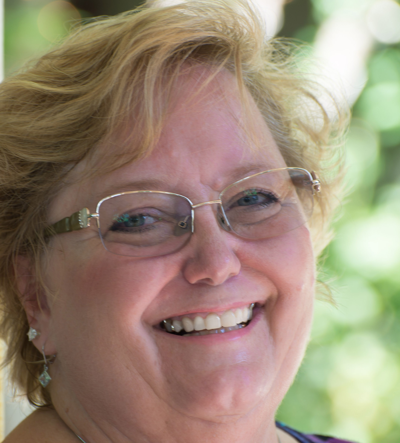} &
    \includegraphics[height=\portraitwidth]{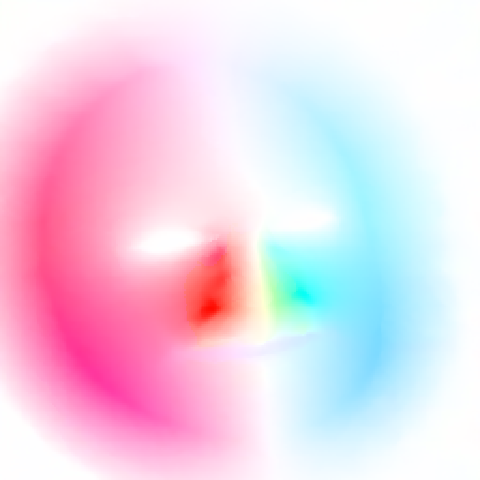} \\
    \end{tabular}
    \caption{{\small
    {\bf Noise experiment setup}. The Gaussian noise image (left) and the face image (middle) are deformed with the same warping field (right). Our model trained on faces can detect the warped noise \camlast{(if well-calibrated)}, but a model trained on noise cannot detect the warped face. %
    }}
    \vspace{-5.5mm}
    \label{fig:noisefig}
\end{figure}

\section{Additional data collection details}
\label{sec:apdxdatainfo}
\begin{figure*}[t]
    \centering
    \def\vallinewidth{.13\linewidth}
    \begin{tabular}{c@{\hspace{8px}} *{6}{c@{\hspace{2px}}}}

    \includegraphics[width=\vallinewidth]{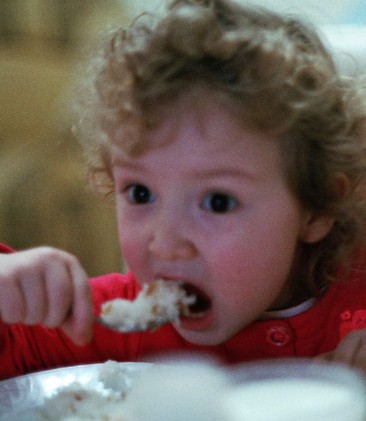} &
    \includegraphics[width=\vallinewidth]{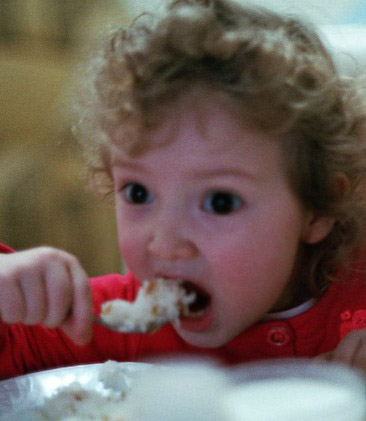}&
    \includegraphics[width=\vallinewidth]{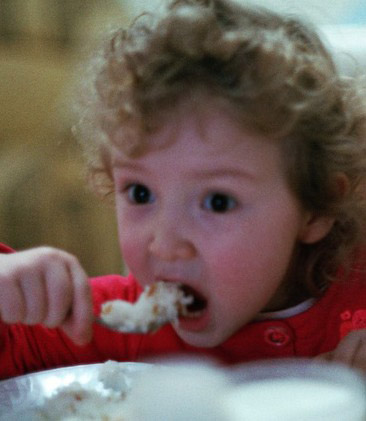}&
    \includegraphics[width=\vallinewidth]{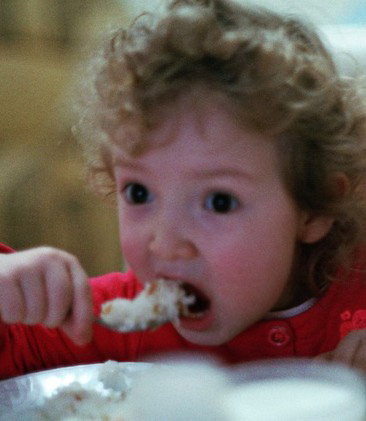}&
    \includegraphics[width=\vallinewidth]{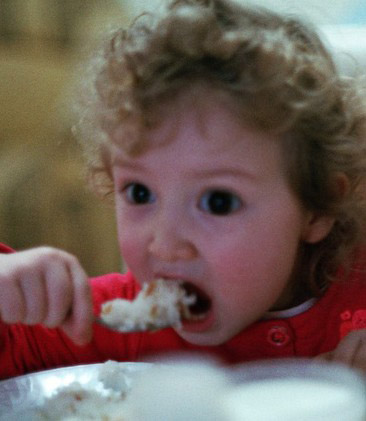}&
    \includegraphics[width=\vallinewidth]{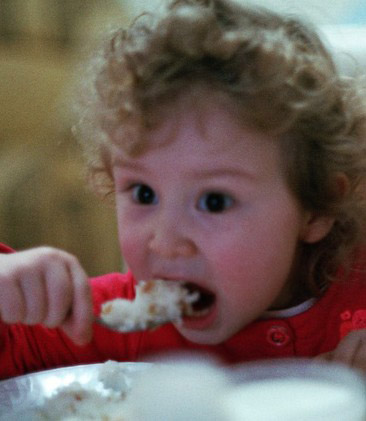}&
    \includegraphics[width=\vallinewidth]{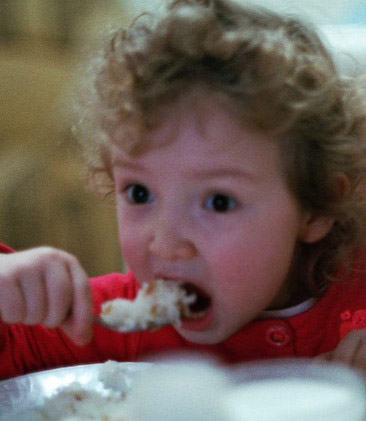}\\
    
    \includegraphics[width=\vallinewidth]{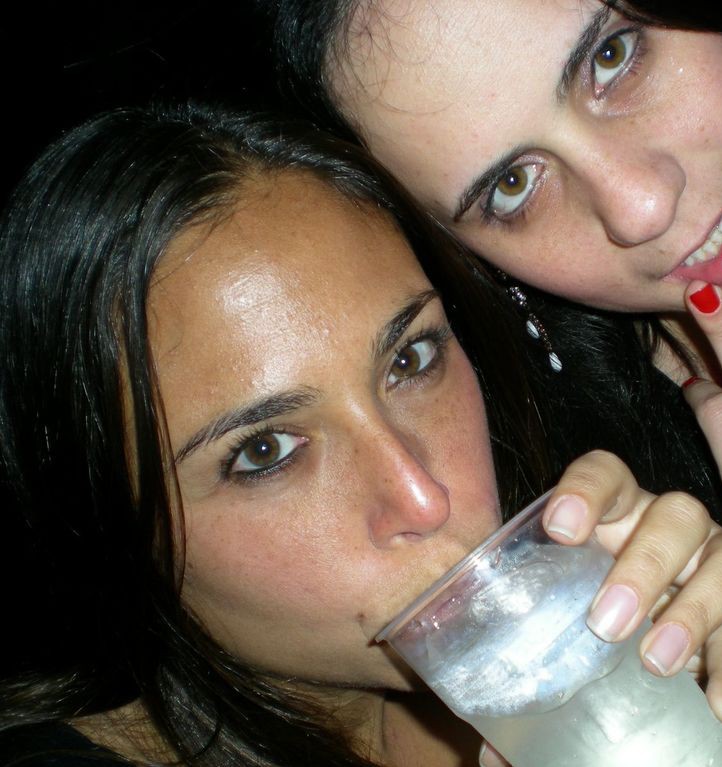} &
    \includegraphics[width=\vallinewidth]{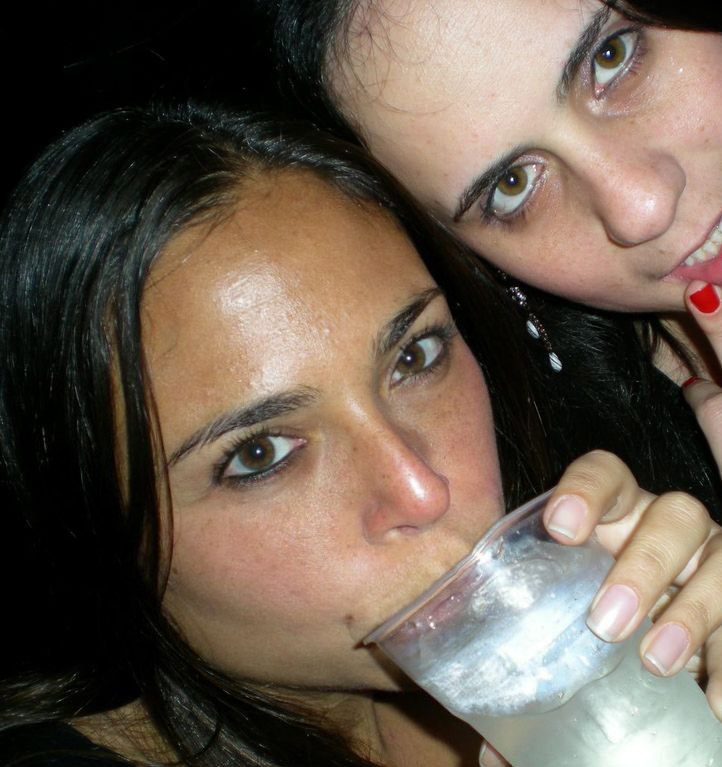}&
    \includegraphics[width=\vallinewidth]{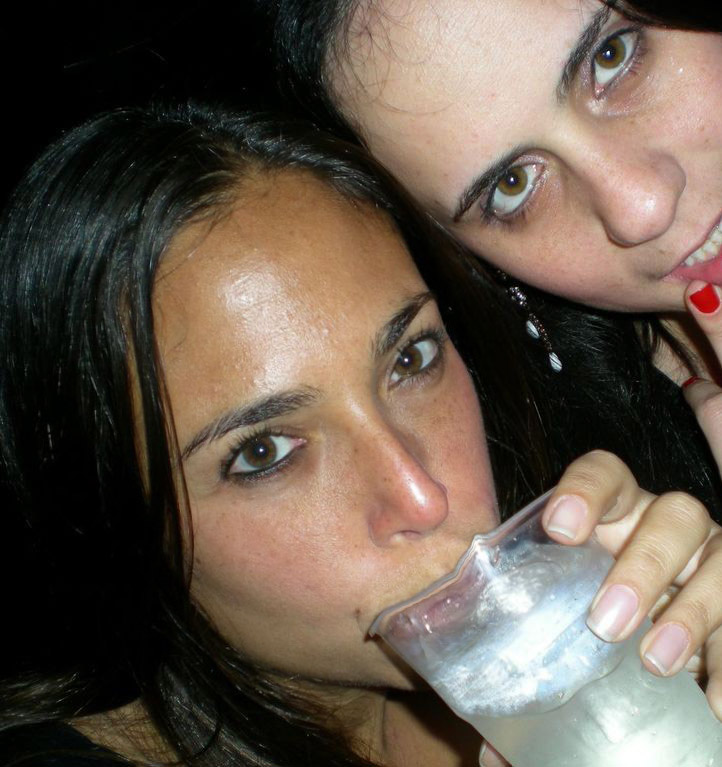}&
    \includegraphics[width=\vallinewidth]{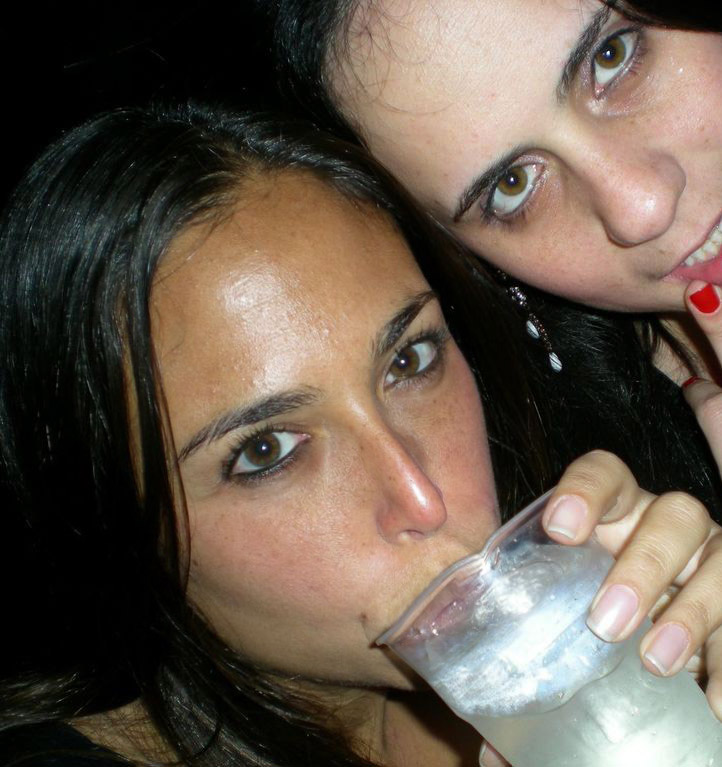}&
    \includegraphics[width=\vallinewidth]{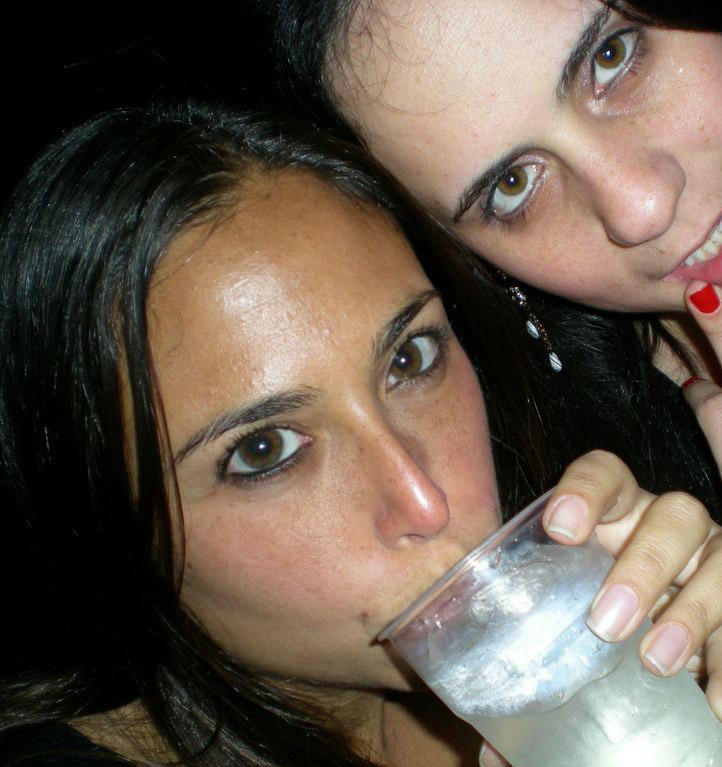}&
    \includegraphics[width=\vallinewidth]{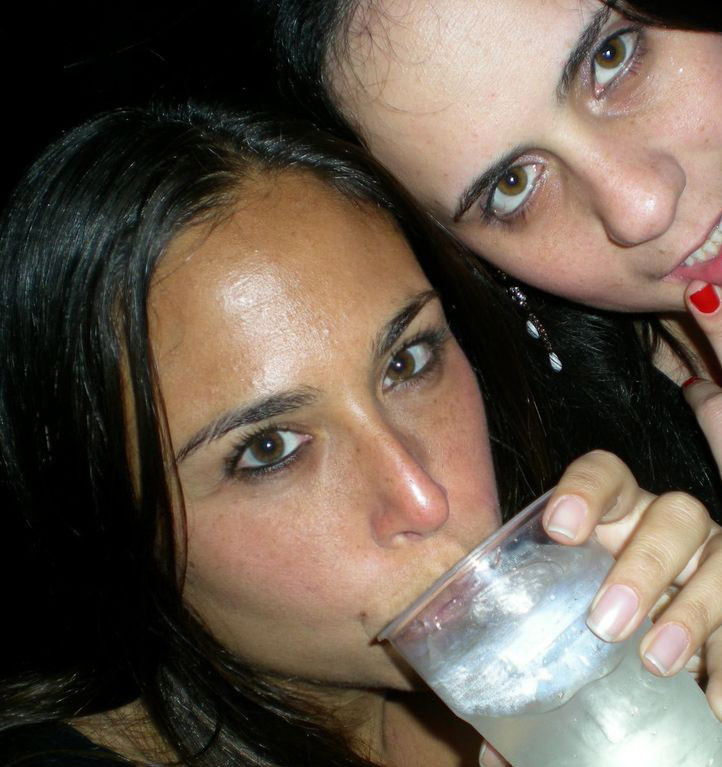}&
    \includegraphics[width=\vallinewidth]{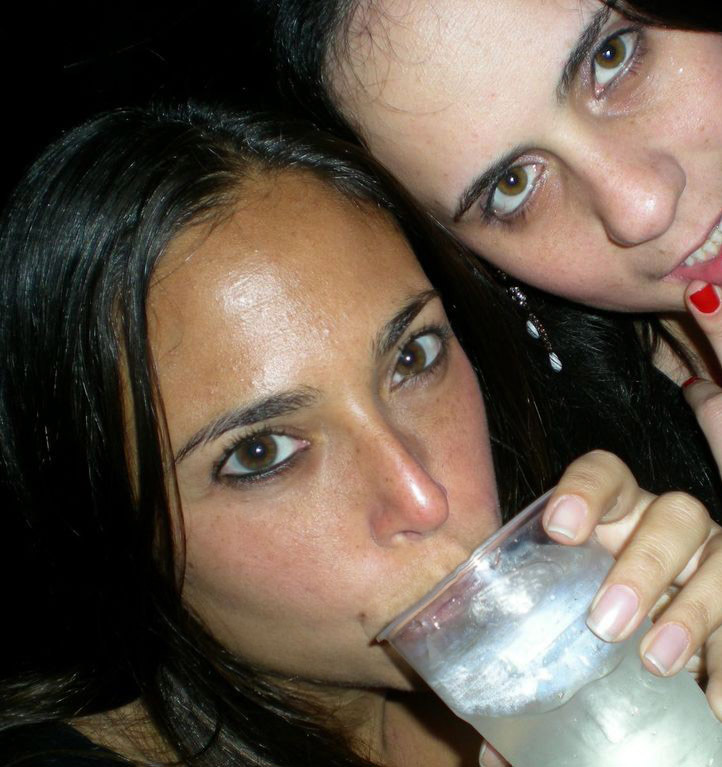}\\
    
    \includegraphics[width=\vallinewidth]{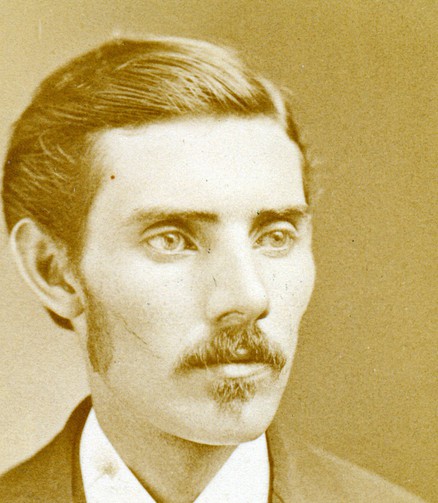} &
    \includegraphics[width=\vallinewidth]{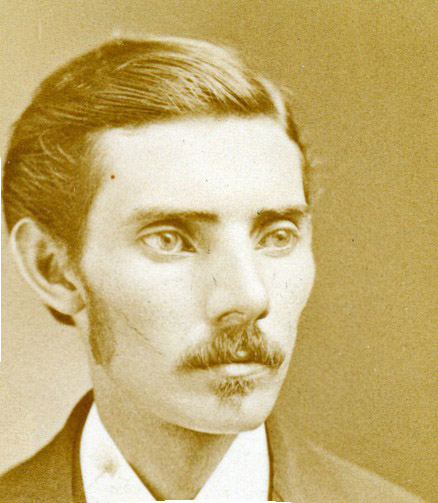}&
    \includegraphics[width=\vallinewidth]{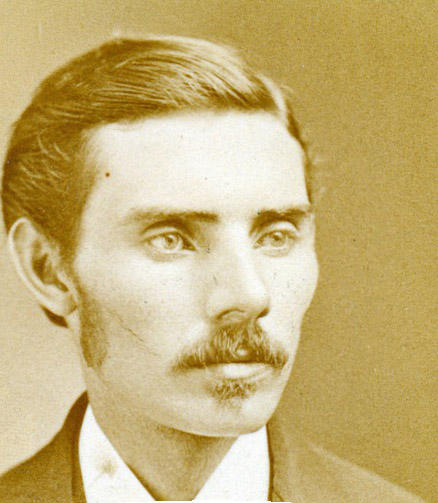}&
    \includegraphics[width=\vallinewidth]{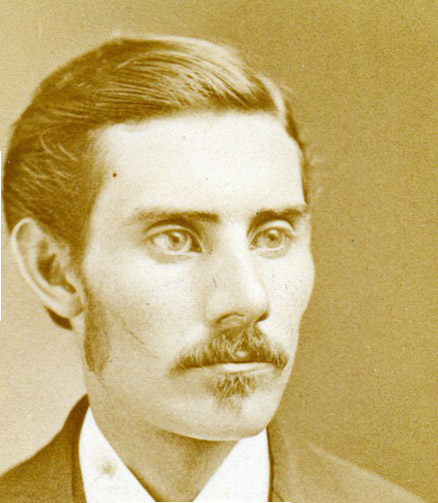}&
    \includegraphics[width=\vallinewidth]{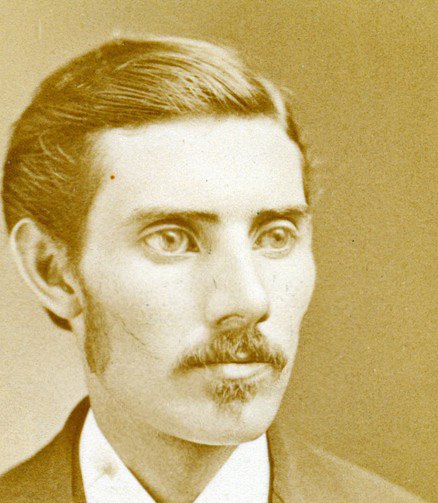}&
    \includegraphics[width=\vallinewidth]{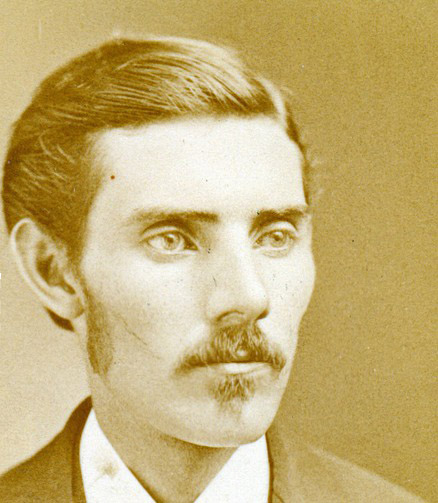}&
    \includegraphics[width=\vallinewidth]{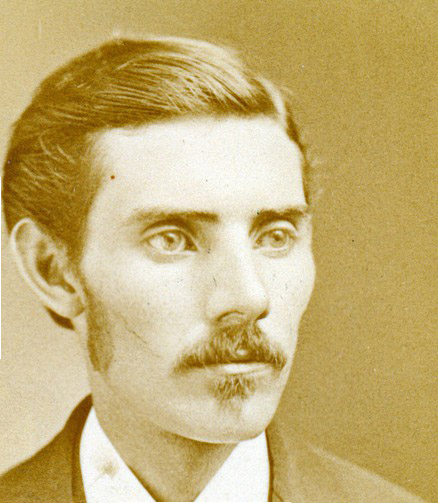}\\
    
    \includegraphics[width=\vallinewidth]{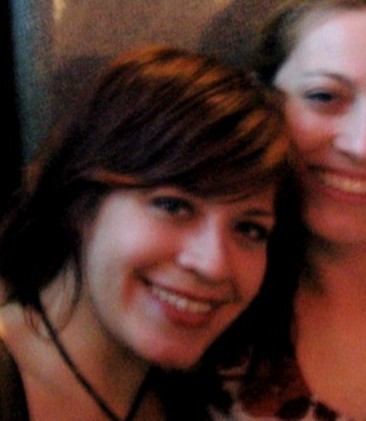} &
    \includegraphics[width=\vallinewidth]{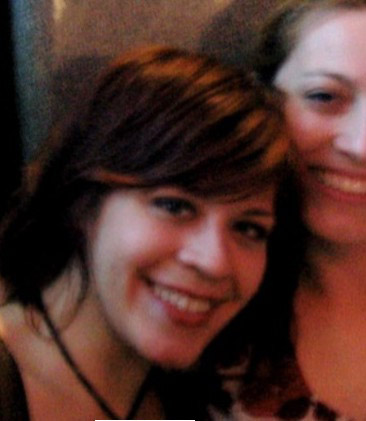}&
    \includegraphics[width=\vallinewidth]{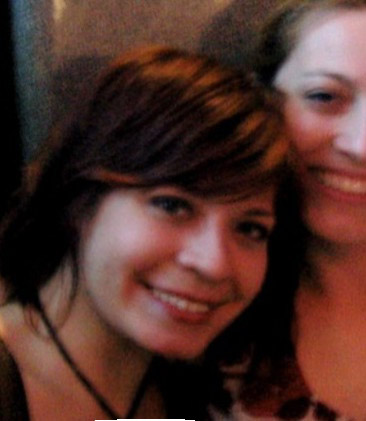}&
    \includegraphics[width=\vallinewidth]{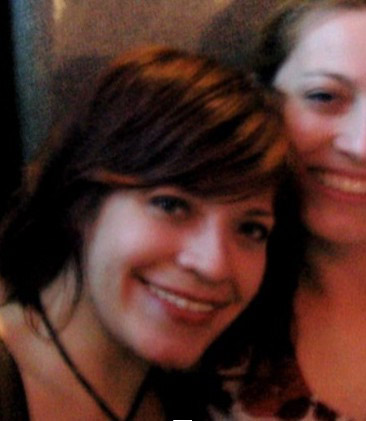}&
    \includegraphics[width=\vallinewidth]{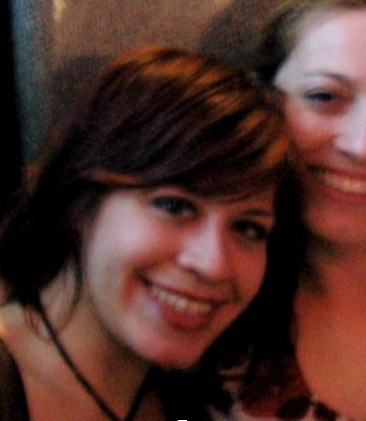}&
    \includegraphics[width=\vallinewidth]{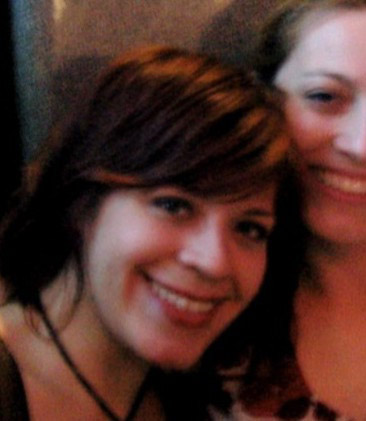}&
    \includegraphics[width=\vallinewidth]{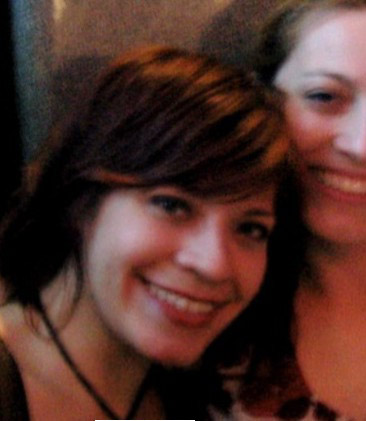}\\
    
    \includegraphics[width=\vallinewidth]{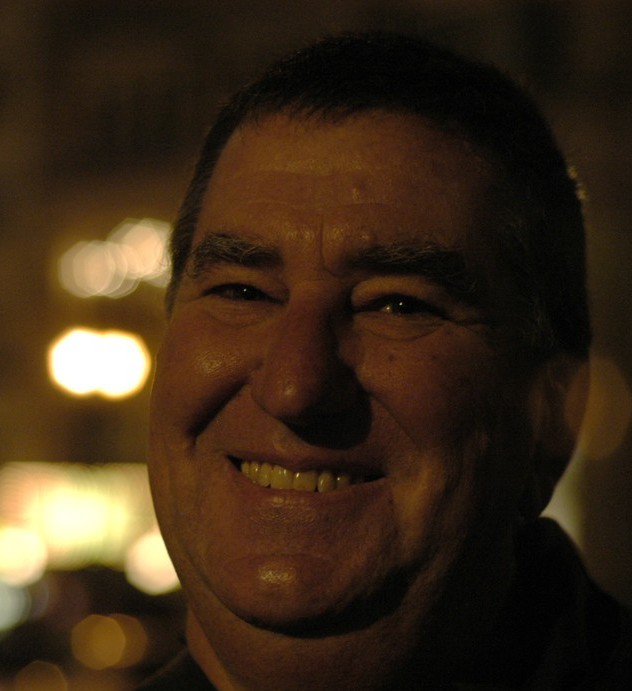} &
    \includegraphics[width=\vallinewidth]{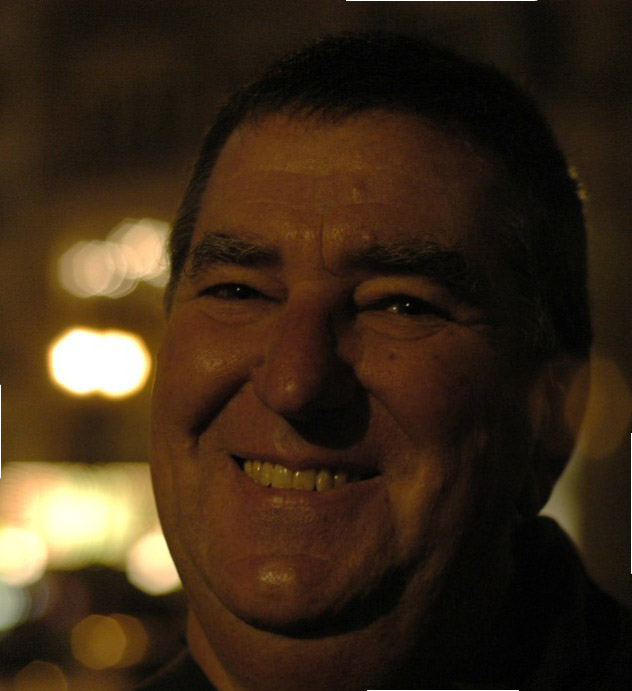}&
    \includegraphics[width=\vallinewidth]{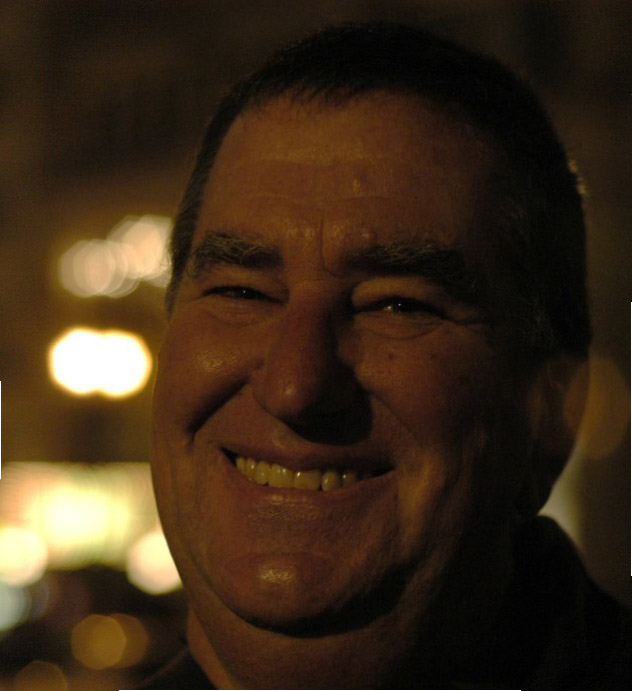}&
    \includegraphics[width=\vallinewidth]{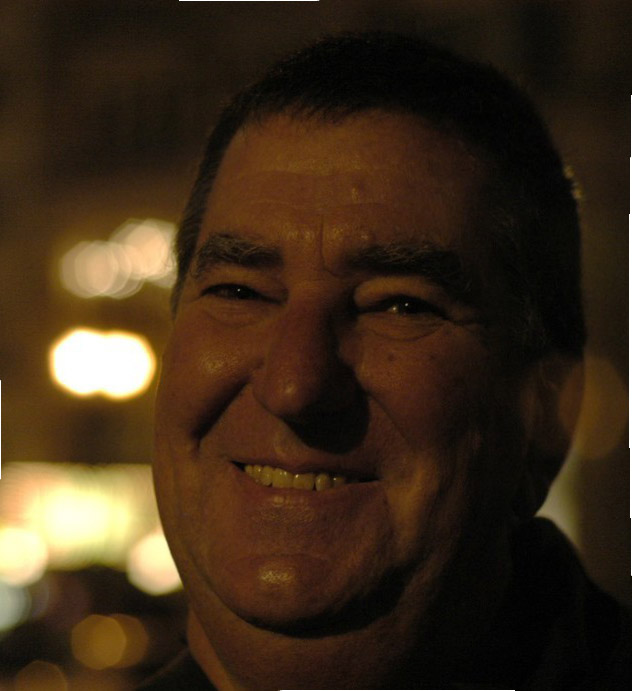}&
    \includegraphics[width=\vallinewidth]{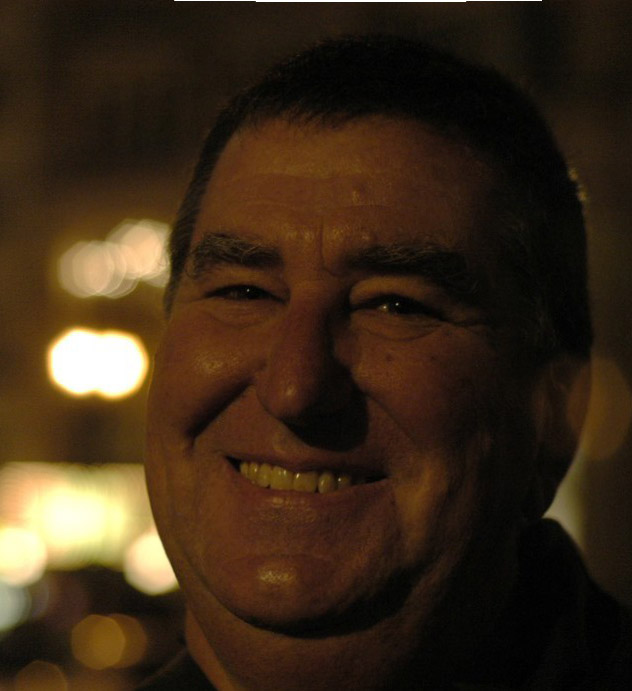}&
    \includegraphics[width=\vallinewidth]{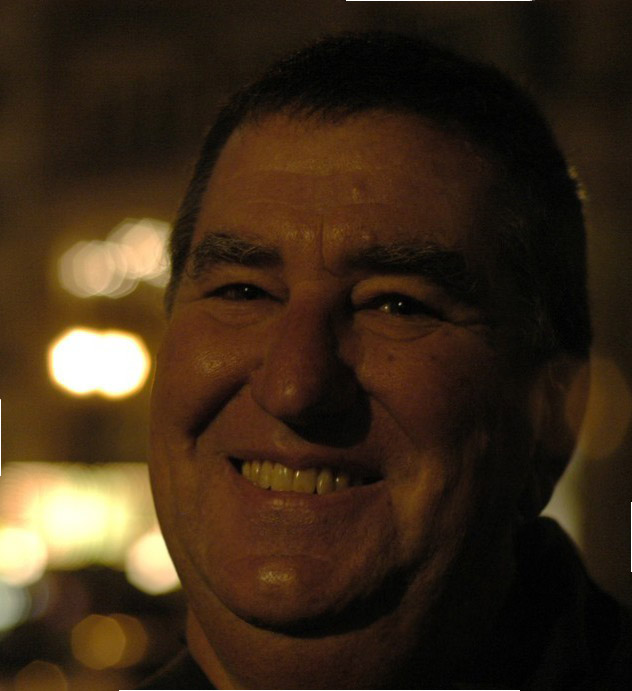}&
    \includegraphics[width=\vallinewidth]{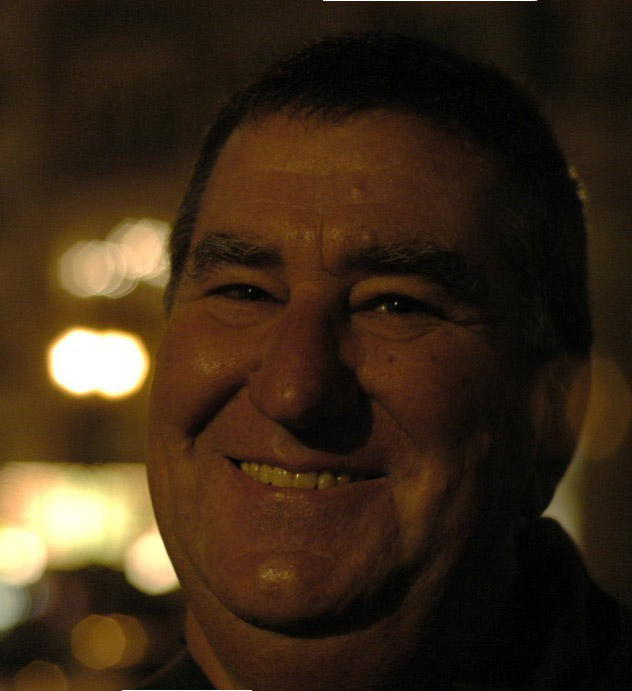}\\

    Input & \multicolumn{6}{c}{Manipulations} \\
    \end{tabular}
    \caption{\label{fig:mods}A random sample of manipulations from our dataset. For each photo, we show all 6 random edits that we made. We note that many of these modifications are subtle.}
\end{figure*}

Figure~\ref{fig:mods} shows a sample of the manipulations in our automatically-generated dataset. For each example photo, we show all 6 random manipulations that were applied to it.

\myparagraph{Collecting real face images} 
To obtain a diverse dataset of faces, we aggregate images from a variety of sources. First, we take all images from the Open Images dataset~\cite{krasin2016openimages} with the ``human face'' label.
This dataset consists of humans in-the-wild, scraped from Flickr. %
We also scrape Flickr specifically for portrait photography images.
To isolate the faces, we use an out-of-the-box CNN-based face detector from dlib~\cite{king2009dlib} and crop
the face region only.  
All together, our face dataset contains 69k and 116k faces from OpenImages and Flickr portrait photos, respectively, of which approximately 65k are high-resolution (at least 700 pixels on the shortest side). 
We note that the our dataset is biased toward Flickr users, who, on average, post higher-quality photographs than users of other Internet platforms.  More problematically, the Flickr user base is predominantly Western.
However, as our method is entirely self-supervised, it is easy to collect and train with new data to match the test distribution for a target application. 

\section{Implementation and training details}
\label{sec:apdxtraininfo}

\myparagraph{Flow consistency mask}
Given the original image $X_{orig}$ and manipulated image $X_{mod}$, we compute the flow from original to manipulated and from manipulated to original using PWC-Net~\cite{sun2018pwc}, which we denote $U_{om}$ and $U_{mo}$, respectively. %

To compute the flow consistency mask, we transform $U_{mo}$ from the manipulated image space into the original image space, which is $U'_{mo} = \mathcal{T} \big( U_{mo};~U_{om} \big)$. We consider the flow to be consistent at a pixel if the magnitude of $U'_{mo}+U_{om}$ is less than a threshold. After this test, pixels corresponding to occlusions and ambiguities (\eg, in low-texture regions) will be marked as inconsistent, and therefore do not contribute to the loss. 

We take relative error of the flow consistency as the criterion. For a pixel $p$,
\begin{equation}
    M_{inconsistent}(p) = \mathbbm{1}\big\{\frac{||U'_{mo}(p)+U_{om}(p)||_2}{||U_{om}(p)||_2 + \epsilon } > \tau \big\}.
    \label{eqn:flowconsistency}
\end{equation}

We take $\epsilon=0.1$ and $\tau=0.85$, then apply a Gaussian blur with $\sigma=7$, denoted by $G$, and take the complement to get the flow consistency mask $M$:
\begin{equation}
    M = 1 - G(M_{inconsistent})
    \label{eqn:flowconsistency2}
\end{equation}

\myparagraph{Training details for local prediction networks}
We use a two-stage training curriculum, where we first train a per-pixel 121-class classifier to predict the discretized warping field. We round the flow values into the closest integer, and assign class to each integer $(u, v)$ value with a cutoff at 5 pixels.  Therefore, we have $u, v \in \{-5, -4, \dots, 4, 5\}$, \ie 121 classes in total. We pretrained the model for 100k iterations with batch size 16. Our strategy is consistent to Zhang et al.~\cite{zhang2016colorful}, which found that (in the context of colorization) pretraining with multinomial classification and then fine-tuning for regression gave better performance than just training for regression directly.

The base-network of the regression model is initialized with the pretrained model weights, and the other weights are initialized with normal distribution with gain $0.02$. We train the models for 250k iterations with batch size 32.

Both models are trained with Adam optimizer~\cite{kingma2014adam} with learning rate $10^{-4}$, $\beta_1 = 0.9, \beta_2 = 0.999$.\\

\myparagraph{Training details for global classification networks}
\camready{
We initialized the base-network of the DRN-C-26~\cite{Yu2017} network with the weights pretrained on the local detection task, and fine-tuned it for the global classification task.
We use the Adam optimizer~\cite{kingma2014adam} with $\beta_1 = 0.9, \beta_2 = 0.999$, minibatch size 32 and 16 for the low and high-res models, respectively, and initial learning rate $10^{-4}$, reduced by $10\times$ when loss plateaus. 
The models are trained for 300k iterations on 135.4k original images and 812.4k modified images, where the original images are sampled $6\times$ more frequently to balance the class distribution.
}

\end{document}